\documentclass[runningheads]{llncs}
\usepackage[a4paper,left=3cm,right=3cm]{geometry}
\usepackage[T1]{fontenc}
\usepackage{graphicx}
\usepackage{amsfonts}
\usepackage{amsmath}
\usepackage{algorithm}
\usepackage{algpseudocode}
\usepackage{censor,caption}
\usepackage{multirow}
\usepackage{graphicx,verbatim}
\usepackage{color}
\usepackage{booktabs}
\usepackage{array}
\usepackage{longtable}
\usepackage{tabularx}
\usepackage{threeparttable}
\usepackage{caption}
\usepackage{xurl}
\usepackage{amssymb}
\usepackage{xcolor}
\usepackage[colorlinks=true,linkcolor=blue,citecolor=blue,urlcolor=blue]{hyperref}

\begin{document}
\title{Automated grounding verification of AI statements about chest radiographs}

\author{
Soroosh Tayebi Arasteh\inst{1,2}$^{\ast}$ \and
Mehdi Joodaki\inst{3} \and
Mahshad Lotfinia\inst{4} \and
Tri-Thien Nguyen\inst{4,5} \and
Sebastian Ziegelmayer\inst{6} \and
Sven Nebelung\inst{1,2} \and
Daniel Truhn\inst{1,2}
}
\institute{
Lab for AI in Medicine, RWTH Aachen University, Aachen, Germany \and
Department of Diagnostic and Interventional Radiology, University Hospital RWTH Aachen, Aachen, Germany \and
Institute of Computational Genomics, RWTH Aachen University, Aachen, Germany \and
Pattern Recognition Lab, Friedrich-Alexander-Universit\"at Erlangen-N\"urnberg, Erlangen, Germany \and
Institute of Radiology, University Hospital Erlangen, Erlangen, Germany \and
Department of Diagnostic and Interventional Radiology, TUM University Clinic, School of Medicine and Health, Klinikum rechts der Isar, Technical University of Munich, Munich, Germany
}

\maketitle
{\footnotesize
\noindent$^{\ast}$Correspondence to:
Soroosh Tayebi Arasteh (\email{soroosh.arasteh@rwth-aachen.de})
}

\begin{abstract}
A grounding verifier checks a statement about a patient's image against the radiology report and the cited evidence. The state-of-the-art methods for testing these verifiers remove or replace the cited evidence while the patient's report stays unchanged. To our knowledge, no existing method checks both the report and the cited evidence. We present Building Report And Citation Examples (BRACE), an automated framework that builds training examples from public radiology reports and Radiopaedia articles. We compared a lexical baseline and 42 verifiers with 0.1--284B parameters on 5 different public chest radiograph datasets. At a threshold accepting 0.90 of the supported test examples, verifiers trained with BRACE accepted a mean of only 0.239 of AI-written statements contradicted by the report. Verifiers trained without contradicted examples passed the state-of-the-art tests, yet accepted most of these statements (0.830 on average). With citations retrieved from held-out Radiopaedia articles, verifiers trained with BRACE accepted 0.169 of AI-written statements with an agreeing report and an unsupporting citation. By contrast, zero-shot large language models mostly checked the report and not the citation (they accepted 0.830 of these statements). We further assessed the framework manually, with two board-certified radiologists independently labeling a sample of test reports. The framework's labels agreed with each radiologist ($\kappa \geq 0.910$) almost as closely as the radiologists agreed with each other ($\kappa = 0.952$). Moreover, instantiating the framework for food allergens in recipes, we showed that it extends beyond clinical applications. A verifier built the same way accepted 0.493 of the statements contradicted by the recipe, whereas a lexical baseline accepted 0.898. Using both the report and the cited evidence, verifiers trained with BRACE could check AI statements before clinicians read them.
\end{abstract}

\section*{Introduction}

\begin{figure*}[!ht]
\centering
\includegraphics[width=\textwidth]{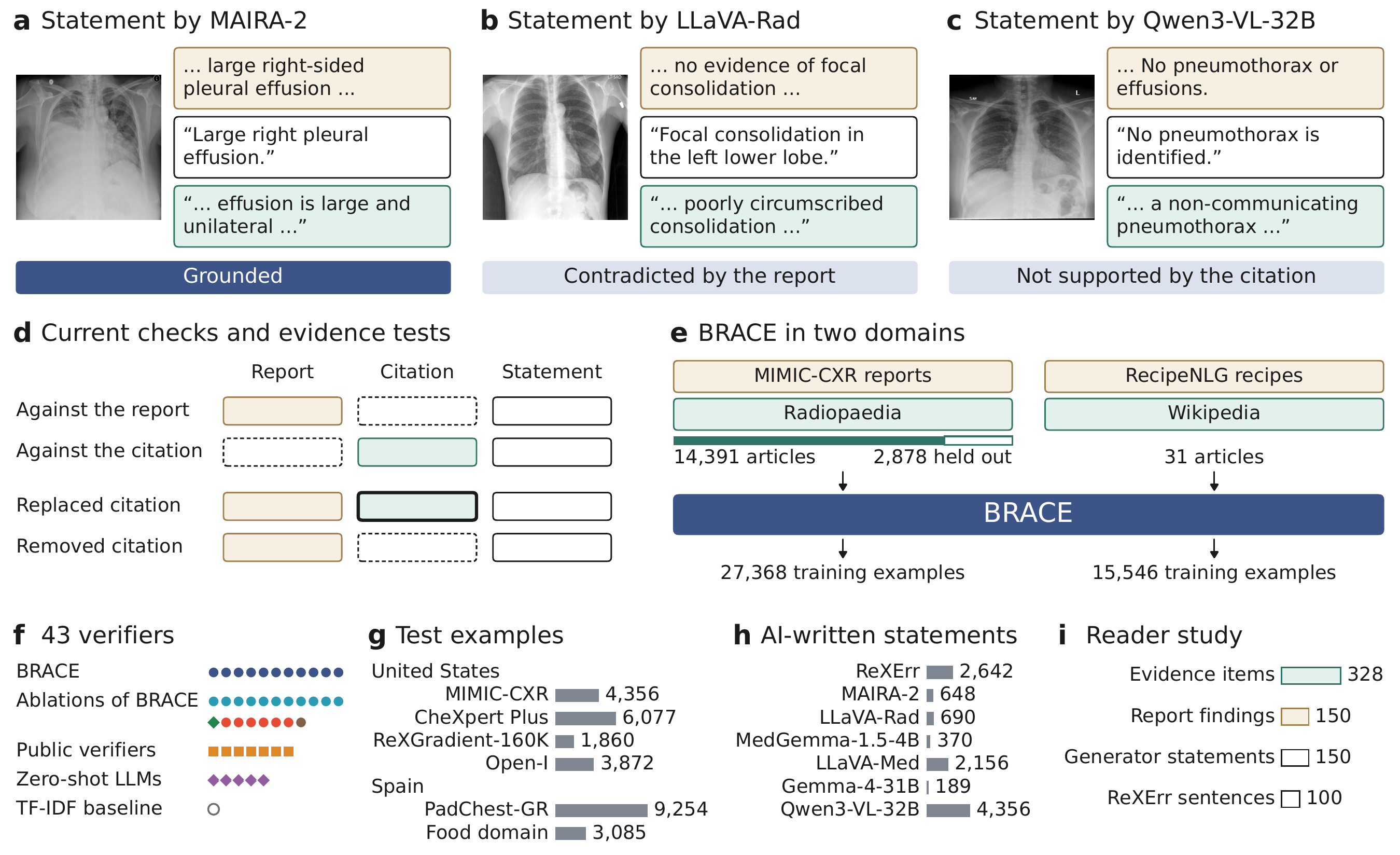}
\caption{From statements written by artificial intelligence (AI) to the study design. AI systems write statements about a patient's radiograph and cite reference text for them. Of three statements written by report generators for Open-I~\cite{demnerfushman2016preparing} radiographs, only the statement of MAIRA-2~\cite{bannur2024maira2} is grounded (\textbf{a}). The report contradicts the statement of LLaVA-Rad~\cite{zambranochaves2025llavarad} (\textbf{b}). The statement of Qwen3-VL-32B~\cite{qwen2025qwen3vl} agrees with its report, but its citation does not support it (\textbf{c}). Each statement is cited with Radiopaedia~\cite{gaillard2011radiopaedia} text. BM25~\cite{robertson2009probabilistic} retrieved the citations of \textbf{a} and \textbf{c} from articles held out from training (outlined in \textbf{e}). Current checks read either the report or the citation (\textbf{d}). A check against the report alone would accept the statement of \textbf{c}. A check against the citation alone would accept the statement of \textbf{b}. The state-of-the-art evidence tests replace the citation (thick outline) or remove it (dashed box) but do not change the report (\textbf{d}). Therefore, a verifier that ignores the report can pass them. To train verifiers that read the report and the citation, Building Report And Citation Examples (BRACE) builds training examples from radiology reports with Radiopaedia text and from recipes with Wikipedia text (\textbf{e}). We test 43 verifiers (\textbf{f}) on five chest radiograph test sets (\textbf{g}) and on AI-written statements (\textbf{h}). A verifier built with BRACE for food allergens is tested on recipes (\textbf{g}). The ablations of BRACE are trained with fewer support sentences (cyan and green), without contradicted examples (red), or on rule-selected support (brown). Two radiologists read samples of the support sentences, the report findings, and the AI-written statements (\textbf{i}). Each marker is one verifier. The bars of \textbf{g} and \textbf{h} share one scale. Colors mark reports and recipes (beige), reference text (green), and BRACE and a grounded statement (navy). Ellipses mark shortened text. LLM, large language model; TF-IDF, term frequency-inverse document frequency.}
\label{fig:story}
\end{figure*}

Generative artificial intelligence (AI) systems now draft chest radiograph reports~\cite{huang2023generative,tanno2025collaboration,buess2025large}, analyze chest radiograph datasets~\cite{tayebi2024treasure}, and answer radiology questions with text retrieved from reference sources~\cite{tayebi2025radiorag,wind2025multi,farajiamiri2026retrieval}. Board-certified radiologists found clinically significant errors only in the AI-drafted report for 22.8\% of inpatient and outpatient cases~\cite{tanno2025collaboration}. Prior work checks generated report sentences against the reference report~\cite{bannur2024maira2,ostmeier2024green} and checks measurements in generated reports against the image~\cite{heiman2025factchexcker}. Other work checks clinical text written by large language models (LLMs) against the patient's record~\cite{chung2025verifact}, medical statements against the record and medical knowledge~\cite{qu2025medieval}, and LLM outputs against grounding documents~\cite{tang2024minicheck}. A grounding verifier is a model that checks an AI-written statement about a patient's image against both the radiology report and the cited evidence (Fig.~\ref{fig:story}a--c).

For example, a Radiopaedia~\cite{gaillard2011radiopaedia} sentence stating that there is no associated cardiomegaly supports the statement ``cardiomegaly is absent'' for a patient whose report describes a normal heart size. The same statement with the same cited sentence is wrong for a patient whose report describes cardiomegaly. For the first patient, the statement is unsupported when the cited sentence says nothing about the absence of cardiomegaly. A verifier that reads only the evidence and the statement assigns the two patients the same score. A verifier that reads only the report assigns the two citations the same score.
Labels for training a verifier can be generated by rules~\cite{mintz2009distant,ratner2017data}, as the finding labels of public chest radiograph datasets are~\cite{peng2018negbio,chexpertmain,smit2020chexbert}. A verifier is usually evaluated by its area under the receiver operating characteristic curve (AUROC) or accuracy on held-out examples and at other hospitals. The state-of-the-art evidence tests remove or replace the cited evidence while the patient's report stays unchanged~\cite{deyoung2020eraser,joren2025sufficient,ming2025faitheval}. Therefore, a verifier that ignores the report can pass those tests. Such shortcuts~\cite{geirhos2020shortcut} are known from accurate chest radiograph classifiers~\cite{zech2018variable,degrave2021ai}. However, to our knowledge, no existing method checks both the report and the cited evidence (Fig.~\ref{fig:story}d).

\begin{figure*}[p]
\centering
\includegraphics[width=\textwidth]{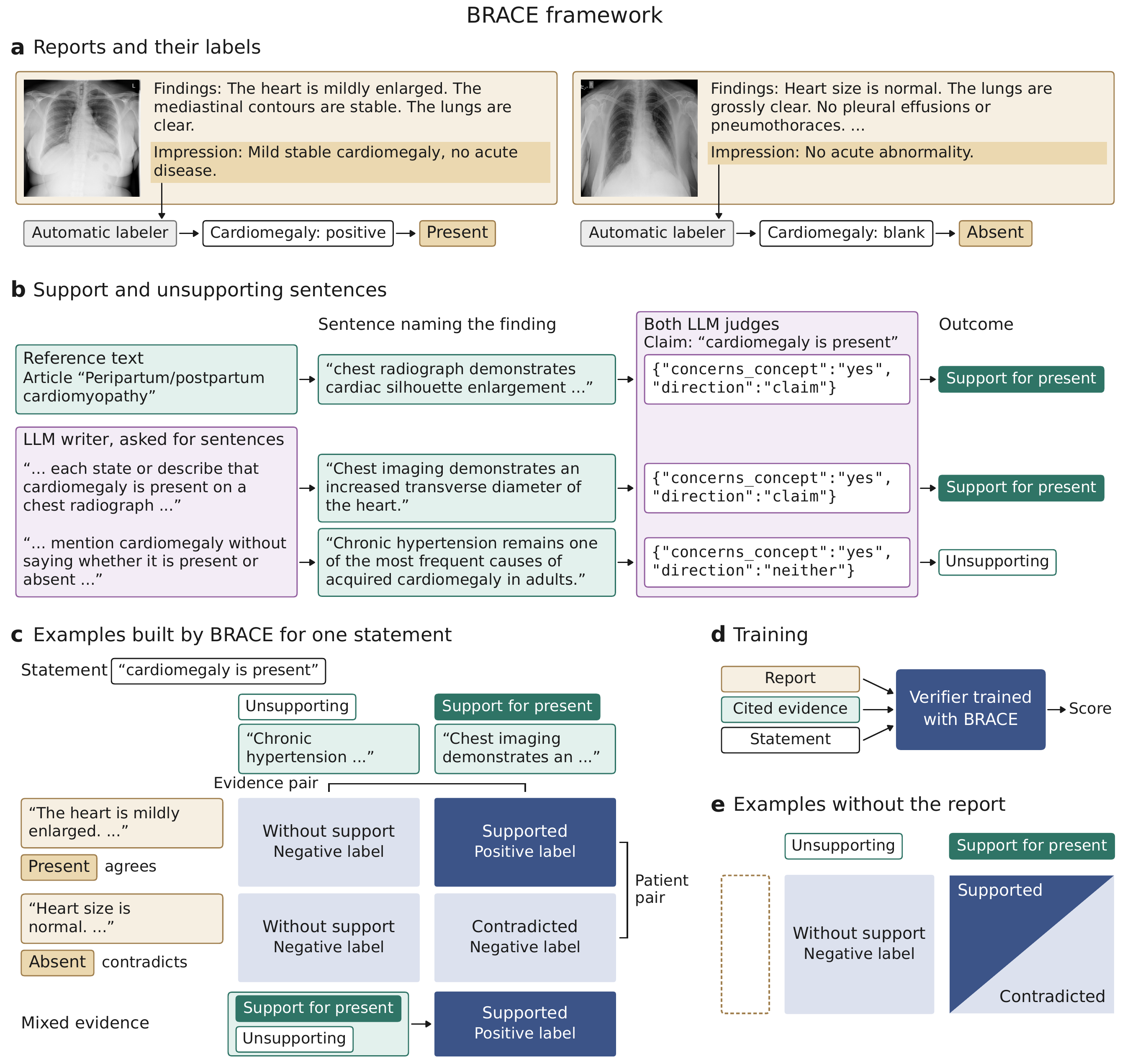}
\caption{BRACE, shown for the finding cardiomegaly. Colors mark reports (beige), evidence (green), LLMs (purple), and BRACE (navy). Ellipses mark shortened text. \textbf{a}, From the highlighted impression, an automatic labeler assigns the finding a class. We map this class to a state. Blank means not mentioned. \textbf{b}, Each LLM judge receives one sentence and the claim that the finding is present. It answers whether the sentence concerns the finding and whether the sentence points toward the claim, the opposite state, or neither. A sentence assigned to the same state by both judges supports that state. A sentence assigned to no state by either judge is unsupporting. \textbf{c}, The label is positive only where the report agrees with the statement and the sentence supports it. The two examples of a patient pair differ only in the report. The two examples of an evidence pair differ only in the evidence. With mixed evidence, the label is unchanged. \textbf{d}, The verifier scores a report, the cited evidence, and a statement. \textbf{e}, Without the report (dashed box), the two examples of a patient pair have the same input. We compute the highest area under the receiver operating characteristic curve (AUROC) achievable when examples with identical evidence and statement receive the same score, i.e., the AUROC attainable without the report. The reports and radiographs are from Open-I~\cite{demnerfushman2016preparing}, and the reference sentence is from Radiopaedia~\cite{gaillard2011radiopaedia}. The labeler is CheXbert~\cite{smit2020chexbert}, the judges are gpt-oss-120b~\cite{openai2025gptoss} and Qwen3.5-122B-A10B~\cite{qwen2026qwen35}, and the writer is Qwen3.5-35B-A3B.}
\label{fig:framework}
\end{figure*}

Here, we present Building Report And Citation Examples (BRACE), an automated framework that builds training examples for grounding verifiers from public radiology reports and Radiopaedia reference texts (Fig.~\ref{fig:framework}). For every report finding, it constructs four types of training examples: (i) a supported example; (ii) a contradicted example, whose statement has supporting evidence and disagrees with the report; (iii) an example without supporting evidence for the statement that the finding is present; and (iv) an example without supporting evidence for the statement that it is absent. A patient pair is a supported and a contradicted example with the same statement and cited evidence and different reports. The report labels come from an automatic labeler. Two LLM judges~\cite{zheng2023judging} select the support sentences. A third LLM writes further support sentences. We compute the AUROC attainable without the report, i.e., the highest AUROC of any verifier that receives only the evidence and the statement. A verifier with a higher AUROC uses the report.

We build the examples from the reports of MIMIC-CXR~\cite{johnson2019mimic} and 14{,}391 Radiopaedia articles, of which 2{,}878 are held out from training. We train eight transformer verifiers and an open LLM finetuned with low-rank adaptation (LoRA)~\cite{hu2022lora}. For ModernBERT-large~\cite{warner2024modernbert} and the LLM, we also train versions without contradicted examples. We compare them with seven public verifiers, five zero-shot LLMs, and a term frequency-inverse document frequency (TF-IDF) baseline that does not receive the report. Counting every seed and training variant, we test 42 verifiers with 0.1--284B parameters and the TF-IDF baseline on 5 different public chest radiograph datasets from the United States and Spain, namely MIMIC-CXR, CheXpert Plus~\cite{baker2024chexpertplus}, ReXGradient-160K~\cite{zhang2025rexgradient}, Open-I~\cite{demnerfushman2016preparing}, and PadChest-GR~\cite{castro2024padchestgr} (Fig.~\ref{fig:story}e--i). We further test the verifiers on AI-written statements from six report generators and ReXErr~\cite{rao2024rexerr}. For a manual assessment of BRACE, two board-certified radiologists independently labeled a sample of the test reports. For generalization assessment beyond radiology in a second domain, we instantiate BRACE for food allergens in RecipeNLG recipes~\cite{bien2020recipenlg}, with Wikipedia sentences as the evidence.

Verifiers trained with BRACE use both the report and the cited evidence on every test set. Without contradicted examples in training, verifiers pass the state-of-the-art evidence tests, yet make little use of the report. Trained on at most four support sentences per finding state, verifiers memorize them and reject most support sentences from the held-out articles. Each verifier accepts more held-out support sentences after training with BRACE. Verifiers trained with BRACE reject most ReXErr statements contradicted by the report, whereas verifiers trained without contradicted examples accept most of them. With citations retrieved from the held-out articles, verifiers trained with BRACE also reject most AI-written statements with an agreeing report and an unsupporting citation. By contrast, the zero-shot LLMs mostly check the report and not the citation, since they accept most of these statements. Most public verifiers also accept them. Each radiologist agrees with the report labels of BRACE almost as closely as with the other radiologist. BRACE also extends beyond clinical applications. In the food domain, a verifier trained with BRACE rejects more of the statements contradicted by the recipe than the TF-IDF baseline does. Supplementary Tables~\ref{stab:glossary} and~\ref{stab:findings} define the terms and summarize the findings. Built with BRACE from public reports and a reference text, verifiers that use both the report and the cited evidence could check AI-written statements and their citations before clinicians read them.

\section*{Results}

We briefly summarize our evaluation metrics and definitions. Patient pair accuracy is the share of patient pairs in which the supported example scores higher than the contradicted example, with ties counted as one half. It is 0.5 for a verifier that does not receive the report. Evidence pair accuracy is the same share for a supported example and the example without supporting evidence of the same report finding. The accepted share is the share of examples or statements scored at or above $\tau_{0.90}$, the threshold at which the accepted share of the supported MIMIC-CXR test examples of a verifier is 0.90. Values are mean $\pm$ standard deviation [95\% confidence interval (CI)] from the bootstrap~\cite{efron1994introduction}, and attainable AUROCs are exact. Two-sided $p$-values are corrected with the Benjamini-Hochberg false discovery rate (FDR)~\cite{benjamini1995controlling} within each family of tests, written $p_{\mathrm{FDR}}$, and considered significant below 0.05.

\subsection*{Verifiers trained with BRACE use the report and cited evidence}

We built the training examples at three levels of training evidence (Table~\ref{tab:datasets}). Level 1 is the pool of four sentences, i.e., the first four Radiopaedia sentences of each finding state accepted by both LLM judges, gpt-oss-120b~\cite{openai2025gptoss} and Qwen3.5-122B-A10B~\cite{qwen2026qwen35}. Level 2 uses every judge-accepted sentence of the training articles. Level 3 is BRACE, which adds sentences written by Qwen3.5-35B-A3B~\cite{qwen2026qwen35} and accepted by both judges. The verifiers trained with BRACE are the 11 verifiers trained on level 3, namely 8 transformer classifiers, 2 further seeds of ModernBERT-large~\cite{warner2024modernbert}, and Qwen3-8B~\cite{yang2025qwen3} finetuned with LoRA. We compare them with 7 public verifiers and 5 zero-shot LLMs (Supplementary Table~\ref{stab:models}).

\begin{table*}[p]
\centering
\caption{Datasets of the study. The first block lists the five chest radiograph test sets and Open-I with the labels of its Medical Subject Headings (MeSH) major topics, with their studies, patients, report findings by state, and test examples by type. The second block lists the MIMIC-CXR training and validation sets of every training set, with the distinct sentences of their support pools and the examples with mixed evidence. The third block lists the statements written by artificial intelligence (AI) in the template form with the training citation, by their label under the definition named in the row, and the statements with a held-out citation. The last block lists the food domain, with recipes in place of studies. Not applicable (N/A) marks patients of Open-I, which releases no patient identifier.}
\label{tab:datasets}
{\scriptsize\hyphenpenalty=10000\exhyphenpenalty=10000
\setlength{\tabcolsep}{3pt}
\renewcommand{\arraystretch}{1.0}
\begin{tabular}{@{}>{\raggedright\arraybackslash}p{0.25\textwidth}rrrrrrr@{}}
\toprule
\addlinespace[2.5pt]
\multicolumn{8}{@{}l}{\textbf{Test sets}} \\
Set & \begin{tabular}[b]{@{}r@{}}Studies\end{tabular} & \begin{tabular}[b]{@{}r@{}}Patients\end{tabular} & \begin{tabular}[b]{@{}r@{}}Present\\findings\end{tabular} & \begin{tabular}[b]{@{}r@{}}Absent\\findings\end{tabular} & \begin{tabular}[b]{@{}r@{}}Supported\end{tabular} & \begin{tabular}[b]{@{}r@{}}Contradicted\end{tabular} & \begin{tabular}[b]{@{}r@{}}Without\\support\end{tabular} \\
\midrule
MIMIC-CXR~\cite{johnson2019mimic} & 655 & 548 & 952 & 249 & 1{,}109 & 845 & 2{,}402 \\
CheXpert Plus~\cite{baker2024chexpertplus} & 853 & 715 & 1{,}141 & 528 & 1{,}508 & 1{,}231 & 3{,}338 \\
ReXGradient-160K~\cite{zhang2025rexgradient} & 395 & 395 & 261 & 261 & 408 & 408 & 1{,}044 \\
Open-I~\cite{demnerfushman2016preparing} & 849 & N/A & 524 & 524 & 888 & 888 & 2{,}096 \\
PadChest-GR~\cite{castro2024padchestgr} & 1{,}934 & 1{,}903 & 1{,}241 & 1{,}241 & 2{,}145 & 2{,}145 & 4{,}964 \\
Open-I (MeSH labels) & 1{,}292 & N/A & 823 & 823 & 1{,}346 & 1{,}346 & 3{,}292 \\
\midrule
\addlinespace[2.5pt]
\multicolumn{8}{@{}l}{\textbf{Training and validation sets (MIMIC-CXR)}} \\
Set & \begin{tabular}[b]{@{}r@{}}Studies\end{tabular} & \begin{tabular}[b]{@{}r@{}}Patients\end{tabular} & \begin{tabular}[b]{@{}r@{}}Report\\findings\end{tabular} & \begin{tabular}[b]{@{}r@{}}Training\\examples\end{tabular} & \begin{tabular}[b]{@{}r@{}}Validation\\examples\end{tabular} & \begin{tabular}[b]{@{}r@{}}Support\\sentences\end{tabular} & \begin{tabular}[b]{@{}r@{}}Mixed\\evidence\end{tabular} \\
\midrule
Pool of four sentences (level 1) & 4{,}095 & 3{,}370 & 7{,}611 & 24{,}687 & 2{,}783 & 30 & 0 \\
Every judge-accepted sentence (level 2) & 4{,}095 & 3{,}370 & 7{,}611 & 24{,}687 & 2{,}783 & 54 & 0 \\
With written sentences (level 3, BRACE) & 4{,}095 & 3{,}370 & 7{,}611 & 27{,}368 & 3{,}076 & 500 & 4{,}564 \\
Level 1 without contradicted examples (three variants) & 4{,}095 & 3{,}370 & 7{,}611 & 19{,}932 & 2{,}244 & 30 & 0 \\
BRACE without contradicted examples & 4{,}095 & 3{,}370 & 7{,}611 & 20{,}526 & 2{,}307 & 500 & 2{,}267 \\
Rule-selected support & 4{,}095 & 3{,}370 & 7{,}611 & 22{,}153 & 2{,}478 & 12 & 0 \\
\midrule
\addlinespace[2.5pt]
\multicolumn{8}{@{}l}{\textbf{AI-written statements (template form)}} \\
Set & \begin{tabular}[b]{@{}r@{}}Statements\end{tabular} & \begin{tabular}[b]{@{}r@{}}Reports\end{tabular} & \begin{tabular}[b]{@{}r@{}}Patients\end{tabular} & \begin{tabular}[b]{@{}r@{}}Consistent\end{tabular} & \begin{tabular}[b]{@{}r@{}}Contradicted\end{tabular} & \begin{tabular}[b]{@{}r@{}}Excluded\end{tabular} & \begin{tabular}[b]{@{}r@{}}With held-out\\citation\end{tabular} \\
\midrule
ReXErr~\cite{rao2024rexerr} (agree definition) & 2{,}642 & 1{,}320 & 485 & 1{,}563 & 472 & 607 & 1{,}466 \\
MAIRA-2~\cite{bannur2024maira2} (report definition) & 648 & 592 & N/A & 349 & 299 & 0 & 308 \\
LLaVA-Rad~\cite{zambranochaves2025llavarad} (report definition) & 690 & 485 & N/A & 484 & 206 & 0 & 272 \\
MedGemma-1.5-4B~\cite{sellergren2025medgemma} (report definition) & 370 & 204 & N/A & 268 & 102 & 0 & 170 \\
LLaVA-Med~\cite{li2023llavamed} (report definition) & 2{,}156 & 2{,}154 & N/A & 45 & 2{,}111 & 0 & 2{,}147 \\
Gemma-4-31B~\cite{gemma2026gemma4} (report definition) & 189 & 138 & N/A & 125 & 64 & 0 & 61 \\
Qwen3-VL-32B~\cite{qwen2025qwen3vl} (report definition) & 4{,}356 & 3{,}067 & N/A & 4{,}116 & 240 & 0 & 281 \\
\midrule
\addlinespace[2.5pt]
\multicolumn{8}{@{}l}{\textbf{Food domain}} \\
Set & \begin{tabular}[b]{@{}r@{}}Recipes\end{tabular} & \begin{tabular}[b]{@{}r@{}}Recipe\\allergens\end{tabular} & \begin{tabular}[b]{@{}r@{}}Present\end{tabular} & \begin{tabular}[b]{@{}r@{}}Absent\end{tabular} & \begin{tabular}[b]{@{}r@{}}Supported\end{tabular} & \begin{tabular}[b]{@{}r@{}}Contradicted\end{tabular} & \begin{tabular}[b]{@{}r@{}}Without\\support\end{tabular} \\
\midrule
Training & 2{,}825 & 3{,}915 & 1{,}987 & 1{,}928 & 3{,}860 & 3{,}856 & 7{,}830 \\
Validation & 371 & 528 & 246 & 282 & 519 & 524 & 1{,}056 \\
Test & 564 & 775 & 376 & 399 & 768 & 767 & 1{,}550 \\
\bottomrule
\end{tabular}
}
\end{table*}

\begin{figure*}[p]
\centering
\includegraphics[width=\textwidth]{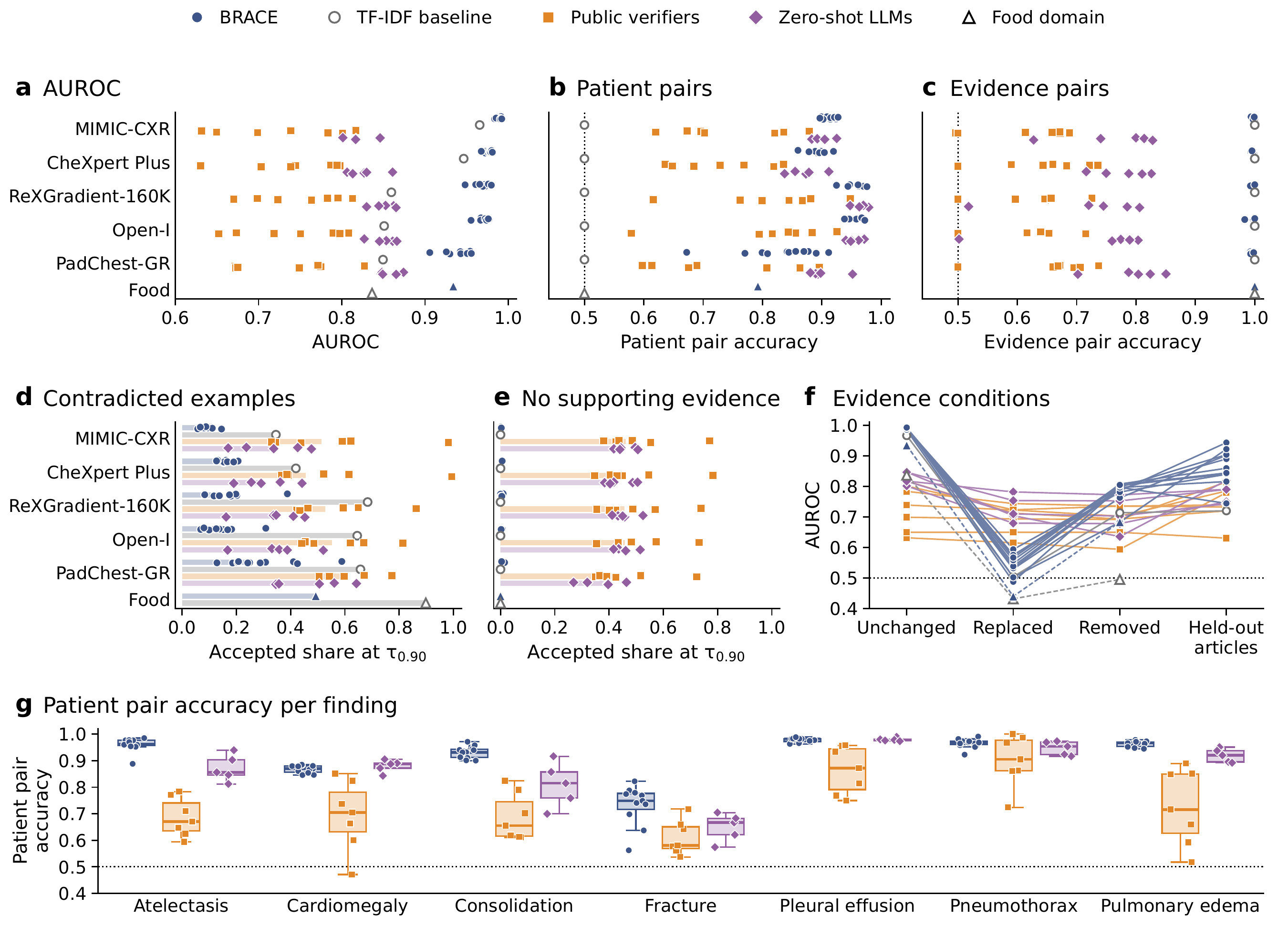}
\caption{Standard validation and patient-level checks on 5 radiology test sets and in the food domain ($n = 1{,}860$ to $9{,}254$ examples per radiology set and $3{,}085$ in the food domain). Colors mark the 11 verifiers trained with BRACE (navy), the seven public verifiers (orange), and the five zero-shot LLMs (purple). Open gray markers represent the TF-IDF baselines. Triangles mark the food domain. Each marker is the bootstrap mean of one verifier over 1,000 resamples of patients, reports on Open-I, or recipes. Bars in \textbf{d} and \textbf{e} are the medians of a family. Each box in \textbf{g} spans the 25th to the 75th percentile of a family, with a line at the median and whiskers to the most extreme value within 1.5 times the interquartile range. No panel has error bars. The 95\% confidence intervals (CIs) of the MIMIC-CXR~\cite{johnson2019mimic} values of \textbf{a}--\textbf{d} are drawn in Supplementary Fig.~\ref{sfig:mimic}. Counts per panel: per radiology set, 15,984 to 588,892 patient pairs (\textbf{b}), 408 to 2,145 evidence pairs (\textbf{c}) and contradicted examples (\textbf{d}), and 1,044 to 4,964 examples without supporting evidence (\textbf{e}); in the food domain, 21,015 recipe pairs, 768 evidence pairs, 767 contradicted examples, and 1,550 examples without supporting evidence; on MIMIC-CXR, 3,697 examples with held-out evidence (\textbf{f}) and 133 to 16,448 patient pairs per finding (\textbf{g}). \textbf{a}--\textbf{c}, AUROC (\textbf{a}), patient or recipe pair accuracy (\textbf{b}), and evidence pair accuracy (\textbf{c}), with one row per test set and one sub-row per family. \textbf{d},\textbf{e}, Accepted shares of the contradicted examples (\textbf{d}) and of the examples without supporting evidence (\textbf{e}) at $\tau_{0.90}$. For each verifier, $\tau_{0.90}$ is the threshold at which the accepted share of its supported MIMIC-CXR or food test examples is 0.90. GREEN~\cite{ostmeier2024green} is not drawn because its threshold accepts every supported example. \textbf{f}, AUROC on MIMIC-CXR and on the food test set under each evidence condition, with one line per verifier and dashed lines for the food domain. \textbf{g}, Patient pair accuracy per finding on MIMIC-CXR, without the TF-IDF baseline, which has 0.5 for every finding by construction. Dotted lines mark 0.5.}
\label{fig:panel}
\end{figure*}

On MIMIC-CXR, the verifiers trained with BRACE use both the report and the cited evidence (Fig.~\ref{fig:panel}a--c and Supplementary Fig.~\ref{sfig:mimic}a). They have an AUROC of 0.984 to 0.992, a patient pair accuracy of 0.897 to 0.927, and an evidence pair accuracy of 0.994 to 1.000. The zero-shot LLMs and the public verifiers use the cited evidence less, with an evidence pair accuracy of 0.627 to 0.828 for the LLMs and 0.496 to 0.688 for the public verifiers. The LLMs use the report about as much, with a patient pair accuracy of 0.883 to 0.925. The public verifiers use it less (0.620 to 0.879), although the interval of three of them overlaps that of a verifier trained with BRACE. A high AUROC is no evidence that a verifier uses the report. The TF-IDF baseline does not receive the report. Its AUROC of 0.966 is higher than that of every LLM and every public verifier.

At $\tau_{0.90}$ on MIMIC-CXR, the verifiers trained with BRACE reject most contradicted examples, with an accepted share of 0.058 to 0.145, and nearly all examples without supporting evidence, with an accepted share of 0.001 to 0.006 (Fig.~\ref{fig:panel}d,e). A verifier passes the state-of-the-art evidence tests when both replacing and removing the evidence lower its AUROC. Except GREEN~\cite{ostmeier2024green}, which does not receive the evidence, every verifier passes them (Fig.~\ref{fig:panel}f and Supplementary Fig.~\ref{sfig:mimic}). The TF-IDF baseline passes them without the report. Per finding, the verifiers trained with BRACE have their lowest patient pair accuracy for fracture (0.562 to 0.822; Fig.~\ref{fig:panel}g).

The verifiers trained with BRACE also use the report and the cited evidence on the four test sets from other hospitals, although their training reports come from MIMIC-CXR alone (Fig.~\ref{fig:panel}a--c). There, they have an AUROC of 0.906 to 0.981, a patient pair accuracy of 0.672 to 0.976, and an evidence pair accuracy of 0.983 to 1.000. Each of them has its lowest patient pair accuracy on PadChest-GR, the test set from Spain (0.672 to 0.911). On the five test sets, their accepted share of the contradicted examples is 0.058 to 0.388, except for RoBERTa-large~\cite{liu2019roberta}, BiomedBERT-base~\cite{biomedbert}, and Longformer-base~\cite{beltagy2020longformer} on PadChest-GR (0.410 to 0.589).

\subsection*{Training without contradicted examples or with few support sentences}

\begin{figure*}[p]
\centering
\includegraphics[width=\textwidth]{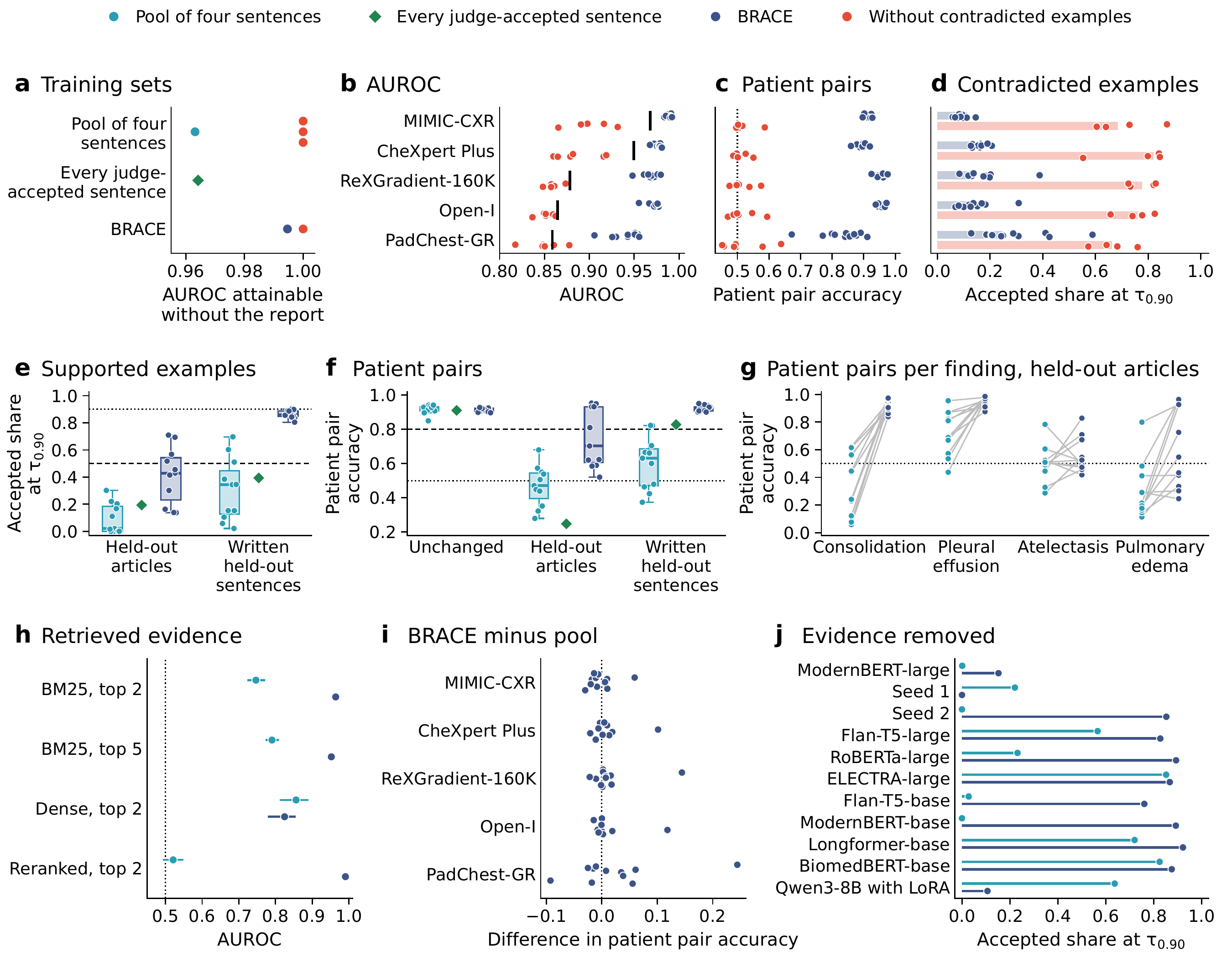}
\caption{Training without contradicted examples and the three levels of training evidence ($n = 1{,}860$ to $9{,}254$ test examples per set). Cyan marks the 11 verifiers trained on the pool of four sentences, and red marks the six verifiers trained without contradicted examples. The green diamond marks ModernBERT-large~\cite{warner2024modernbert} trained on every judge-accepted sentence. Markers are bootstrap means over 1,000 resamples of patients, or of reports on Open-I, except the exact attainable AUROCs (\textbf{a}) and the differences of two means (\textbf{i}). Bars in \textbf{d} are medians of a group. Lollipops in \textbf{j} are point values on the test set. Each box in \textbf{e} and \textbf{f} spans the 25th to the 75th percentile of a level, with a line at the median and whiskers to the most extreme value within 1.5 times the interquartile range. Only \textbf{h} has error bars, the 95\% CIs from the 2.5th to the 97.5th percentile of the resamples. Counts per panel: 19,932 to 27,368 examples per training set (\textbf{a}); per test set, 15,984 to 588,892 patient pairs (\textbf{c}, \textbf{i}) and 408 to 2,145 contradicted examples (\textbf{d}); on MIMIC-CXR, 695 and 1,201 supported examples (\textbf{e}), 33,441, 29,907, and 4,078 patient pairs (\textbf{f}), 477 to 16,720 patient pairs per finding (\textbf{g}), 2,402 retrieved examples per condition with 34 to 335 supported examples (\textbf{h}), and 1,109 supported examples (\textbf{j}). \textbf{a}, Training split of each training set. \textbf{b}--\textbf{d} and \textbf{i}, Test sets with unchanged evidence. Black ticks in \textbf{b} mark the AUROC attainable without the report. Two verifiers without contradicted examples are not drawn in \textbf{d} because their threshold accepts more than 0.95 of the supported examples. Dashed lines mark the criterion for accepting unseen evidence, an accepted share of 0.50 and a patient pair accuracy of 0.80. Gray lines in \textbf{g} join the two values of a verifier. \textbf{h}, ModernBERT-large with evidence retrieved from the held-out articles by BM25, a dense retriever, or BM25 with reranking. Seed 1 and Seed 2 (\textbf{j}) are two further runs of ModernBERT-large. Dotted lines mark 0.5 in \textbf{c}, \textbf{f}, \textbf{g}, and \textbf{h}, 0.90 in \textbf{e}, and 0 in \textbf{i}. LoRA, low-rank adaptation.}
\label{fig:ablations}
\end{figure*}

We trained six verifiers on four training sets without contradicted examples, namely three variants of level 1 and BRACE without its contradicted examples (Supplementary Table~\ref{stab:models}). These verifiers pass the state-of-the-art evidence tests and make little use of the report. The AUROC attainable without the report is 1.000 on each of their training sets. On the training sets with contradicted examples, it is 0.963 at level 1, 0.964 at level 2, and 0.995 at level 3 (Fig.~\ref{fig:ablations}a). On the five test sets, the AUROC interval of the six verifiers is below the AUROC attainable without the report in 25 of the 30 combinations and above it only for Qwen3-8B with LoRA at level 1 on PadChest-GR (Fig.~\ref{fig:ablations}b). Every verifier trained on the three levels has an AUROC above this value on every test set.

The patient pair accuracy of the six verifiers is 0.496 to 0.587 on MIMIC-CXR and 0.452 to 0.639 on the other test sets (Fig.~\ref{fig:ablations}c). On every test set, each of them has a lower patient pair accuracy than every verifier trained with BRACE. The intervals overlap only on PadChest-GR, for the random-evidence variant of ModernBERT-large (0.639) and Longformer-base (0.672). On the five test sets, the accepted share of the contradicted examples is 0.553 to 0.871 for four of the six verifiers (Fig.~\ref{fig:ablations}d).

We also tested the verifiers with support sentences never seen in training, namely 12 sentences of the held-out articles for 6 finding states and 112 written sentences for all 14 states. By a criterion fixed before we trained the verifiers with BRACE, a verifier accepts unseen evidence when its accepted share of the supported examples with these sentences is at least 0.50 and its patient pair accuracy is at least 0.80. Trained on the pool of four sentences, the 11 verifiers reject most supported examples whose support sentences come from the held-out articles (Fig.~\ref{fig:ablations}e,f and Supplementary Fig.~\ref{sfig:unseen}a). Their accepted share of the supported MIMIC-CXR examples is 0.000 to 0.302 with these sentences and 0.021 to 0.697 with the written held-out sentences.

Trained with BRACE, all 11 verifiers meet the criterion on the written held-out sentences, which come from the same writer (Supplementary Fig.~\ref{sfig:unseen}b). Their accepted share of the supported examples is 0.805 to 0.901 with these sentences and 0.137 to 0.709 with the support sentences of the held-out articles, on which 4 of the 11 verifiers meet the criterion. Per finding, their patient pair accuracy with the held-out sentences is higher than after training on the pool in 38 of the 44 combinations of verifier and finding (Fig.~\ref{fig:ablations}g). With evidence retrieved from the held-out articles, the AUROC of ModernBERT-large is 0.825 to 0.991 after training with BRACE and 0.521 to 0.856 after training on the pool (Fig.~\ref{fig:ablations}h). It is higher after training with BRACE in three of the four retrieval conditions. On the unchanged test examples, training with BRACE in place of the pool changes the AUROC of a verifier by $-0.037$ to $+0.050$ and its patient pair accuracy by $-0.092$ to $+0.244$ (Fig.~\ref{fig:ablations}i). Without evidence, 8 of the 11 verifiers trained with BRACE accept most supported examples (0.761 to 0.922), while ModernBERT-large, its first seed, and Qwen3-8B with LoRA reject most of them (0.000 to 0.152; Fig.~\ref{fig:ablations}j).

\subsection*{Checking the report and the citation of AI-written statements}

\begin{figure*}[p]
\centering
\includegraphics[width=\textwidth]{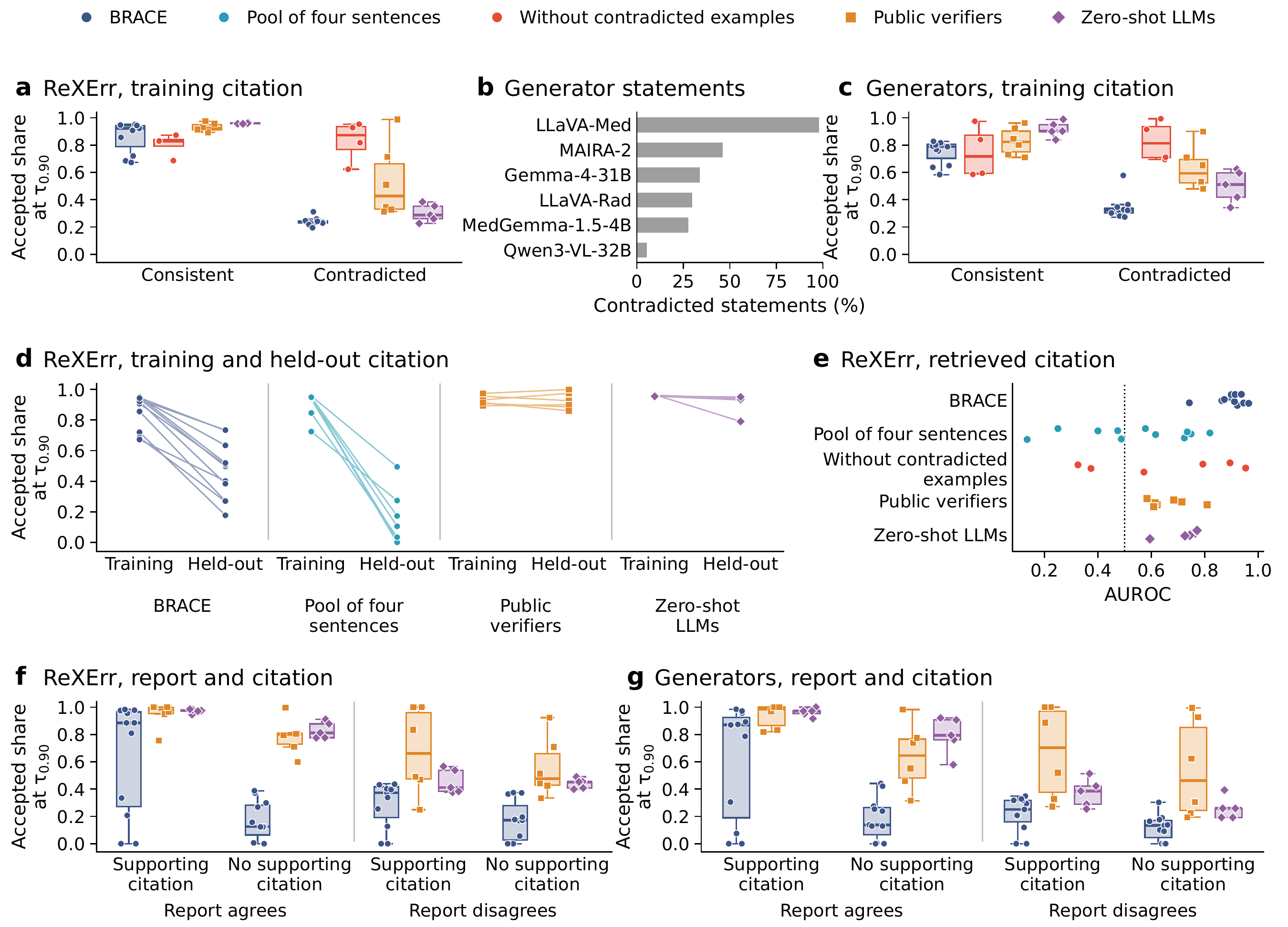}
\caption{AI-written statements with up to three citations ($n = 2{,}642$ ReXErr~\cite{rao2024rexerr} statements from 485 MIMIC-CXR test patients and 8,409 statements of 6 generators on 3,180 Open-I reports). Each marker is the bootstrap mean of one verifier over 1,000 resamples of patients or reports. Markers in \textbf{c} and \textbf{g} are means over the generators with at least one statement, without an interval. Each box spans the 25th to the 75th percentile of a group, with a line at the median and whiskers to the most extreme value within 1.5 times the interquartile range. No panel has error bars. The 95\% CIs of \textbf{a} and \textbf{d} are drawn in Supplementary Fig.~\ref{sfig:rexerr}. Counts per panel: 1,563 consistent and 472 contradicted statements (\textbf{a}); 189 to 4,356 statements per generator (\textbf{b}); 45 to 4,116 consistent and 64 to 2,111 contradicted statements per generator (\textbf{c}); 1,563 and 762 consistent statements (\textbf{d}); 401 positive and 2,241 negative joint labels (\textbf{e}); 401, 1,444, 205, and 592 statements in the four combinations (\textbf{f}); and 0 to 4,109 statements per generator and combination (\textbf{g}). GREEN and two verifiers trained without contradicted examples are left out of the accepted shares, since their threshold accepts more than 0.95 of the supported examples. \textbf{a}, Accepted shares of the consistent and the contradicted ReXErr statements at $\tau_{0.90}$, with a support sentence of the training articles as the citation. \textbf{b}, Share of contradicted statements per generator. \textbf{c}, The shares of \textbf{a} for the generator statements. \textbf{d}, Accepted share of the consistent ReXErr statements with a support sentence of the training or the held-out articles as the citation. Each line joins the two values of one verifier. \textbf{e}, AUROC on the ReXErr statements with the two sentences retrieved by BM25 from the held-out articles as the citation. A positive joint label requires both report agreement and citation support. The dotted line marks 0.5. \textbf{f},\textbf{g}, Accepted shares of the ReXErr statements (\textbf{f}) and the generator statements (\textbf{g}) with the retrieved citation, by report agreement and citation support.}
\label{fig:statements}
\end{figure*}

We tested the verifiers on AI-written statements about one of the seven findings, from ReXErr on the MIMIC-CXR test reports and from six report generators on the Open-I radiographs (Table~\ref{tab:datasets}). A statement is consistent when the report of its study agrees with it and contradicted otherwise. Each statement is scored with up to three citations, namely support sentences of the training articles, support sentences of the held-out articles where its finding state has them, and two sentences retrieved from the held-out articles by BM25~\cite{robertson2009probabilistic}. With the training citation, the verifiers trained with BRACE accept most consistent ReXErr statements (0.673 to 0.951) and reject most contradicted statements, with an accepted share of 0.239 on average (range 0.195 to 0.312; Fig.~\ref{fig:statements}a and Supplementary Fig.~\ref{sfig:rexerr}). Of the verifiers trained without contradicted examples, the four that are compared by their accepted shares accept most contradicted statements (0.830 on average, range 0.623 to 0.953). The zero-shot LLMs also check the report, with an accepted share of 0.226 to 0.383 for the contradicted statements.

Per generator, 6\% (Qwen3-VL-32B~\cite{qwen2025qwen3vl}) to 98\% (LLaVA-Med~\cite{li2023llavamed}) of the statements contradict the report (Fig.~\ref{fig:statements}b). With the generators weighted equally, the accepted share of the contradicted generator statements is 0.273 to 0.578 for the verifiers trained with BRACE and 0.695 to 0.993 for the four verifiers trained without contradicted examples (Fig.~\ref{fig:statements}c). With the held-out citation, the accepted share of the consistent ReXErr statements is 0.177 to 0.734 for the verifiers trained with BRACE and 0.791 to 0.952 for the LLMs (Fig.~\ref{fig:statements}d). Only the intervals of Flan-T5-large and ELECTRA-large overlap the interval of Mistral-Small-4-119B.

A retrieved citation supports a statement when one of its sentences is a support sentence of the stated finding state. Under the label that requires report agreement and citation support, the AUROC is 0.742 to 0.964 for the verifiers trained with BRACE and at most 0.809 for the LLMs and the public verifiers (Fig.~\ref{fig:statements}e). A verifier can have a high AUROC under this label without using the report, as Qwen3-8B with LoRA trained on BRACE without its contradicted examples does (0.953, with a patient pair accuracy of 0.502 on MIMIC-CXR). With the retrieved citation, every verifier trained with BRACE rejects most statements without report agreement or without citation support (Fig.~\ref{fig:statements}f,g). Seven of them accept most statements with report agreement and citation support, while Longformer-base, ModernBERT-large, and its two seeds reject most statements in every combination, in both statement sets. By contrast, the zero-shot LLMs mostly check the report and not the citation. For the ReXErr statements with an agreeing report and an unsupporting citation, the accepted share averages 0.169 over the verifiers trained with BRACE (range 0.000 to 0.388) and 0.830 over the LLMs (range 0.775 to 0.911). The six public verifiers other than GREEN also accept most of these statements (0.598 to 0.997).

Finetuned with LoRA on the training examples of BRACE, Qwen3-8B checks the cited evidence in addition to the report. Its AUROC is higher than that of its zero-shot version on every test set ($+0.090$ to $+0.176$, all $p_{\mathrm{FDR}} < 0.001$; Table~\ref{tab:lora}). Its patient pair accuracy is higher by $+0.013$ to $+0.041$, significantly on three of the five test sets ($p_{\mathrm{FDR}} \leq 0.010$). The finetuned model rejects nearly all MIMIC-CXR examples without supporting evidence (accepted share 0.003), while the zero-shot model accepts about half of them (0.506). For the ReXErr statements with an agreeing report and an unsupporting retrieved citation, the accepted share is 0.123 for the finetuned model and 0.878 for the zero-shot model.

\begin{table*}[p]
\centering
\caption{Qwen3-8B~\cite{yang2025qwen3} zero-shot and finetuned with low-rank adaptation (LoRA)~\cite{hu2022lora}, on the five test sets with unchanged evidence ($n = 1{,}860$ to $9{,}254$ examples) and on the contradicted ReXErr statements with the training citation ($n = 472$). The difference compares the model finetuned with BRACE with the zero-shot model by a paired bootstrap. Contradicted accepted is the share accepted at the threshold at which a model's accepted share of its supported MIMIC-CXR test examples is 0.90. N/A marks the model whose accepted share of these examples at this threshold is above 0.95. Each cell is the bootstrap mean $\pm$ standard deviation [95\% confidence interval (CI)] over 1{,}000 resamples of patients, or of reports on Open-I. AUROC, area under the receiver operating characteristic curve; FDR, false discovery rate.}
\label{tab:lora}
{\scriptsize\hyphenpenalty=10000\exhyphenpenalty=10000
\setlength{\tabcolsep}{4pt}
\renewcommand{\arraystretch}{1.05}
\begin{tabular}{@{}lrrrrr@{}}
\toprule
Metric & \begin{tabular}[b]{@{}r@{}}Zero-shot\end{tabular} & \begin{tabular}[b]{@{}r@{}}LoRA,\\BRACE\end{tabular} & \begin{tabular}[b]{@{}r@{}}Difference\\{}[95\% CI] and $p_{\mathrm{FDR}}$\end{tabular} & \begin{tabular}[b]{@{}r@{}}LoRA, pool of\\four sentences\end{tabular} & \begin{tabular}[b]{@{}r@{}}LoRA, BRACE\\without\\contradicted\\examples\end{tabular} \\
\midrule
\addlinespace[2.5pt]
\multicolumn{6}{@{}l}{\textbf{MIMIC-CXR}} \\
AUROC & \begin{tabular}[t]{@{}r@{}}0.817 $\pm$ 0.006\\{[}0.805, 0.828{]}\end{tabular} & \begin{tabular}[t]{@{}r@{}}0.992 $\pm$ 0.001\\{[}0.989, 0.994{]}\end{tabular} & \begin{tabular}[t]{@{}r@{}}+0.176 [+0.165, +0.186]\\$p_{\mathrm{FDR}} < 0.001$\end{tabular} & \begin{tabular}[t]{@{}r@{}}0.994 $\pm$ 0.001\\{[}0.992, 0.996{]}\end{tabular} & \begin{tabular}[t]{@{}r@{}}0.865 $\pm$ 0.003\\{[}0.860, 0.871{]}\end{tabular} \\
Patient pairs & \begin{tabular}[t]{@{}r@{}}0.905 $\pm$ 0.018\\{[}0.866, 0.937{]}\end{tabular} & \begin{tabular}[t]{@{}r@{}}0.927 $\pm$ 0.019\\{[}0.889, 0.962{]}\end{tabular} & \begin{tabular}[t]{@{}r@{}}+0.022 [$-$0.002, +0.044]\\$p_{\mathrm{FDR}} = 0.064$\end{tabular} & \begin{tabular}[t]{@{}r@{}}0.941 $\pm$ 0.017\\{[}0.905, 0.970{]}\end{tabular} & \begin{tabular}[t]{@{}r@{}}0.502 $\pm$ 0.001\\{[}0.500, 0.505{]}\end{tabular} \\
\addlinespace[2.5pt]
\multicolumn{6}{@{}l}{\textbf{CheXpert Plus}} \\
AUROC & \begin{tabular}[t]{@{}r@{}}0.813 $\pm$ 0.004\\{[}0.805, 0.822{]}\end{tabular} & \begin{tabular}[t]{@{}r@{}}0.981 $\pm$ 0.002\\{[}0.977, 0.984{]}\end{tabular} & \begin{tabular}[t]{@{}r@{}}+0.168 [+0.159, +0.176]\\$p_{\mathrm{FDR}} < 0.001$\end{tabular} & \begin{tabular}[t]{@{}r@{}}0.981 $\pm$ 0.002\\{[}0.977, 0.984{]}\end{tabular} & \begin{tabular}[t]{@{}r@{}}0.865 $\pm$ 0.002\\{[}0.860, 0.869{]}\end{tabular} \\
Patient pairs & \begin{tabular}[t]{@{}r@{}}0.878 $\pm$ 0.013\\{[}0.853, 0.902{]}\end{tabular} & \begin{tabular}[t]{@{}r@{}}0.920 $\pm$ 0.010\\{[}0.898, 0.938{]}\end{tabular} & \begin{tabular}[t]{@{}r@{}}+0.041 [+0.025, +0.060]\\$p_{\mathrm{FDR}} < 0.001$\end{tabular} & \begin{tabular}[t]{@{}r@{}}0.915 $\pm$ 0.011\\{[}0.893, 0.934{]}\end{tabular} & \begin{tabular}[t]{@{}r@{}}0.501 $\pm$ 0.001\\{[}0.499, 0.504{]}\end{tabular} \\
\addlinespace[2.5pt]
\multicolumn{6}{@{}l}{\textbf{ReXGradient-160K}} \\
AUROC & \begin{tabular}[t]{@{}r@{}}0.844 $\pm$ 0.008\\{[}0.828, 0.860{]}\end{tabular} & \begin{tabular}[t]{@{}r@{}}0.980 $\pm$ 0.004\\{[}0.972, 0.986{]}\end{tabular} & \begin{tabular}[t]{@{}r@{}}+0.135 [+0.121, +0.150]\\$p_{\mathrm{FDR}} < 0.001$\end{tabular} & \begin{tabular}[t]{@{}r@{}}0.983 $\pm$ 0.003\\{[}0.976, 0.989{]}\end{tabular} & \begin{tabular}[t]{@{}r@{}}0.848 $\pm$ 0.005\\{[}0.837, 0.858{]}\end{tabular} \\
Patient pairs & \begin{tabular}[t]{@{}r@{}}0.963 $\pm$ 0.012\\{[}0.938, 0.983{]}\end{tabular} & \begin{tabular}[t]{@{}r@{}}0.976 $\pm$ 0.009\\{[}0.956, 0.991{]}\end{tabular} & \begin{tabular}[t]{@{}r@{}}+0.013 [$-$0.007, +0.033]\\$p_{\mathrm{FDR}} = 0.223$\end{tabular} & \begin{tabular}[t]{@{}r@{}}0.969 $\pm$ 0.011\\{[}0.944, 0.988{]}\end{tabular} & \begin{tabular}[t]{@{}r@{}}0.498 $\pm$ 0.002\\{[}0.492, 0.501{]}\end{tabular} \\
\addlinespace[2.5pt]
\multicolumn{6}{@{}l}{\textbf{Open-I}} \\
AUROC & \begin{tabular}[t]{@{}r@{}}0.845 $\pm$ 0.006\\{[}0.834, 0.856{]}\end{tabular} & \begin{tabular}[t]{@{}r@{}}0.976 $\pm$ 0.003\\{[}0.971, 0.981{]}\end{tabular} & \begin{tabular}[t]{@{}r@{}}+0.131 [+0.121, +0.141]\\$p_{\mathrm{FDR}} < 0.001$\end{tabular} & \begin{tabular}[t]{@{}r@{}}0.985 $\pm$ 0.002\\{[}0.981, 0.989{]}\end{tabular} & \begin{tabular}[t]{@{}r@{}}0.837 $\pm$ 0.004\\{[}0.829, 0.844{]}\end{tabular} \\
Patient pairs & \begin{tabular}[t]{@{}r@{}}0.949 $\pm$ 0.010\\{[}0.928, 0.967{]}\end{tabular} & \begin{tabular}[t]{@{}r@{}}0.972 $\pm$ 0.006\\{[}0.959, 0.983{]}\end{tabular} & \begin{tabular}[t]{@{}r@{}}+0.024 [+0.012, +0.038]\\$p_{\mathrm{FDR}} < 0.001$\end{tabular} & \begin{tabular}[t]{@{}r@{}}0.973 $\pm$ 0.006\\{[}0.959, 0.983{]}\end{tabular} & \begin{tabular}[t]{@{}r@{}}0.499 $\pm$ 0.001\\{[}0.497, 0.501{]}\end{tabular} \\
\addlinespace[2.5pt]
\multicolumn{6}{@{}l}{\textbf{PadChest-GR}} \\
AUROC & \begin{tabular}[t]{@{}r@{}}0.865 $\pm$ 0.003\\{[}0.859, 0.871{]}\end{tabular} & \begin{tabular}[t]{@{}r@{}}0.956 $\pm$ 0.002\\{[}0.951, 0.960{]}\end{tabular} & \begin{tabular}[t]{@{}r@{}}+0.090 [+0.083, +0.097]\\$p_{\mathrm{FDR}} < 0.001$\end{tabular} & \begin{tabular}[t]{@{}r@{}}0.968 $\pm$ 0.002\\{[}0.964, 0.971{]}\end{tabular} & \begin{tabular}[t]{@{}r@{}}0.817 $\pm$ 0.003\\{[}0.811, 0.824{]}\end{tabular} \\
Patient pairs & \begin{tabular}[t]{@{}r@{}}0.898 $\pm$ 0.009\\{[}0.881, 0.915{]}\end{tabular} & \begin{tabular}[t]{@{}r@{}}0.911 $\pm$ 0.007\\{[}0.898, 0.925{]}\end{tabular} & \begin{tabular}[t]{@{}r@{}}+0.014 [+0.003, +0.023]\\$p_{\mathrm{FDR}} = 0.009$\end{tabular} & \begin{tabular}[t]{@{}r@{}}0.873 $\pm$ 0.011\\{[}0.852, 0.895{]}\end{tabular} & \begin{tabular}[t]{@{}r@{}}0.452 $\pm$ 0.004\\{[}0.444, 0.460{]}\end{tabular} \\
\addlinespace[2.5pt]
\multicolumn{6}{@{}l}{\textbf{ReXErr statements, training citation}} \\
Contradicted accepted & \begin{tabular}[t]{@{}r@{}}0.383 $\pm$ 0.024\\{[}0.336, 0.429{]}\end{tabular} & \begin{tabular}[t]{@{}r@{}}0.195 $\pm$ 0.020\\{[}0.156, 0.233{]}\end{tabular} & \begin{tabular}[t]{@{}r@{}}$-$0.189 [$-$0.232, $-$0.145]\\$p_{\mathrm{FDR}} < 0.001$\end{tabular} & \begin{tabular}[t]{@{}r@{}}0.180 $\pm$ 0.018\\{[}0.144, 0.215{]}\end{tabular} & N/A \\
\bottomrule
\end{tabular}
}
\end{table*}

\subsection*{Patient pair accuracy and the acceptance of contradicted statements}

\begin{figure*}[p]
\centering
\includegraphics[width=\textwidth]{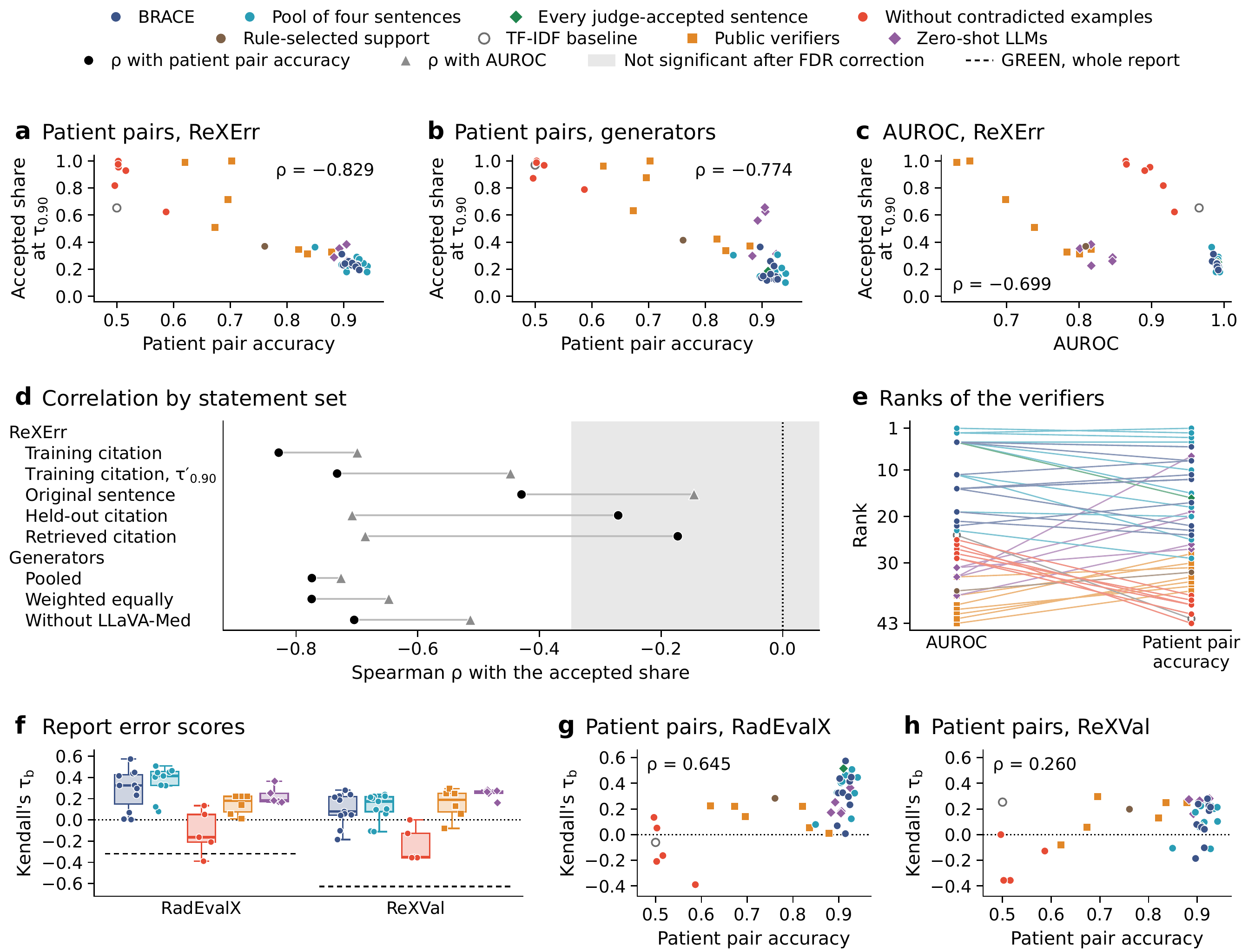}
\caption{Test-set measures, statement acceptance, and report error scores ($n = 43$ verifiers). Accepted shares, AUROCs, and patient pair accuracies are bootstrap means over 1,000 resamples of patients, or of reports for the generators. Every $\rho$ is a Spearman correlation across the verifiers. Each box in \textbf{f} spans the 25th to the 75th percentile of a group, with a line at the median and whiskers to the most extreme value within 1.5 times the interquartile range. No panel has error bars. Counts per panel: 472 contradicted ReXErr statements (\textbf{a}, \textbf{c}) and 3,022 contradicted generator statements (\textbf{b}); 38 verifiers on 33 RadEvalX~\cite{calamida2023radiology} and 68 ReXVal~\cite{yu2023evaluating} candidate reports (\textbf{f}); 41 verifiers (\textbf{g}, \textbf{h}). \textbf{a},\textbf{b}, Accepted share of the contradicted ReXErr statements (\textbf{a}) and of the pooled contradicted generator statements (\textbf{b}) with the training citation at $\tau_{0.90}$, as a function of the patient pair accuracy on the MIMIC-CXR test set. \textbf{c}, The accepted share of \textbf{a} as a function of the AUROC on the same test set. \textbf{d}, $\rho$ of the patient pair accuracy and of the AUROC with the accepted share among negative labels of each set. Each row after the first of its block changes one setting of the first row. The gray band marks $|\rho| < 0.348$. Of the 688 correlations of this analysis, those outside the band have a $p$-value corrected for the false discovery rate (FDR), $p_{\mathrm{FDR}}$, below 0.05. Those inside it have $p_{\mathrm{FDR}} \geq 0.05$. \textbf{e}, Rank of each verifier by the AUROC and by the patient pair accuracy. Lines join the two ranks of one verifier. Values equal at three decimals share a rank. \textbf{f}, Kendall's $\tau_b$ of the report error score of each verifier with the radiologists' counts of clinically significant errors per candidate report. Dashed lines mark $\tau_b$ of the GREEN score of the whole report, which falls as errors rise. \textbf{g},\textbf{h}, $\tau_b$ on RadEvalX (\textbf{g}) and on ReXVal (\textbf{h}) as a function of the patient pair accuracy. Dotted lines mark 0.}
\label{fig:checks}
\end{figure*}

Across the 43 verifiers, the patient pair accuracy on MIMIC-CXR is rank-correlated with the accepted share of the contradicted ReXErr statements with the training citation (Spearman $\rho = -0.829$, $p_{\mathrm{FDR}} < 0.001$; Fig.~\ref{fig:checks}a and Supplementary Table~\ref{stab:correlations}). With the pooled generator statements, $\rho$ is $-0.774$ ($p_{\mathrm{FDR}} < 0.001$; Fig.~\ref{fig:checks}b). The AUROC on MIMIC-CXR is also correlated with the accepted share of the contradicted ReXErr statements ($\rho = -0.699$, $p_{\mathrm{FDR}} < 0.001$; Fig.~\ref{fig:checks}c). In all 96 variants of this analysis with the training citation, $|\rho|$ is larger with the patient pair accuracy than with the AUROC (Fig.~\ref{fig:checks}d). With the original ReXErr sentence in place of the template statement, $\rho$ is $-0.429$ ($p_{\mathrm{FDR}} = 0.010$) with the patient pair accuracy and $-0.146$ ($p_{\mathrm{FDR}} = 0.496$) with the AUROC.

By the AUROC, the TF-IDF baseline and the six verifiers trained without contradicted examples rank directly below the 23 verifiers trained on the three levels (Fig.~\ref{fig:checks}e). By the patient pair accuracy, these seven verifiers rank last. Under the held-out citation, the accepted share is correlated with the AUROC ($\rho = -0.708$, $p_{\mathrm{FDR}} < 0.001$) and not significantly with the patient pair accuracy ($\rho = -0.271$, $p_{\mathrm{FDR}} = 0.154$). Under the retrieved citation, $\rho$ is $-0.173$ with the patient pair accuracy ($p_{\mathrm{FDR}} = 0.409$) and $-0.686$ with the AUROC ($p_{\mathrm{FDR}} < 0.001$).

The report error score of a verifier is the share of the statements of a generated report that score below $\tau_{0.90}$. It is correlated with the radiologists' counts of clinically significant errors for 4 verifiers trained with BRACE on the 33 RadEvalX~\cite{calamida2023radiology} reports with statements (Kendall's $\tau_b = 0.415$ to $0.574$, $p_{\mathrm{FDR}} \leq 0.013$). On the 68 ReXVal~\cite{yu2023evaluating} reports, the correlations for the 11 verifiers trained with BRACE range from $\tau_b = -0.186$ to $0.279$ (all $p_{\mathrm{FDR}} > 0.05$; Fig.~\ref{fig:checks}f and Supplementary Table~\ref{stab:report_metric}). For the GREEN score of the whole report, $\tau_b$ is $-0.319$ on RadEvalX and $-0.626$ on ReXVal, each with $p_{\mathrm{FDR}} < 0.001$. Across the 41 verifiers with a defined score, the patient pair accuracy is correlated with $\tau_b$ on RadEvalX ($\rho = 0.645$, $p_{\mathrm{FDR}} < 0.001$) and not significantly on ReXVal ($\rho = 0.260$, $p_{\mathrm{FDR}} = 0.187$; Fig.~\ref{fig:checks}g,h).

\subsection*{Robustness of the verifiers and of the labels}

\begin{figure*}[p]
\centering
\includegraphics[width=\textwidth]{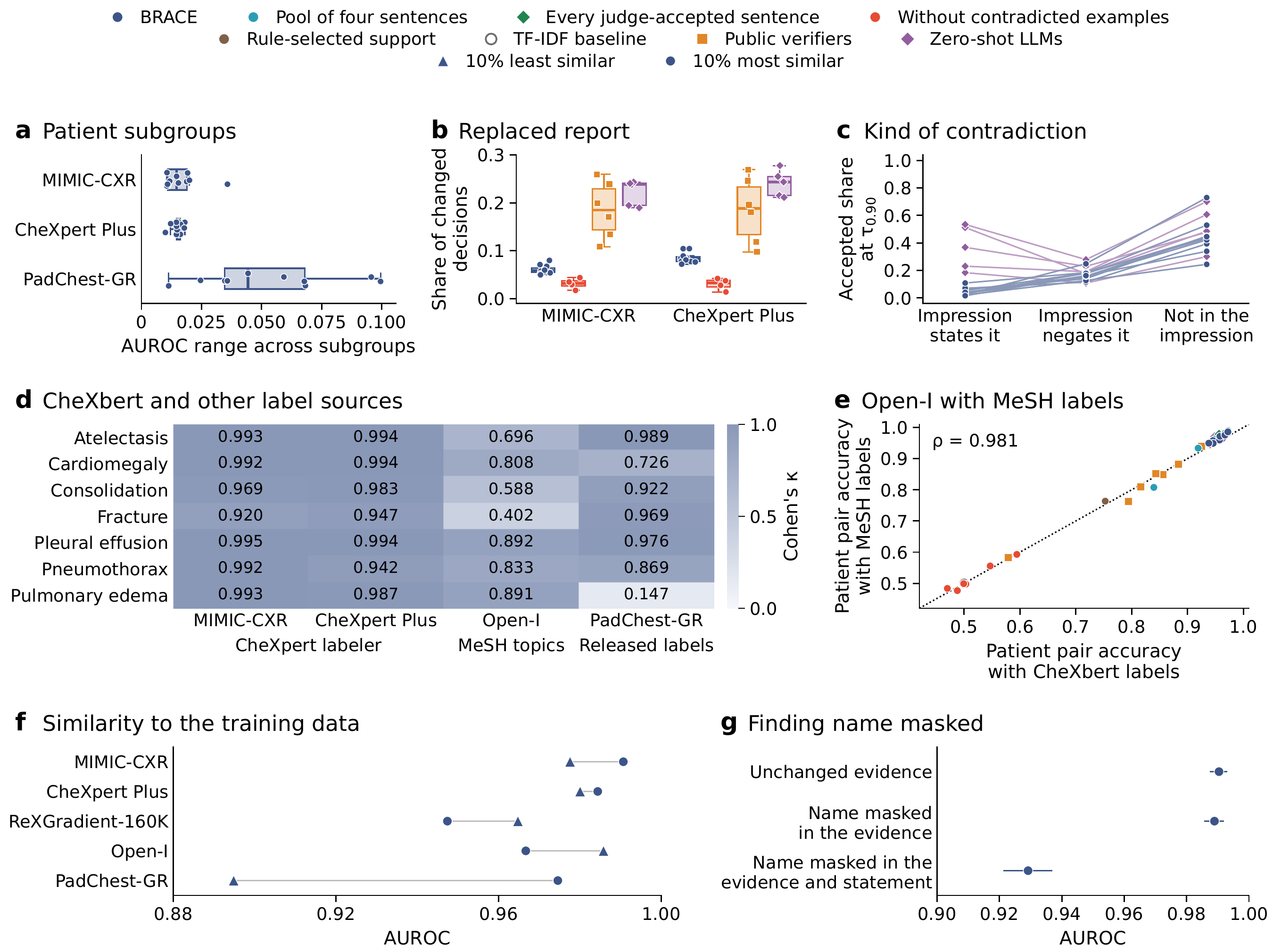}
\caption{Robustness of the verifiers and of the labels ($n = 43$ verifiers). Markers are bootstrap means over 1,000 resamples of patients, or of reports on Open-I, except the ranges of these means in \textbf{a}. Each box in \textbf{a} and \textbf{b} spans the 25th to the 75th percentile of a group, with a line at the median and whiskers to the most extreme value within 1.5 times the interquartile range. Error bars in \textbf{g} are 95\% CIs from the 2.5th to the 97.5th percentile. Counts per panel: 9, 10, and 6 subgroups with at least 30 patients (\textbf{a}); 4,356 and 6,077 examples (\textbf{b}); 596, 213, and 36 contradicted examples (\textbf{c}); 560 to 5,215 reports per finding and label source (\textbf{d}); 101,242 and 198,520 patient pairs (\textbf{e}); 195 to 925 examples per subset (\textbf{f}); and 4,356 examples (\textbf{g}). \textbf{a}, Range of the AUROC of each verifier across the subgroups by sex, age group, and race on MIMIC-CXR, CheXpert Plus~\cite{baker2024chexpertplus}, and PadChest-GR~\cite{castro2024padchestgr}. \textbf{b}, Share of decisions at $\tau_{0.90}$ that change when the report is replaced by that of another patient with the same finding state. GREEN and two verifiers without contradicted examples are left out of \textbf{b}, since their threshold accepts more than 0.95 of the supported examples. \textbf{c}, Accepted share at $\tau_{0.90}$ of the contradicted MIMIC-CXR examples whose impression states the finding called absent by the statement, negates the finding called present, or does not mention it. Lines join the three values of one verifier. \textbf{d}, Cohen's $\kappa$ of the CheXbert labels with the labels of the CheXpert labeler~\cite{chexpertmain}, the Medical Subject Headings (MeSH) topics, and the released labels. \textbf{e}, Patient pair accuracy on Open-I under the MeSH labels as a function of that under the CheXbert labels, with Spearman's $\rho$ across the verifiers. The dotted line is the identity. \textbf{f}, AUROC of ModernBERT-large on the 10\% least and most similar report findings of each set by unigram similarity. \textbf{g}, AUROC of ModernBERT-large on MIMIC-CXR with the finding name masked.}
\label{fig:robustness}
\end{figure*}

Within sex, age group, and race, the AUROC intervals of the patient subgroups with at least 30 patients overlap for 8 of the 11 verifiers trained with BRACE on MIMIC-CXR and 9 on CheXpert Plus (Fig.~\ref{fig:robustness}a and Supplementary Fig.~\ref{sfig:subgroups}). On PadChest-GR, the AUROC is 0.853 to 0.944 for the patients aged 40 to 59 and 0.950 to 0.978 for those aged 80 and over, with nonoverlapping intervals for 10 of the 11 verifiers.

Under the replaced report, i.e., the findings section of another patient with the same finding state, the share of changed decisions over all test examples of MIMIC-CXR and CheXpert Plus is 0.190 to 0.278 for the LLMs and 0.050 to 0.104 for the verifiers trained with BRACE (Fig.~\ref{fig:robustness}b). By the kind of contradiction on MIMIC-CXR, the accepted share of the verifiers trained with BRACE is 0.016 to 0.108 when the impression states the finding, 0.115 to 0.248 when it negates the finding, and 0.244 to 0.729 when it does not mention the finding (Fig.~\ref{fig:robustness}c).

Per finding, the CheXbert~\cite{smit2020chexbert} labels of the training reports and of four test sets have a Cohen's $\kappa$ of 0.920 to 0.995 with the labels of the CheXpert labeler~\cite{chexpertmain} released with MIMIC-CXR and CheXpert Plus and 0.402 to 0.892 with the Medical Subject Headings (MeSH) topics of Open-I (Fig.~\ref{fig:robustness}d). With the released labels of PadChest-GR, it is 0.726 to 0.989 for six findings and 0.147 for pulmonary edema. On Open-I, the ranking of the 43 verifiers by patient pair accuracy is nearly the same under CheXbert and MeSH labels ($\rho = 0.981$, $p_{\mathrm{FDR}} < 0.001$; Fig.~\ref{fig:robustness}e). The two judges gave the same answer for 0.865 of the 1,646 judged combinations of a sentence and a finding ($\kappa = 0.410$, $p_{\mathrm{FDR}} < 0.001$).

On the test report findings least and most similar to the training data~\cite{elangovan2021memorization}, the intervals of the AUROC of ModernBERT-large overlap on every test set except PadChest-GR (0.895 on the 10\% least similar and 0.975 on the 10\% most similar findings by unigrams; Fig.~\ref{fig:robustness}f and Supplementary Fig.~\ref{sfig:unseen}d,e). On MIMIC-CXR, its AUROC is 0.990 with unchanged evidence, 0.989 with the finding name masked in the evidence, and 0.929 with the name masked also in the statement (Fig.~\ref{fig:robustness}g and Supplementary Fig.~\ref{sfig:unseen}f).

\subsection*{Agreement of the automatic labels and the judges with radiologists}

\begin{figure*}[p]
\centering
\includegraphics[width=\textwidth]{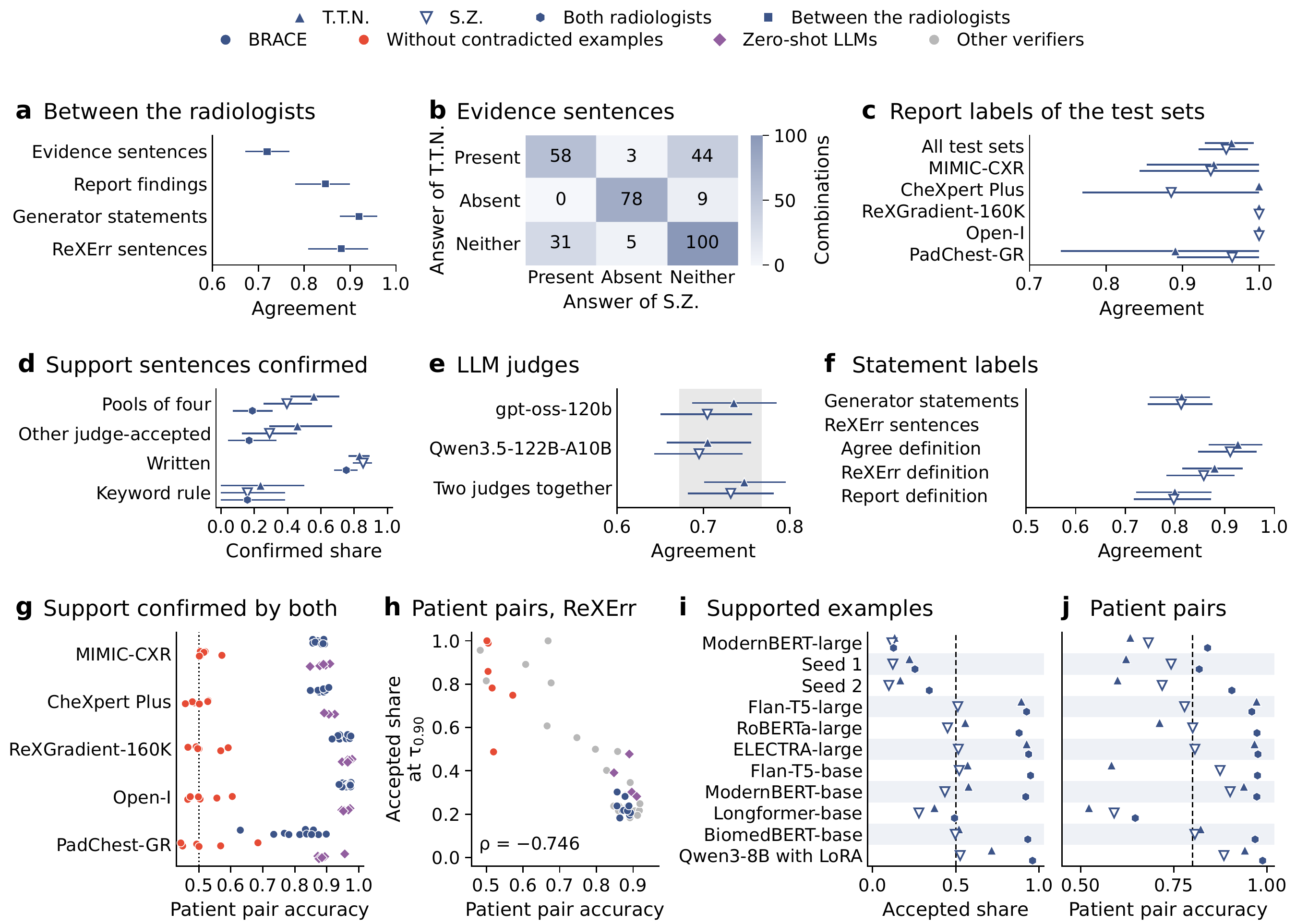}
\caption{Reader study and evaluation on confirmed support ($n = 728$ items per radiologist). Filled triangles mark T.T.N. and open triangles mark S.Z., or the support confirmed by that radiologist. Hexagons mark the support confirmed by both radiologists. Gray marks the 21 other verifiers in \textbf{h}. Markers are bootstrap means over 1,000 resamples of sentences, patients, or reports (\textbf{a}, \textbf{c}--\textbf{f}) or of patients or reports (\textbf{g}--\textbf{j}). Error bars in \textbf{a} and \textbf{c}--\textbf{f} are 95\% CIs from the 2.5th to the 97.5th percentile. Counts per panel: 99 to 328 items (\textbf{a}); 328 combinations of a sentence and a finding (\textbf{b}, \textbf{e}); 141 and 139 report findings (\textbf{c}); 13 to 150 combinations of a sentence and a finding (\textbf{d}); 79 to 150 statements (\textbf{f}); 11,890 to 502,124 patient pairs per test set (\textbf{g}); 92 contradicted ReXErr statements (\textbf{h}); and 177 to 501 supported examples and 2,132 to 28,948 patient pairs (\textbf{i}, \textbf{j}). \textbf{a}, Agreement between the two radiologists, i.e., the share of items with the same answer. \textbf{b}, Answers of the two radiologists on the state supported by each evidence sentence. \textbf{c}, Agreement of the report labels with each radiologist, with a finding not mentioned counted as absent. \textbf{d}, Share of the combinations in each group confirmed, i.e., read as supporting their assigned state. Other judge-accepted sentences are level 2 sentences outside the pools. \textbf{e}, Agreement of the LLM judges with each radiologist. The gray band is the 95\% CI of the first row of \textbf{a}. \textbf{f}, Agreement of the statement labels with each radiologist, with the generator statements under the report definition. \textbf{g}, Patient pair accuracy with the support confirmed by both radiologists. \textbf{h}, Accepted share at $\tau_{0.90}$ of the contradicted ReXErr statements with the training citation as a function of the patient pair accuracy on MIMIC-CXR, both with this support. Here, $\tau_{0.90}$ is recomputed for each support. \textbf{i},\textbf{j}, Accepted share at $\tau_{0.90}$ of the supported MIMIC-CXR test examples (\textbf{i}) and patient pair accuracy (\textbf{j}) with the confirmed sentences of the held-out articles. Dashed lines mark the criterion for accepting unseen evidence. The dotted line in \textbf{g} marks 0.5.}
\label{fig:reader}
\end{figure*}

Two board-certified radiologists, T.T.N. and S.Z., each read 328 combinations of an evidence sentence and a finding, 150 report findings, 150 generator statements, and 100 ReXErr sentences. They gave the same answer for 0.719 of the combinations (Cohen's $\kappa = 0.569$, $p_{\mathrm{FDR}} < 0.001$) and for 0.846 to 0.920 of the other items (Fig.~\ref{fig:reader}a and Supplementary Table~\ref{stab:reader}). In 89 of the 92 combinations with different answers, only one radiologist assigned a state (Fig.~\ref{fig:reader}b). The report labels of BRACE agree with each radiologist almost as closely as the radiologists agree with each other. The agreement is 0.964 with T.T.N. ($\kappa = 0.927$), 0.957 with S.Z. ($\kappa = 0.910$), and 0.977 between the radiologists ($\kappa = 0.952$), each with $p_{\mathrm{FDR}} < 0.001$ (Fig.~\ref{fig:reader}c).

A support sentence is confirmed when a radiologist reads it as supporting its assigned state. Both radiologists confirmed most of the 150 written sentences, with a confirmed share of 0.753, and fewer than a fifth of the 43 combinations of a Radiopaedia sentence and a finding in the pools of four, with 0.189 (Fig.~\ref{fig:reader}d). On the 328 combinations, the LLM judges agree with each radiologist about as often as the radiologists agree with each other, with overlapping intervals (0.695 to 0.735 for each judge and 0.732 to 0.747 for both judges together; Fig.~\ref{fig:reader}e). Under the report definition, the labels of the generator statements agree with each radiologist for 0.812 to 0.813 of the statements (Fig.~\ref{fig:reader}f).

Both radiologists confirmed support sentences for 5 of the 14 finding states. We repeated the evaluation on this support. There, every verifier trained with BRACE has a higher patient pair accuracy than every verifier trained without contradicted examples on every test set except PadChest-GR (Fig.~\ref{fig:reader}g). Across the 43 verifiers, the patient pair accuracy on MIMIC-CXR remains rank-correlated with the accepted share of the contradicted ReXErr statements ($\rho = -0.746$, $p_{\mathrm{FDR}} < 0.001$; Fig.~\ref{fig:reader}h). Both radiologists confirmed 2 of the 12 support sentences of the held-out articles. With these sentences, every verifier trained with BRACE except Longformer-base and ModernBERT-large with its two seeds meets the criterion for unseen evidence (Fig.~\ref{fig:reader}i,j). In the template form, the support of all 606 ReXErr statements with a supporting retrieved citation comes from two other sentences of the held-out articles. Neither of them was confirmed by both radiologists. With the threshold calibrated on the confirmed support, the verifiers trained with BRACE still reject most of the 1,444 statements with an agreeing report and an unsupporting citation, with an accepted share of 0.000 to 0.387. The LLMs accept most of them (0.587 to 0.904).

\subsection*{Generalization of BRACE to a second domain}

BRACE extends beyond clinical applications. We instantiated it for food allergens in recipes, with Wikipedia sentences as the evidence. The directions of a recipe take the place of the report, and recipe pairs take the place of patient pairs. We trained ModernBERT-large on these examples. This food verifier uses the recipe and the evidence, with a recipe pair accuracy of 0.792 $\pm$ 0.025 [0.744, 0.839] and an evidence pair accuracy of 1.000 (Fig.~\ref{fig:panel}b,c). Its AUROC of 0.934 is below the AUROC of 0.968 attainable without the recipe. At the $\tau_{0.90}$ of the food test set, the food verifier rejects about half of the examples contradicted by the recipe, with an accepted share of 0.493, whereas the TF-IDF baseline accepts most of them (0.898; Fig.~\ref{fig:panel}d). The two verifiers accept none of the examples without supporting evidence and pass the evidence tests (Fig.~\ref{fig:panel}e,f).

\section*{Discussion}

AI systems that draft chest radiograph reports or answer radiology questions make statements about individual patients and cite reference text for them. We asked whether training on examples built from public radiology reports and a reference text yields verifiers that check such a statement against both the patient's report and the cited evidence. Our main finding is that verifiers trained with BRACE use both the report and the cited evidence on five test sets from the United States and Spain. They reject most ReXErr statements contradicted by the report, with an accepted share of 0.239 on average. Verifiers trained without the contradicted examples pass the state-of-the-art evidence tests, yet accept most of these statements. With citations retrieved from held-out Radiopaedia articles, the accepted share of the ReXErr statements with an agreeing report and an unsupporting citation averages 0.169 over the verifiers trained with BRACE and 0.830 over five zero-shot LLMs. In a second domain, a verifier trained with BRACE for food allergens in recipes uses the recipe and the cited evidence.

Classifiers of chest radiographs have performed worse on data from other hospitals and patient populations than on data like their training data~\cite{zech2018variable,tayebi2023enhancing,lotfinia2025boosting}. The verifiers trained with BRACE also have lower AUROCs on the four test sets from other hospitals than on MIMIC-CXR, while they still use the report and the cited evidence there. Their patient pair accuracy is lowest on PadChest-GR, whose reports are English translations of sentences labeled by radiologists. We cannot tell whether this form of the reports or this source of the labels lowers the accuracy.

Trained on documents and on versions revised by hand to flip the label, models relied less on spurious features~\cite{kaushik2020learning}. Within a patient pair, changing only the report flips the label. FEVER-trained models lost accuracy on an evaluation set built without the cues that let the statement alone predict the label~\cite{schuster2019towards}. In a patient pair, such cues are absent, since its two examples share the statement and the evidence. A dataset can contain shortcuts although a partial-input baseline performs poorly on it~\cite{feng2019misleading}. The AUROC attainable without the report needs no trained baseline. It is 1.000 on each training set without contradicted examples and 0.995 for BRACE, whose verifiers use the report. Therefore, before training, we can use this AUROC only to find a training set in which no label needs the report. After training, the patient pair accuracy measures whether a verifier uses the report.

A verifier can have a high AUROC with unseen support sentences and still reject them. Trained on the pool of four sentences, ModernBERT-large has an AUROC of 0.898 with support sentences from the held-out articles. At $\tau_{0.90}$, it accepts none of the supported examples that cite these sentences. The AUROC depends on the order of the scores and not on their level. In open-domain question answering, models answered test questions without overlap with the training data much less accurately than questions with overlap~\cite{lewis2021question}. Among the verifiers trained with BRACE, Longformer-base and ModernBERT-large with its two seeds have the four lowest accepted shares of the supported examples with support sentences from the held-out articles. They are also the only verifiers trained with BRACE that reject most AI-written statements with report agreement and citation support.

Public verifiers such as MiniCheck~\cite{tang2024minicheck} and AlignScore~\cite{zha2023alignscore} check a statement against one grounding document. In our setting, the report and the cited evidence form this document, in which the report alone can support the statement. The prompt of the zero-shot LLMs defines a statement as supported only if the report agrees with it and the cited evidence supports it. The LLMs nevertheless use the report more than the cited evidence. MiniCheck models trained on synthetic errors written by GPT-4 have matched GPT-4 on grounding documents~\cite{tang2024minicheck}. Qwen3-8B finetuned with BRACE also has a higher AUROC with retrieved citations than every zero-shot LLM, including four LLMs with 27 to 284 billion parameters. Without the contradicted examples in its training data, the same model makes little use of the report. The two finetuned versions of this model were trained with the same recipe on different training examples. Of the responses of LLMs to medical questions, 50\% to 90\% were not fully supported by their cited sources~\cite{wu2025automated}. A verifier that checks only the report would pass such citations on to the radiologist. With retrieved citations, however, 1,444 of the 1,845 ReXErr statements with an agreeing report have no supporting sentence among their two retrieved sentences. Since the verifiers trained with BRACE reject most of these statements, the retrieval would limit how many correct statements reach the radiologist with a supporting citation.

Across many models, accuracy on a test set has been strongly correlated with accuracy under a distribution shift~\cite{miller2021accuracy}. A verifier must use the report to distinguish two examples with the same statement and citation but opposite report states. The AUROC does not measure whether a verifier uses the report. Therefore, a verifier with a higher AUROC can accept more contradicted statements. The TF-IDF baseline has a higher AUROC than every zero-shot LLM. Yet it accepts more of the contradicted ReXErr statements with the training citation than any LLM (0.653). Its patient pair accuracy of 0.500 is below that of every LLM.

The replaced report keeps the state and the CheXbert class of the finding. Over all test examples of MIMIC-CXR and CheXpert Plus, every zero-shot LLM changes more decisions under the replaced report than every verifier trained with BRACE. The accuracy of LLMs on arithmetic word problems has also decreased when irrelevant sentences were added~\cite{shi2023large}. For each verifier trained with BRACE, the mean accepted share of the contradicted examples is lowest when the impression states the finding, higher when it negates the finding, and highest when it does not mention the finding. For 5 of the 11 verifiers, the intervals of the three kinds do not overlap. Transformers have struggled with examples of natural language inference in which negation is important~\cite{hossain2020analysis}. When the impression does not mention the finding, the statement is contradicted only because we map a finding not mentioned to absent, as in the labeling convention of the chest radiograph datasets~\cite{chexpertmain,johnson2019mimic}.

A strong LLM judge has agreed with human preferences as often as two humans agreed with each other~\cite{zheng2023judging}. In our reader study, the automatic report labels and the LLM judges agree with each radiologist about as often as the radiologists agree with each other. The radiologists agree less on whether a sentence supports a finding state ($\kappa = 0.569$) than on the state of a finding in a report ($\kappa = 0.952$). Between the radiologists and between a judge and a radiologist, nearly every disagreement on a sentence concerns whether it supports a state, and not which state it supports. On the support confirmed by both radiologists, the verifiers trained with BRACE still use the report. With their thresholds recalibrated on this support, they reject most statements labeled as lacking citation support.

In the FEVER~\cite{thorne2018fever} and SciFact~\cite{wadden2020fact} datasets, the label depends only on the statement and the evidence. In our food domain, the label also depends on the recipe, as it depends on the report in radiology. From its recipe pairs, we can tell that the food verifier uses the recipe, although its AUROC is below the AUROC attainable without the recipe.

This study has several limitations. First, all training reports come from one hospital (MIMIC-CXR), since we kept the four other radiology datasets as external test sets. However, we applied BRACE to a completely unrelated domain, food allergens in recipes. As on chest radiographs, the food verifier uses both the recipe and the cited evidence. Future work should train verifiers on data from other hospitals, imaging modalities, and domains to test whether BRACE transfers. Second, every statement names one of seven findings as present or absent, since CheXbert labels these states in the report. Future work should extend BRACE with labels for the location, severity, and change of a finding. Third, apart from the written sentences, all support sentences come from Radiopaedia. Since both judges assign the same state to 94 of its 1,646 combinations of a sentence and a finding, its held-out articles have 12 support sentences for 6 of the 14 finding states. On these sentences, 4 of the 11 verifiers trained with BRACE meet the criterion for unseen evidence. However, 7 of the 11 meet it on the 2 of these sentences confirmed by both radiologists. All 11 meet it on the written held-out sentences of all 14 finding states. Future work should test unseen evidence in every finding state with radiologist-confirmed support sentences from a second reference text. Fourth, 8 of the 11 verifiers trained with BRACE accept most supported examples when no evidence is given, since every training example has evidence. However, all 11 reject nearly all examples without supporting evidence. Future work should add training examples without evidence and with a negative label. Fifth, no AI system of this study cited sources for its statements. However, we scored each statement with up to three citations. With citations retrieved by BM25 from held-out Radiopaedia articles, the verifiers trained with BRACE still reject most statements without citation support. Future work should have radiologists judge the sources cited by an AI system and test the verifier on its statements. Sixth, each zero-shot LLM received one fixed prompt without example statements and answered without reasoning first, since its score is taken at the first answer token. We used the same prompt for every LLM, so that no LLM was tuned on the test sets. However, the prompt already required the cited evidence to support the statement. Future work should test whether prompts with example statements or with reasoning before the answer make the LLMs reject more statements without citation support.

Training on examples built from the chest radiograph reports of one hospital and from one reference text yielded verifiers that use both the report and the cited evidence on five test sets from two countries. Unlike the zero-shot LLMs, these verifiers reject most AI-written statements with an agreeing report and an unsupporting retrieved citation. BRACE also extends beyond clinical applications, to food allergens in recipes. The patient pair accuracy needs only test examples built with BRACE and the scores of a verifier. With a retrieval step that returns supporting sentences more often than our BM25 retrieval, verifiers trained with BRACE could check AI-written statements and their citations before clinicians read them.

\section*{Methods}

\subsection*{Ethics statement}

This study was conducted in accordance with the relevant national and international guidelines and regulations. All data of this study are public. Each dataset was used under the conditions of its release. Radiopaedia approved the use of its content for noncommercial machine learning research in this study (permission number 1163-5571). Sentences derived from Radiopaedia were not redistributed where the project-specific agreement did not permit public sharing of derived dataset content. The approval and the consent procedure reported for each dataset are given in Supplementary Note~\ref{snote:data}. No new patient data were generated and no re-identification was attempted. Therefore, this secondary analysis required no further ethics approval or informed consent.

\subsection*{Datasets}

An example consists of a report, at most two evidence sentences, and a statement about one of seven findings, namely atelectasis, cardiomegaly, consolidation, fracture, pleural effusion, pneumothorax, and pulmonary edema. The report is the findings section of a radiology report. The reports come from five public chest radiograph datasets. The training and validation sets and one test set come from MIMIC-CXR~\cite{johnson2019mimic} and have no patient in common. The four other test sets come from CheXpert Plus~\cite{baker2024chexpertplus}, ReXGradient-160K~\cite{zhang2025rexgradient}, and Open-I~\cite{demnerfushman2016preparing} in the United States and from PadChest-GR~\cite{castro2024padchestgr} in Spain. The evidence sentences come from a snapshot of the articles of Radiopaedia~\cite{gaillard2011radiopaedia}, divided into training articles and held-out articles. The AI-written statements come from ReXErr-v1~\cite{rao2024rexerr} and from the reports of six report generators for the Open-I radiographs. ReXErr-v1 contains the sentences of MIMIC-CXR reports with and without an error injected by an LLM. The verifiers are also evaluated as metrics of whole generated reports on ReXVal~\cite{yu2023evaluating} and RadEvalX~\cite{calamida2023radiology}, in which radiologists counted the errors of each report. In the food domain, the reports are the directions of recipes from RecipeNLG~\cite{bien2020recipenlg}, the findings are six food allergens, and the evidence sentences come from Wikipedia articles. Table~\ref{tab:datasets} lists every set with its size. Supplementary Tables~\ref{stab:examples} and~\ref{stab:examples_other} list the examples per finding. The source, the selection, and the labels of each dataset are given in Supplementary Note~\ref{snote:data}.

A report finding is a study together with one of the seven findings and its state in the report, present or absent. On MIMIC-CXR, CheXpert Plus, ReXGradient-160K, and Open-I, we labeled the impression section of each report with CheXbert~\cite{smit2020chexbert}, a BERT labeler of radiology reports (Fig.~\ref{fig:framework}a). CheXbert assigns each finding the class positive, negative, uncertain, or blank, where blank means that the impression does not mention the finding. As in the labeling convention of the chest radiograph datasets~\cite{chexpertmain,johnson2019mimic}, we map the class positive to present, map the classes negative and blank to absent, and drop the class uncertain. On PadChest-GR, the state comes from the finding labels assigned by radiologists to the sentences of each report. For a second Open-I test set, the state comes from the MeSH topics assigned to each report by hand. Every present finding of a study enters the set. Per split and finding, the absent findings are drawn at random, first from the reports that negate the finding and then from the reports that do not mention it. On the second Open-I test set, they are drawn from all reports whose MeSH topics do not name the finding. The random generator of each draw is seeded by 13, the dataset, the split, and the finding. On MIMIC-CXR and CheXpert Plus, a preset number of absent findings is drawn per split and finding (Table~\ref{tab:datasets} and Supplementary Note~\ref{snote:data}). On ReXGradient-160K, Open-I, and PadChest-GR, as many absent findings are drawn per finding as there are present findings.

\subsection*{BRACE}

BRACE builds the training examples of a grounding verifier from two inputs (Fig.~\ref{fig:framework}), namely reports with an automatic label of the state of each finding and a reference text split into sentences. Two LLM judges select the support sentences of each finding state from the reference text. An LLM writer adds further support and unsupporting sentences, each accepted or rejected by the same judges. For every report finding, BRACE then builds a supported example, a contradicted example, and two examples without supporting evidence. The definitions of these example types and of the checks of a verifier do not depend on the domain. In the food domain, BRACE builds the same four types of examples, with support sentences selected by a keyword rule in place of the judges and without written sentences (Supplementary Note~\ref{snote:data}).

A statement $y$ has the form ``$f$ is $s$'' for a finding $f$ and a state $s$, as in ``pleural effusion is present''. The support pool $P_f^{s}$ of a finding state $(f, s)$ is the set of reference sentences selected as support of the statement ``$f$ is $s$''. For a report $c$ in which finding $f$ has the state $g(c, f)$, evidence $e$, and a statement $y$ about $f$ in state $s$, the label is the product of a report factor $\lambda$ and an evidence factor $\sigma$,
\begin{equation}\label{eq:label}
\ell(c, e, y) = \lambda(c, y)\,\sigma(e, y), \qquad \lambda(c, y) = \mathbf{1}\big[g(c, f) = s\big], \qquad \sigma(e, y) = \max_{u \in e} \mathbf{1}\big[u \in P_f^{s}\big].
\end{equation}
For a report finding with state $s$ and opposite state $\bar{s}$, we build up to four examples (Fig.~\ref{fig:framework}c). A supported example states $s$ with evidence from $P_f^{s}$, so $\lambda = \sigma = 1$ and $\ell = 1$. A contradicted example states $\bar{s}$ with evidence from $P_f^{\bar{s}}$, so $\lambda = 0$, $\sigma = 1$, and $\ell = 0$. Two examples without supporting evidence state present and absent with unsupporting sentences, so $\sigma = 0$ and $\ell = 0$. A supported or contradicted example exists only where its support pool is not empty.

A Radiopaedia sentence names a finding when it contains a term of the finding in the evidence lexicon. It is a chest sentence when it passes the cleanup filters and contains a chest term (Supplementary Note~\ref{snote:data}). The two LLM judges are gpt-oss-120b~\cite{openai2025gptoss} and Qwen3.5-122B-A10B~\cite{qwen2026qwen35} with GPTQ 4-bit weights. The two judges decode at temperature 0. A judge receives one sentence and one finding in the messages of Supplementary Note~\ref{snote:inputs} (Fig.~\ref{fig:framework}b). It answers whether the sentence concerns the finding and whether the sentence points toward the state present, toward the state absent, or toward neither. We assign the sentence to the state toward which it points when it concerns the finding, and to no state otherwise. The judges answered for all 1,646 combinations of a sentence and a finding named in the sentence. Of these combinations, 1,325 come from the training articles, and 321 come from the held-out articles. Both judges assign the same state to 94 combinations of 93 sentences. The candidates are the 67 of these combinations whose sentence is a chest sentence, 55 in the training articles and 12 in the held-out articles. The support pool of a finding state and an article set contains the first four candidates of that state in snapshot order, i.e., the order in which the snapshot lists its sentences, article by article. The training pools contain 31 combinations of 30 distinct sentences in 12 of the 14 finding states. The held-out pools contain 12 sentences in 6 states (Supplementary Fig.~\ref{sfig:support}a).

The hard unsupporting sentences of a finding $f$ are the chest sentences that name $f$ and are assigned to no state of $f$ by either judge (Supplementary Fig.~\ref{sfig:support}f). There are 502 such combinations of 481 sentences with a finding in the training articles and 121 combinations of 115 sentences in the held-out articles. Before drawing evidence, we remove support sentences of every finding from the hard unsupporting sentences of the same article set. The easy unsupporting sentences are the chest sentences that name none of the seven findings, 2,401 in the training articles and 605 in the held-out articles. A supported or contradicted example takes up to two sentences drawn at random from its support pool, kept in snapshot order. An example without supporting evidence takes two unsupporting sentences, drawn from the hard sentences first for the statement present and from the easy sentences first for the statement absent, and completed from the other set. For each dataset and article set, one random generator draws the evidence in the order of the report findings and is seeded by 13, the dataset, and the article set. At levels 2 and 3, the seed also contains the name of the training set, and the test set with the written held-out sentences is drawn with a seed of 13, the dataset, and the name of this set. The evidence of each example is drawn once and used in training and in evaluation. The construction is written in pseudocode in Supplementary Algorithm~\ref{salg:construction}.

We built the training evidence at three levels. Level 1 uses the pools of four sentences. Level 2 keeps every candidate of the training articles, 55 combinations of 54 sentences in the same 12 states. Level 3, i.e., BRACE, adds sentences written by an LLM in the style of the reference text. The writer is Qwen3.5-35B-A3B~\cite{qwen2026qwen35} with GPTQ 4-bit weights, sampled at a temperature of 0.8 with a top-$p$ of 0.95. Its system message and its requests are given in Supplementary Note~\ref{snote:inputs}. For each of the 14 finding states, the writer was asked twice for 30 sentences that describe the state, with up to 5 judge-accepted sentences of the state as examples of the style. It was also asked twice per finding for 30 sentences that mention the finding without saying whether it is present or absent, and once for 3 rewrites of each of the 55 training candidates. In all, the writer was called 97 times. A returned line counted as a written candidate when it had at least six words and ended with a period. Written candidates that repeat a style example or an earlier written candidate of the same finding were dropped. Of the 1,439 written candidates that remained, 844 are support sentences, 430 are unsupporting sentences, and 165 are rewrites. The judges answered for every written candidate with the finding of its request (Supplementary Fig.~\ref{sfig:support}c--e,g). Both judges assign 484 of the support sentences and 77 of the rewrites to the state named in the request. They assign 297 of the unsupporting sentences to no state. A leakage filter then dropped a written sentence when its text equaled a snapshot sentence, when the Jaccard similarity of its word set with a chest sentence of the held-out articles was at least 0.6, or when that similarity with a written sentence kept before it was at least 0.8. The filter dropped 3 near duplicates, and 855 written sentences remained. For each finding state with at least five written support sentences, a random fifth of them was held out from training, with the number rounded to the nearest integer and at least two. The random generator of this draw is seeded by 13, the finding, and the state. The draw left 446 training and 112 held-out written support sentences. Each of the two sets has sentences of all 14 states (Supplementary Fig.~\ref{sfig:support}b). At level 3, the written unsupporting sentences are added to the hard unsupporting sentences of their finding. With a probability of 30\%, the evidence of each supported and contradicted example of the training and validation splits of level 3 is replaced by its first support sentence and one hard unsupporting sentence of the finding drawn at random, in random order (Table~\ref{tab:datasets}). The label of an example with such mixed evidence is unchanged under Eq.~\ref{eq:label}. At level 3, every report finding has a supported and a contradicted example, since every finding state has a support pool.

Four further training sets contain no contradicted examples. Three of them are variants of level 1 that keep its supported examples. One variant also keeps the examples without supporting evidence of level 1. In the easy-evidence variant, the examples without supporting evidence take two easy unsupporting sentences. In the random-evidence variant, they take two sentences drawn at random from the sentences of the training articles that pass the cleanup filters and are assigned to no state of the finding by either judge. The fourth set is BRACE without its contradicted examples. In each of these sets, no combination of evidence and statement occurs with both labels. So the evidence and the statement determine every label. In the comparison set, i.e., level 1 with rule-selected support, the support pools are the 13 combinations of 12 sentences in 8 finding states selected as support by a keyword rule (Supplementary Note~\ref{snote:data}). These sentences are removed from the hard unsupporting sentences of the comparison set. The test examples of the chest radiograph test sets are built twice, once with the training pools and once with the held-out pools. We call the evidence from the training pools unchanged evidence. The MIMIC-CXR test set is built a third time with the written held-out support sentences as its support pools and with the hard and easy unsupporting sentences of the held-out articles.

\subsection*{Verifiers}

The radiology domain has 43 verifiers (Supplementary Table~\ref{stab:models}). We trained 30 of them on the constructed examples, namely 11 on each of levels 1 and 3, 1 on level 2, 6 on the four training sets without contradicted examples, and 1 on the comparison set. The 11 verifiers trained on level 3 are the verifiers trained with BRACE. The other 13 verifiers are the TF-IDF baseline, 7 public verifiers, and 5 zero-shot LLMs. In the food domain, the food verifier is ModernBERT-large trained on the food examples, and the TF-IDF baseline is fitted on the same examples.

The eight transformer verifiers are sequence classifiers with two output classes, initialized from ModernBERT-large~\cite{warner2024modernbert}, Flan-T5-large~\cite{falnt5}, RoBERTa-large~\cite{liu2019roberta}, ELECTRA-large~\cite{clark2020electra}, Flan-T5-base~\cite{falnt5}, ModernBERT-base~\cite{warner2024modernbert}, Longformer-base~\cite{beltagy2020longformer}, and BiomedBERT-base~\cite{biomedbert}. A transformer verifier receives the report, the evidence sentences joined by a blank line and cut at 12,000 characters, and the statement in the fixed template of Supplementary Note~\ref{snote:inputs} (Fig.~\ref{fig:framework}d). When an input exceeds the input length of the model, words are removed from the end of its report. The evidence and the statement are kept whole. Each transformer verifier was finetuned on the training split of its training set with a cross-entropy loss that weights the supported class by the ratio of the training examples with $\ell = 0$ to those with $\ell = 1$. Each run trained for 15 epochs with AdamW~\cite{loshchilov2019decoupled} at a constant learning rate of $8 \times 10^{-6}$ and a weight decay of 0.01, shuffled batches of 4 examples with the last incomplete batch dropped, gradient clipping at a norm of 1.0, and bfloat16 autocasting over float32 weights. The validation AUROC was computed under bfloat16 autocasting. The checkpoint of the epoch with the highest validation AUROC was evaluated in float32. The seed of a run sets the data order, the dropout, and the initialization of the classification head. ELECTRA-large trained with BRACE used a learning rate of $4 \times 10^{-6}$. Every other transformer run used these settings. Supplementary Table~\ref{stab:models} lists the parameters, the input lengths, the training sets, and the seeds of the runs.

The TF-IDF baseline is a logistic regression on TF-IDF features of the statement followed by the evidence. As a partial-input baseline~\cite{poliak2018hypothesis,feng2019misleading}, it does not receive the report. Its features are the lowercase unigrams and bigrams that occur in at least two training examples, weighted with sublinear term frequencies. The regression uses balanced class weights and the liblinear solver with at most 5,000 iterations. We chose its inverse regularization strength $C$ from 0.01, 0.1, 1, 10, and 100 by the validation AUROC, which is 0.967 for radiology and 0.838 for food at the chosen values of Supplementary Table~\ref{stab:models}. The features and the regression are implemented in scikit-learn~\cite{pedregosa2011scikit}.

The public verifiers are MiniCheck-Flan-T5-Large~\cite{tang2024minicheck}, Bespoke-MiniCheck-7B~\cite{bespokelabs2024minicheck}, AlignScore-large~\cite{zha2023alignscore}, HHEM-2.1-Open~\cite{li2024hhem}, LettuceDetect~\cite{kovacs2025lettucedetect}, Lynx-8B~\cite{ravi2024lynx}, and GREEN~\cite{ostmeier2024green}. We evaluated their released checkpoints without tuning, with one fixed mapping of the report, the evidence, and the statement to the inputs of each verifier. Supplementary Table~\ref{stab:models} lists the mapping and the score of each verifier, and their inputs are given in Supplementary Note~\ref{snote:inputs}. GREEN retains the first 300 words of each reference and candidate input and decodes greedily in float16 with a maximum length of 2,048 tokens, Lynx-8B decodes at temperature 0 with at most 600 new tokens, and Bespoke-MiniCheck-7B was loaded in bfloat16.

The zero-shot LLM verifiers are DeepSeek-V4-Flash~\cite{deepseek2026v4}, Qwen3.8-Flash-Next~\cite{qwen2026qwen38flashnext}, Mistral-Small-4-119B~\cite{mistral2026small4}, MedGemma-27B~\cite{sellergren2025medgemma}, and Qwen3-8B~\cite{yang2025qwen3}, with the sizes and settings of Supplementary Table~\ref{stab:models}. Each LLM receives the report, the evidence, and the statement in the prompt of Supplementary Note~\ref{snote:inputs}, which asks whether the statement is supported and requests an answer of Yes or No. Each LLM decoded at temperature 0 for at most 4 new tokens and returned the 20 most probable tokens at every position. The score is $P(\mathrm{Yes}) / (P(\mathrm{Yes}) + P(\mathrm{No}))$ at the first generated token that contains a letter or a digit. Each of the two probabilities is the sum over the returned tokens whose lowercased text without surrounding spaces is yes or no. The score is 0.5 when neither answer is among the returned tokens. In every answer of these verifiers, Yes or No was among the returned tokens.

Qwen3-8B was also finetuned with LoRA~\cite{hu2022lora} at levels 1 and 3, with and without the contradicted examples. It receives the prompt of the zero-shot LLM verifiers in its chat template with thinking disabled and is trained to produce Yes as its first answer token for a supported example and No otherwise. The loss is the cross-entropy over the two logits of Yes and No at the answer position, with the class weights of the transformer verifiers. LoRA adapters of rank 16 with a LoRA $\alpha$ of 32 and a dropout of 0.05 were trained on the query, key, value, output, gate, up, and down projections of every layer. Training used AdamW at a constant learning rate of $10^{-4}$ and no weight decay, an effective batch of 16 examples, gradient clipping at 1.0, and gradient checkpointing. The base model was frozen in bfloat16. The adapters were trained in float32 under bfloat16 autocasting. Each run trained for three epochs. The validation AUROC was computed from the same two logits. The adapter of the epoch with the highest validation AUROC was evaluated. The finetuned model is scored as the zero-shot LLM verifiers are. Every answer of it had Yes or No among the returned tokens. No model is both a judge and a verifier or both a generator and a verifier. The writer has no other role.

\subsection*{Standard validation, evidence tests, and patient-level checks}

In the standard validation, we report the AUROC of every verifier on the test examples of every set with unchanged evidence. The state-of-the-art evidence tests~\cite{deyoung2020eraser,joren2025sufficient,ming2025faitheval} change only the evidence of the unchanged test examples. Under replaced evidence, every example keeps its label and takes the evidence of another example of the same test set with a different evidence text, through a derangement drawn with seed 13. Under removed evidence, the examples keep their labels and have no evidence. We report the AUROC under the evidence tests on the MIMIC-CXR and the food test sets. A verifier passes the evidence tests when both its AUROC with replaced evidence and its AUROC with removed evidence are below its AUROC with unchanged evidence.

A patient pair is a supported and a contradicted example with the same evidence and statement. Its two examples differ only in the report. In the food domain, the two examples of a recipe pair differ only in the recipe. With $\mathcal{S}$ and $\mathcal{K}$ the supported and the contradicted examples of a test set and $z_i$ the score of example $i$, the patient pairs and their accuracy are
\begin{equation}\label{eq:pairs}
\Pi = \big\{(i, j) \in \mathcal{S} \times \mathcal{K} : e_i = e_j,\ y_i = y_j\big\}, \qquad \mathrm{PA} = \frac{1}{|\Pi|} \sum_{(i, j) \in \Pi} \Big( \mathbf{1}[z_i > z_j] + \tfrac{1}{2}\,\mathbf{1}[z_i = z_j] \Big),
\end{equation}
where $e_i = e_j$ means that the two verifier inputs contain the same evidence text. When the score of a verifier is a function $h$ of the evidence and the statement alone, both examples of every pair receive the same score, so
\begin{equation}\label{eq:half}
z_i = h(e_i, y_i) \ \text{for every example } i \quad \Longrightarrow \quad \mathrm{PA} = \tfrac{1}{2}
\end{equation}
on every test set (Supplementary Note~\ref{snote:proofs}). An evidence pair is a supported example and the example without supporting evidence of the same report finding whose statement has the same state. The evidence pair accuracy is computed by Eq.~\ref{eq:pairs} with these pairs. Patient pairs are reported with unchanged evidence, with evidence from the held-out articles, and with the written held-out sentences. The evidence pair accuracy is reported with unchanged evidence. The patient pair accuracy is computed over all pairs of a test set and within the pairs of each finding.

For a share $q$, the threshold $\tau_q$ of a verifier and the share $a_q(\mathcal{X})$ of a set $\mathcal{X}$ of examples accepted at this threshold are
\begin{equation}\label{eq:operating}
\tau_q = \max\Big\{ t : \frac{1}{|\mathcal{S}|}\sum_{i \in \mathcal{S}} \mathbf{1}[z_i \ge t] \ge q \Big\}, \qquad a_q(\mathcal{X}) = \frac{1}{|\mathcal{X}|}\sum_{j \in \mathcal{X}} \mathbf{1}\big[z_j \ge \tau_q\big],
\end{equation}
for $q = 0.90$ and $q = 0.95$, with $\mathcal{S}$ the supported MIMIC-CXR test examples with unchanged evidence. The set $\mathcal{X}$ is the contradicted examples, the examples without supporting evidence, or the supported examples of any test set and condition. In the food domain, the threshold is set on the supported examples of the food test set. A verifier with tied scores at $\tau_q$ accepts more than the share $q$ of the supported examples. A verifier whose accepted share of the supported MIMIC-CXR examples differs from $q$ by more than 0.05 is compared with the other verifiers by its AUROC alone. The correlations across the verifiers and the report error score include every verifier. At $q = 0.90$, this applies to ModernBERT-large trained on the easy-evidence variant, to Qwen3-8B with LoRA trained on BRACE without its contradicted examples, and to GREEN, whose score takes few distinct values. The threshold $\tau_{0.90}$ is the operating point of every analysis. A verifier accepts unseen evidence when its accepted share at $\tau_{0.90}$ of the supported MIMIC-CXR examples with the support sentences of the held-out articles is at least 0.50 and its patient pair accuracy with these sentences is at least 0.80. This criterion was fixed before the verifiers were trained with BRACE. The same two thresholds apply to the written held-out sentences.

An input channel $x$ of an example is a part of its input, such as its evidence and statement $(e, y)$. A verifier that receives only $x$ assigns the same score to all examples of a set that share $x$ (Fig.~\ref{fig:framework}e). The AUROC attainable from $x$ is obtained by scoring each example with the share of supported examples among the examples with the same $x$,
\begin{equation}\label{eq:bound}
B_x = \mathrm{AUROC}\big(\{(\ell_i, \hat{p}_x(i))\}_i\big), \qquad \hat{p}_x(i) = \frac{\sum_{j : x_j = x_i} \ell_j}{\big|\{j : x_j = x_i\}\big|},
\end{equation}
and no verifier that receives only $x$ has a higher AUROC on the same examples (Supplementary Note~\ref{snote:proofs}). We report $B_{(e, y)}$, the AUROC attainable without the report, on the five test sets and the food test set with unchanged evidence and on the training splits of the three levels and of the four sets without contradicted examples. A verifier whose AUROC exceeds $B_{(e, y)}$ on the same examples uses the report. The computation of the patient pair accuracy, the accepted share of Eq.~\ref{eq:operating}, and the attainable AUROC from the examples and scores of one set is written in pseudocode in Supplementary Algorithm~\ref{salg:checks}.

In the retrieval control, we retrieved the evidence of each MIMIC-CXR test report finding from the chest sentences of the held-out articles, with the statement as the query. We retrieved it once for the statement of the state in the report and once for the opposite statement. BM25~\cite{robertson2009probabilistic} with $k_1 = 1.2$ and $b = 0.75$ returned the top 2 or the top 5 sentences. Its inverse document frequency of a term that occurs in $n_t$ of the $N$ sentences is $\log\big(1 + (N - n_t + 0.5)/(n_t + 0.5)\big)$. The dense retriever Qwen3-Embedding-8B~\cite{zhang2025qwen3embedding} returned the top 2 sentences by the cosine similarity of unit-length embeddings. Its query was prefixed by the instruction to retrieve reference sentences that support a statement about a chest radiograph. In a fourth condition, Qwen3-Reranker-8B~\cite{zhang2025qwen3embedding} reranked the 80 sentences with the highest BM25 scores and kept the top 2. A retrieved example is labeled as supported when its statement agrees with the state in the report and a retrieved sentence belongs to the held-out pool of the stated finding state. For each condition, we report the share of the MIMIC-CXR test report findings whose retrieved sentences for the statement of their state contain a pool sentence (Supplementary Fig.~\ref{sfig:unseen}c). Where a condition has at least 30 supported examples, we also report the AUROC of ModernBERT-large trained on level 1 and with BRACE.

\subsection*{AI-written statements and their citations}

We split the text of each source into sentences after a period, a question mark, or an exclamation mark and removed a leading section label or list number. We kept a sentence when it names a single one of the seven findings, as decided by the mention patterns of Supplementary Note~\ref{snote:data}. We labeled each kept sentence with CheXbert. A sentence labeled positive states its finding as present, a sentence labeled negative states it as absent, and a sentence labeled blank or uncertain is dropped. We also labeled the impression of the study with CheXbert. The report state of a statement is the state of its finding in the impression under the mapping of the report findings. A statement whose report state is uncertain is dropped. The report of the statement is the findings section of its study. Each statement has two forms, the template ``$f$ is $s$'' and the original sentence. The template is the primary form. Within a study, we kept one statement in the template form per finding, state, and sentence status (unchanged or error) and one statement in the original form per sentence text and status.

The ReXErr statements come from the studies of the MIMIC-CXR test patients with a findings and an impression section. They are the unchanged sentences and the sentences with a false prediction or a false negation. Under the report definition, a statement is consistent when its state equals the report state. Under the ReXErr definition, an unchanged sentence is consistent and an error sentence is contradicted. Under the agree definition, a statement keeps its label when the two definitions agree and is excluded otherwise. The agree definition is the primary definition. The generator statements come from findings sections written for the Open-I radiographs by MAIRA-2~\cite{bannur2024maira2}, LLaVA-Rad~\cite{zambranochaves2025llavarad}, MedGemma-1.5-4B~\cite{sellergren2025medgemma}, LLaVA-Med~\cite{li2023llavamed}, Gemma-4-31B~\cite{gemma2026gemma4}, and Qwen3-VL-32B~\cite{qwen2025qwen3vl}. Each generator received the frontal image of a report at $224 \times 224$ pixels (Supplementary Note~\ref{snote:data}) and decoded greedily with at most 512 new tokens. MAIRA-2 received the image in its reporting input format, and the other generators received the prompt of Supplementary Note~\ref{snote:inputs}. LLaVA-Med was served from its released weights converted to the Hugging Face format. The generator statements take the report definition. Table~\ref{tab:datasets} lists the statements of every source with their labels. We also pooled the statements of the six generators. Their accepted shares are further averaged over the generators with equal weight and computed without LLaVA-Med, which wrote most of the contradicted generator statements.

Each statement is scored with up to three citations. The training citation is the most common unchanged evidence of the supported MIMIC-CXR test examples of the stated finding state, with ties broken by the alphabetical order of the evidence. Therefore, we left out the statements of atelectasis absent and pulmonary edema absent, the two finding states without a training pool. The held-out citation is the most common evidence of the same examples with the held-out pools. A statement whose finding state has no held-out pool has no held-out citation. The retrieved citation consists of the two sentences retrieved by BM25 from the held-out chest sentences with the statement text as the query. Its evidence factor is $\sigma = 1$ when one of its sentences belongs to the held-out pool of the stated finding state and $\sigma = 0$ otherwise. Under the retrieved citation, the label of a statement is $\lambda\sigma$ as in Eq.~\ref{eq:label}, with $\lambda$ its label under the report definition. Under this joint label, 401 ReXErr template statements satisfy both requirements and 2,241 fail at least one. We report the accepted share of a verifier in each of the four combinations of report agreement $\lambda$ and citation support $\sigma$, namely statements with agreement and support, with agreement and no support, with support and no agreement, and with neither.

For every verifier, source, form, and label definition, we computed the share of consistent and of contradicted statements accepted at the threshold $\tau_q$ of the test set and at a threshold $\tau'_q$ that accepts the share $q$ of the consistent statements of the source, for both values of $q$. We also computed the AUROC that separates the consistent from the contradicted statements. The test-set threshold is primary, since it is set before any AI-written statement is seen.

\subsection*{Correlations across the verifiers and the report error score}

Across the 43 verifiers of the radiology domain, we computed the Spearman rank correlation between the patient pair accuracy on the MIMIC-CXR test set with unchanged evidence and the share of contradicted statements accepted. We computed it for every source, form, label definition, threshold basis, $q$, and citation, and again with the AUROC on the MIMIC-CXR test set in place of the patient pair accuracy. The primary correlation uses the ReXErr statements in the template form under the agree definition with the training citation at $\tau_{0.90}$. A second main correlation uses the pooled generator statements under the report definition.

The candidate reports of ReXVal and RadEvalX were split into statements by the rules of the AI-written statements, in the template form and with the training citation. For ReXVal, the report is the released reference report, i.e., the impression of the study. The 14 studies whose patients belong to the MIMIC-CXR training or validation split were left out. The 144 candidates of the other 36 studies and 30 patients remain. The ReXVal patients do not overlap with the MIMIC-CXR test patients. For RadEvalX, the report is the findings section of the Open-I report of the same study. The report error score of a verifier for a candidate report is the share of the statements of the candidate that score below $\tau_{0.90}$. A candidate without a statement is excluded. After this exclusion, 68 ReXVal candidates from 21 patients remain and contain 88 statements. The 33 remaining RadEvalX candidates contain 47 statements. The reference of a candidate is the sum of its clinically significant errors over the error categories, averaged over the radiologists on ReXVal. We report Kendall's $\tau_b$~\cite{kendall1945treatment} between the report error score and the reference per verifier and dataset. As the radiology metric of reference, we report the same statistic for the GREEN score of the whole candidate report, computed with its reference report on all 144 ReXVal and 100 RadEvalX candidates. We also report the Spearman correlation across the verifiers between their patient pair accuracy and their $\tau_b$.

\subsection*{Robustness of the verifiers and of the labels}

The patient subgroups are the sex (female, male), the age group (under 40, 40 to 59, 60 to 79, and 80 and over), and the race (White, Black, Asian, Hispanic or Latino, and other) on MIMIC-CXR and CheXpert Plus, and the sex and the age group on PadChest-GR. An unknown race counts as other. A group with fewer than 30 patients is reported by its counts only. For each verifier trained with BRACE, we report the AUROC per set and group. Under the replaced report, the findings section of each MIMIC-CXR and CheXpert Plus test report finding is replaced by that of another patient whose impression has the same state and the same CheXbert class for the finding. The replacement is a derangement over the patients of each group of finding, state, and class, drawn with a random generator seeded by 13, the dataset, the finding, the state, and the class. The rule for a group without a derangement and the number of report findings replaced by each rule are given in Supplementary Note~\ref{snote:data}. A decision changes when an example is accepted with one of the two reports and rejected with the other at the point value of $\tau_{0.90}$. We report the share of decisions that change over all test examples. We also classified each contradicted example by the kind of contradiction between its report and its statement. The impression states the finding when the statement says absent and its CheXbert class is positive, negates it when the statement says present and the class is negative, and does not mention it when the statement says present and the class is blank. The share of contradicted examples accepted at $\tau_{0.90}$ is reported per kind.

The report labels of CheXbert were compared per finding with three other label sources, namely the labels of the CheXpert labeler~\cite{chexpertmain} released with MIMIC-CXR and CheXpert Plus under the mapping of the CheXbert classes, the MeSH major topics of Open-I, and the finding labels of the PadChest-GR release. The two judges were compared with each other on the state assigned to each combination of a sentence and a finding. Each comparison is reported as Cohen's $\kappa$~\cite{cohen1960coefficient}. The comparison of the judges is also reported as the agreement share. Every verifier was also scored on the Open-I set with the MeSH labels. The patient pair accuracies of the verifiers under the two label sources of Open-I were correlated across the verifiers.

In the similarity control, the similarity of a test report finding to the training data is computed as in Elangovan et al.~\cite{elangovan2021memorization}. It is the cosine similarity between the word $n$-gram count vectors of its findings section and of the nearest of the 3,658 distinct findings sections of the MIMIC-CXR training split, for $n = 1$, 2, and 3, over lowercase tokens of letters and digits without English stop words. The training reference is the same similarity between the findings section of each training report finding and the nearest findings section of another training patient, computed once on all training report findings and once without the findings sections shared by several patients. In the food domain, the same measure is computed between the directions of the test recipes and those of the training recipes. For ModernBERT-large trained with BRACE, we report the AUROC on the examples of the 10\% and 25\% of test report findings that are most and least similar to the training data in each test set, ranked by unigram and by bigram similarity. In two further conditions on the MIMIC-CXR test set, the token [FINDING] replaces the terms of the evidence lexicon finding by finding, in the evidence alone or in the evidence and the statement. Only the term blunting of the costophrenic angle, which none of these evidence sentences contains, is kept. ModernBERT-large trained on levels 1 and 3 was scored under these two conditions.

\subsection*{Reader study}

Two board-certified radiologists read the same four packets of text items. T.T.N. has 8 years of experience in diagnostic and interventional radiology, and S.Z. has 6 years of experience. The two radiologists received the same instructions. Where a packet contains a random sample, the sample was drawn with seed 0. No item was selected by a verifier score. Packet A contains 328 combinations of an evidence sentence and a finding. For each combination, the radiologists answered whether the sentence supports the statement that the finding is present on a patient's chest radiograph, supports the statement that it is absent, or supports neither, reading the sentence as general radiology knowledge. The combinations are the 43 combinations of the training and held-out pools; the 24 candidates of the training articles outside the pool of four; 150 written sentences, all 112 held-out and 38 training sentences; the 11 rule-selected sentences of the comparison set not accepted by both judges; 50 combinations outside the candidates assigned to a state by at least one judge; and 50 combinations assigned to a state by neither judge. Packet B contains 150 report findings of the five test sets, namely 35 from MIMIC-CXR, 30 from CheXpert Plus, 25 from ReXGradient-160K, 30 from Open-I, and 30 from PadChest-GR. Each report finding is shown with its finding and with the findings section and the impression of its report, or with the English sentences of a PadChest-GR report. For each report finding, the radiologists answered whether the report states the finding as present, as absent, or as uncertain, or does not mention it. Packet C contains 150 generator statements in their original wording, 25 per generator with 12 consistent and 13 contradicted under the report definition. Each statement is shown with its finding and with the findings section and the impression of its Open-I report. The radiologists answered whether the report agrees with the sentence about the finding, contradicts it, or does not address it. Packet D contains 100 ReXErr sentences, namely 25 false predictions, 25 false negations, and 50 unchanged sentences. Each sentence is shown with the report of its study and the question of packet C. The composition of each packet and the reading instructions are given in Supplementary Note~\ref{snote:reader}.

The items of each packet were shown in random order, without any label and without the name of their source, generator, dataset, or judges. The instructions asked the radiologists to start with 10 practice items from outside the packets, namely 3 for packet A, 3 for packet B, and 2 each for packets C and D. The expected answers of the practice items were the automatic labels. In packet B, the instructions asked the radiologists to answer from the impression where the findings section and the impression differ.

In packet A, a support sentence is confirmed by a radiologist when the answer is its assigned state, i.e., the state of its pool, of its request to the writer, or of the keyword rule. It is confirmed by both radiologists when both answers are its assigned state. We report the share of confirmed sentences per group for each radiologist and for both radiologists. For each judge, the answer is its assigned state, or neither if it assigned no state. For both judges together, the answer is their state if they assigned the same state, and neither otherwise. Each judge and both judges together are compared with each radiologist on all 328 combinations. In packets B to D, the answers are converted to labels by the labeling convention of the report findings (Supplementary Note~\ref{snote:reader}). These labels are compared with the report labels of the test sets in packet B, with the report definition in packet C, and with each of the three label definitions in packet D, on the sentences labeled under that definition. The four answers of packet B are also compared with the four CheXbert classes. A report finding answered uncertain by a radiologist, or an item left unanswered by a radiologist, is left out of the comparisons with that radiologist. The agreement of the two radiologists is the share of the items answered by both radiologists to which they gave the same answer. It is computed on the answers of each packet and on the labels of packets B to D. Every comparison is reported as an agreement share. Cohen's $\kappa$ is reported for every comparison over all items of a packet or a label definition (Supplementary Table~\ref{stab:reader}).

We then repeated the evaluation on the support confirmed by both radiologists and again on the support confirmed by each radiologist, without retraining and without changing a pool. A supported or contradicted test example is kept when one of its evidence sentences is confirmed in its finding state. A statement with the training or the held-out citation is kept by the same rule. Every example without supporting evidence is kept. With the retrieved citation, a statement whose retrieved sentences include a held-out pool sentence of its finding state is kept when one of these sentences is confirmed. Every other statement is kept. On the five test sets with unchanged evidence and on the MIMIC-CXR test set with the held-out and the written held-out sentences, we recomputed the AUROC, the pair accuracies, and the accepted shares of the 43 verifiers. The results on the AI-written statements and the correlations across the verifiers were recomputed in the same way (Supplementary Table~\ref{stab:correlations}). The threshold $\tau_q$ of each verifier is recomputed on the restricted MIMIC-CXR test set with unchanged evidence.

\subsection*{Statistical analysis}

The AUROCs, the pair accuracies, the accepted shares, the shares of changed decisions, the agreement shares, the shares of confirmed sentences, and the shares of test report findings with a retrieved support sentence are reported as the bootstrap mean $\pm$ standard deviation [95\% CI], with the CI from the 2.5th to the 97.5th percentile of the bootstrap resamples. The attainable AUROCs of Eq.~\ref{eq:bound} are exact properties of a set. The accepted shares of each example type under the evidence tests and the accepted shares averaged over the generators with equal weight are descriptive values without an interval. Paired differences between two verifiers are reported with their CI. The differences between the levels of training evidence are differences of the bootstrap means and have no interval.

Each bootstrap uses 1,000 resamples~\cite{efron1994introduction} with seed 0. The resampling unit is the patient for radiology, except on Open-I, for which it is the report; the recipe for food; and the evidence sentence for support-label agreement. All examples of one unit enter a resample together. Each metric of a resample weights every example by its multiplicity. Each evidence pair receives the shared multiplicity of its two examples once. A patient pair or a recipe pair receives the product of the multiplicities of its two examples. Every AUROC counts tied scores as one half. The threshold $\tau_q$ is recomputed in every resample of the set on which it is defined and kept at its point value elsewhere. The comparison of Qwen3-8B finetuned with LoRA and its zero-shot version uses a paired bootstrap over the same examples, in which the difference in AUROC, in patient pair accuracy, or in the share of contradicted ReXErr statements accepted at the threshold of each model is computed on every resample. Its two-sided $p$-value is the share of resamples whose difference deviates from the observed difference by at least the absolute observed difference. For Kendall's $\tau_b$ of each verifier, the $p$-value is the share of resamples whose $\tau_b$ deviates from the observed $\tau_b$ by at least its absolute value. A resample with an undefined $\tau_b$ counts toward this share. Where several items share a resampling unit, the $p$-value of Cohen's $\kappa$ is computed in the same way. Otherwise, it is the analytic $p$-value under the null hypothesis of a $\kappa$ of 0. The Spearman correlations across the verifiers use the bootstrap mean of each metric. Their $p$-values are analytic and treat each verifier as one observation. The $p$-values are FDR-corrected~\cite{benjamini1995controlling} within seven families: (i) every correlation across verifiers with the accepted share of contradicted statements; (ii) the Kendall and Spearman tests of the report metrics; (iii) the $\kappa$ tests of the label sources, of the judges, of the comparisons with each radiologist, and of the agreement between the two radiologists, together with the correlation between the two Open-I label sources; (iv) the paired tests that compare each of the four versions of Qwen3-8B finetuned with LoRA with the zero-shot model and with the other three versions; and (v) to (vii) the correlations of family (i) recomputed on the support confirmed by T.T.N., by S.Z., and by both radiologists, one family for each. The corrected $p$-values are reported and written $p_{\mathrm{FDR}}$. The significance threshold is 0.05. Every other comparison is descriptive and is stated by the overlap of the intervals.

\section*{Data availability}

Every dataset and reference text of this study comes from an existing public source. None of their images, reports, or derived records is redistributed here.

\begingroup
\footnotesize
\sloppy
\textbf{Chest radiograph datasets:}
\begin{itemize}
  \item MIMIC-CXR~\cite{johnson2019mimic}, reports and the labels of the CheXpert labeler, available from PhysioNet under credentialed access: \url{https://physionet.org/content/mimic-cxr/} and \url{https://physionet.org/content/mimic-cxr-jpg/}.
  \item MIMIC-IV~\cite{johnson2023mimiciv}, the age, sex, and race fields, available from PhysioNet under credentialed access: \url{https://physionet.org/content/mimiciv/}.
  \item CheXpert Plus~\cite{baker2024chexpertplus}, reports, labels, and patient demographics, available from the Stanford AIMI portal after registration and acceptance of its terms: \url{https://aimi.stanford.edu/datasets/chexpert-plus}.
  \item ReXGradient-160K~\cite{zhang2025rexgradient}, reports, provided by Gradient Health and available on the Hugging Face Hub after acceptance of its terms: \url{https://huggingface.co/datasets/rajpurkarlab/ReXGradient-160K}.
  \item Open-I~\cite{demnerfushman2016preparing}, reports and radiographs, the source of the examples in Figs.~\ref{fig:story} and~\ref{fig:framework}, openly available: \url{https://openi.nlm.nih.gov/}.
  \item PadChest-GR~\cite{castro2024padchestgr}, reports, labels, and patient demographics, available on request from the BIMCV repository: \url{https://bimcv.cipf.es/bimcv-projects/padchest-gr/}.
\end{itemize}
\textbf{Reports with annotated errors:}
\begin{itemize}
  \item ReXErr-v1~\cite{rao2024rexerr}, openly available from PhysioNet: \url{https://physionet.org/content/rexerr-v1/1.0.0/}.
  \item ReXVal~\cite{yu2023evaluating}, available from PhysioNet under credentialed access: \url{https://physionet.org/content/rexval-dataset/1.0.0/}.
  \item RadEvalX~\cite{calamida2023radiology}, openly available from PhysioNet: \url{https://physionet.org/content/rad-eval-x/1.0.0/}.
\end{itemize}
\textbf{Reference text:}
\begin{itemize}
  \item Radiopaedia~\cite{gaillard2011radiopaedia}, articles, available on request from Radiopaedia: \url{https://radiopaedia.org/}.
\end{itemize}
\textbf{Food domain:}
\begin{itemize}
  \item RecipeNLG~\cite{bien2020recipenlg}, available from its project page after acceptance of its terms: \url{https://recipenlg.cs.put.poznan.pl/}.
  \item English Wikipedia, the 31 articles of Supplementary Note~\ref{snote:data}, openly available: \url{https://en.wikipedia.org/}.
\end{itemize}
\endgroup

\section*{Code availability}

The analysis code is publicly available at \url{https://github.com/tayebiarasteh/grounding}. The repository does not redistribute the model weights or the underlying datasets.

Every pretrained model of this study has open weights. No model ran on a cloud service or through a third-party application programming interface. The TF-IDF baseline uses no pretrained model. All models were accessed and all experiments were run between March and October 2026. The URLs of the checkpoints used in this study are:

\begingroup
\footnotesize
\sloppy
\textbf{Starting checkpoints of the trained verifiers:}
\begin{itemize}
  \item ModernBERT-large: \url{https://huggingface.co/answerdotai/ModernBERT-large}
  \item Flan-T5-large: \url{https://huggingface.co/google/flan-t5-large}
  \item RoBERTa-large: \url{https://huggingface.co/FacebookAI/roberta-large}
  \item ELECTRA-large: \url{https://huggingface.co/google/electra-large-discriminator}
  \item Flan-T5-base: \url{https://huggingface.co/google/flan-t5-base}
  \item ModernBERT-base: \url{https://huggingface.co/answerdotai/ModernBERT-base}
  \item Longformer-base: \url{https://huggingface.co/allenai/longformer-base-4096}
  \item BiomedBERT-base: \url{https://huggingface.co/microsoft/BiomedNLP-BiomedBERT-base-uncased-abstract-fulltext}
  \item Qwen3-8B, also run zero-shot: \url{https://huggingface.co/Qwen/Qwen3-8B}
\end{itemize}
\textbf{Public verifiers:}
\begin{itemize}
  \item MiniCheck-Flan-T5-Large: \url{https://huggingface.co/lytang/MiniCheck-Flan-T5-Large}
  \item Bespoke-MiniCheck-7B: \url{https://huggingface.co/bespokelabs/Bespoke-MiniCheck-7B}
  \item AlignScore-large: \url{https://huggingface.co/yzha/AlignScore}
  \item HHEM-2.1-Open: \url{https://huggingface.co/vectara/hallucination_evaluation_model}
  \item LettuceDetect: \url{https://huggingface.co/KRLabsOrg/lettucedect-large-modernbert-en-v1}
  \item Lynx-8B: \url{https://huggingface.co/PatronusAI/Llama-3-Patronus-Lynx-8B-Instruct-v1.1}
  \item GREEN: \url{https://huggingface.co/StanfordAIMI/GREEN-RadLlama2-7b}
\end{itemize}
\textbf{Zero-shot LLM verifiers:}
\begin{itemize}
  \item DeepSeek-V4-Flash: \url{https://huggingface.co/MJPansa/DeepSeek-V4-Flash-0731-NVFP4}
  \item Qwen3.8-Flash-Next: \url{https://huggingface.co/nvidia/Qwen3.8-Flash-Next-NVFP4}
  \item Mistral-Small-4-119B: \url{https://huggingface.co/mistralai/Mistral-Small-4-119B-2603-NVFP4}
  \item MedGemma-27B: \url{https://huggingface.co/google/medgemma-27b-it}
\end{itemize}
\textbf{Judges and writer of the evidence sentences:}
\begin{itemize}
  \item gpt-oss-120b: \url{https://huggingface.co/openai/gpt-oss-120b}
  \item Qwen3.5-122B-A10B: \url{https://huggingface.co/Qwen/Qwen3.5-122B-A10B-GPTQ-Int4}
  \item Qwen3.5-35B-A3B: \url{https://huggingface.co/Qwen/Qwen3.5-35B-A3B-GPTQ-Int4}
\end{itemize}
\textbf{Report generators:}
\begin{itemize}
  \item MAIRA-2: \url{https://huggingface.co/microsoft/maira-2}
  \item LLaVA-Rad: \url{https://huggingface.co/microsoft/llava-rad}
  \item LLaVA-Rad code: \url{https://github.com/microsoft/llava-rad}
  \item Vicuna-7B v1.5, the base language model of LLaVA-Rad: \url{https://huggingface.co/lmsys/vicuna-7b-v1.5}
  \item MedGemma-1.5-4B: \url{https://huggingface.co/google/medgemma-1.5-4b-it}
  \item LLaVA-Med: \url{https://huggingface.co/microsoft/llava-med-v1.5-mistral-7b}
  \item Gemma-4-31B: \url{https://huggingface.co/google/gemma-4-31B-it}
  \item Qwen3-VL-32B: \url{https://huggingface.co/Qwen/Qwen3-VL-32B-Instruct}
\end{itemize}
\textbf{Report labeling, image selection, and retrieval:}
\begin{itemize}
  \item CheXbert (from the RRG scorers): \url{https://huggingface.co/StanfordAIMI/RRG_scorers}
  \item BiomedCLIP: \url{https://huggingface.co/microsoft/BiomedCLIP-PubMedBERT_256-vit_base_patch16_224}
  \item Qwen3-Embedding-8B: \url{https://huggingface.co/Qwen/Qwen3-Embedding-8B}
  \item Qwen3-Reranker-8B: \url{https://huggingface.co/Qwen/Qwen3-Reranker-8B}
\end{itemize}
\endgroup

The code ran under Python~3.11 in two environments. The construction of the examples, the training and scoring of the verifiers, and the statistical analyses ran in the first environment, with PyTorch~2.9 (CUDA~13.0), Transformers~5.13, huggingface-hub~1.22, PEFT~0.19, NumPy~2.4, SciPy~1.17, pandas~2.3, and scikit-learn~1.4. MAIRA-2, LLaVA-Rad, and Bespoke-MiniCheck-7B ran in separate processes under Transformers~4.51, tokenizers~0.21, and huggingface-hub~0.35. The models on the serving system of our institution ran under vLLM and were accessed with the OpenAI Python client~2.26 and httpx~0.28. The dataset statistics and the figures were produced in the second environment, with NumPy~2.4, pandas~2.3, and Matplotlib~3.10. No figure contains an image produced by generative AI. The models on the serving system ran on NVIDIA RTX PRO 6000 GPUs. The training runs, the in-process verifiers, MAIRA-2, LLaVA-Rad, and CheXbert ran on NVIDIA L40S GPUs with 48~GB of memory and Intel Xeon Silver 4310 CPUs, and the statistical analyses ran on CPUs.

\section*{Acknowledgements}

STA is supported by the German Research Foundation (DFG; 585806541), the German Federal Ministry of Research, Technology and Space (ARISTOTLE, 01ZU2602B), and the Excellence Strategy of the German Federal Government, the L\"ander, and RWTH ERS (START\_526-26). SZ is supported by the German Federal Ministry of Research, Technology and Space (ARISTOTLE, 01ZU2602A). SN is supported by the DFG (701010997, 517243167). DT is supported by the German Federal Ministry of Research, Technology and Space (TRANSFORM LIVER, 031L0312C; DECIPHER-M, 01KD2420B), the DFG (515639690), and the European Union (Horizon Europe, ODELIA, GA 101057091; ERC Starting Grant SAGMA, GA 101222556).

\section*{Author contributions}

The formal analysis was conducted by STA and DT. The original draft was written by STA. STA developed the code. The experiments were performed by STA. The statistical analyses were performed by STA, MJ, ML, SN, and DT. TTN and SZ performed the reader study. TTN, SZ, SN, and DT provided clinical expertise. STA, MJ, ML, and DT provided technical expertise. The study was defined by STA. All authors read the manuscript and agreed to the submission of this paper.

\section*{Competing interests}

STA received honoraria for lectures from RWTH International Academy gGmbH, Germany and is on the editorial boards of Communications Medicine and of European Radiology Experimental, and on the trainee editorial board of Radiology: Artificial Intelligence. ML is employed by Generali Deutschland Services GmbH, Germany, and is on the editorial board of European Radiology Experimental. DT received honoraria for lectures from Bayer, GE, Roche, AstraZeneca, and Philips and holds shares in StratifAI GmbH, Germany, and in Synagen GmbH, Germany. The authors declare no other competing financial or non-financial interests.

\bibliographystyle{splncs04}
\bibliography{references}

\setcounter{table}{0}
\setcounter{figure}{0}
\setcounter{equation}{0}
\renewcommand{\tablename}{Supplementary Table}
\renewcommand{\figurename}{Supplementary Fig.}
\floatname{algorithm}{Supplementary Algorithm}
\renewcommand{\thealgorithm}{\arabic{algorithm}}
\renewcommand{\theequation}{S\arabic{equation}}
\newcounter{snote}\setcounter{snote}{0}

\section*{Supplementary information}

\begin{figure*}[p]
\centering
\includegraphics[width=\textwidth]{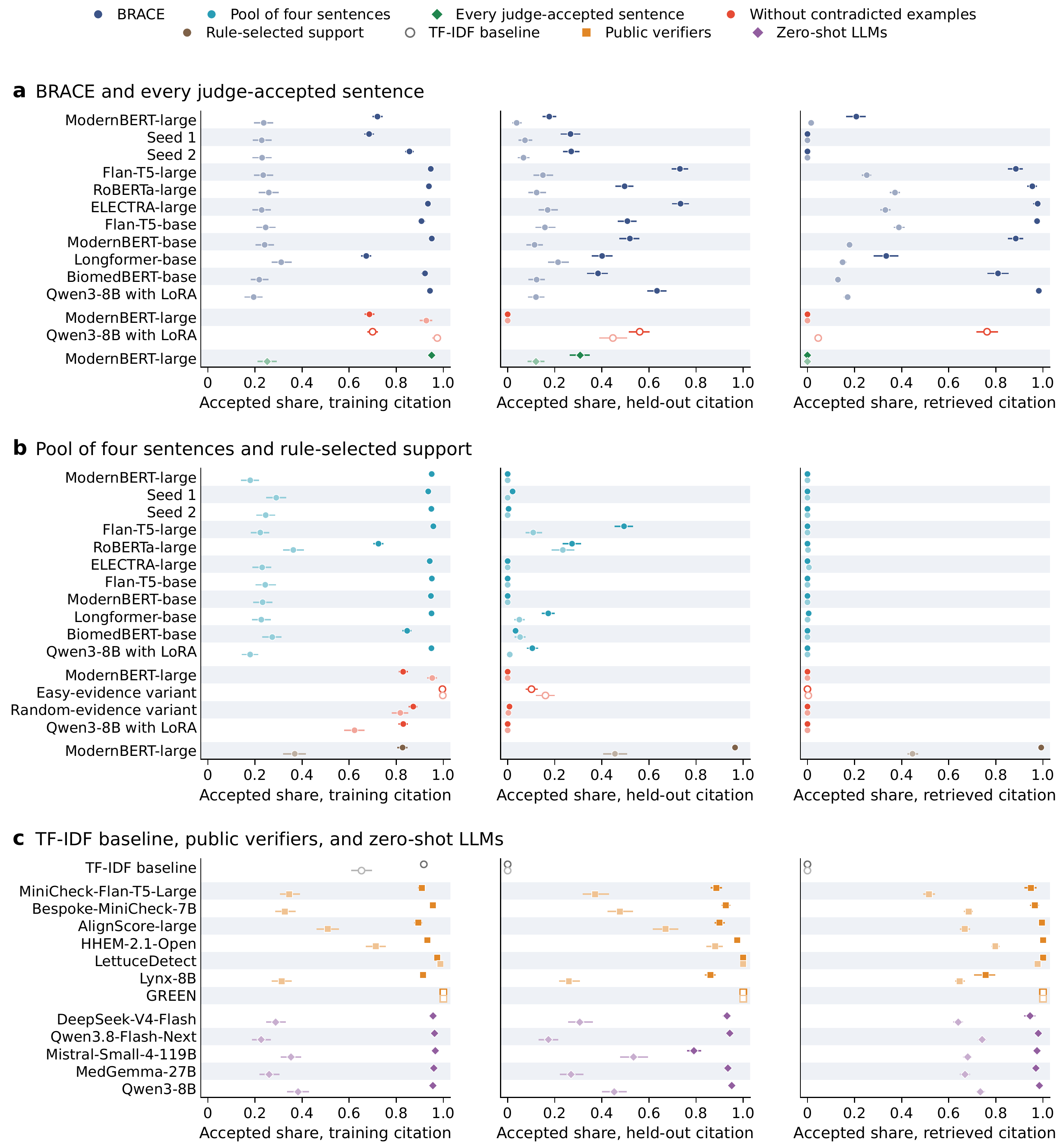}
\caption{Accepted share at $\tau_{0.90}$ (Eq.~\ref{eq:operating}) by label of the ReXErr~\cite{rao2024rexerr} statements of MIMIC-CXR test patients for the 43 verifiers. Color and marker shape mark the verifier group. Upper markers in solid color have positive labels. Lower markers in a light tint have negative labels. Light bands mark every second verifier. Markers are bootstrap means over 1,000 resamples of the patients. Error bars in every panel are 95\% CIs from the 2.5th to the 97.5th percentile. Counts per citation: 1,563 positive and 472 negative labels (training), 762 and 343 (held-out), and 401 and 2,241 (retrieved). \textbf{a}--\textbf{c}, One group per row. The labels take the agree definition with the training and held-out citations, i.e., a support sentence of the training or the held-out articles. The retrieved citation is two sentences retrieved by BM25 from the held-out articles. Its joint label is positive when the report agrees and a retrieved sentence supports the statement, and negative otherwise. Open orange and red markers are the three verifiers that accept more than 0.95 of the supported examples.}
\label{sfig:rexerr}
\end{figure*}

\begin{figure*}[p]
\centering
\includegraphics[width=\textwidth]{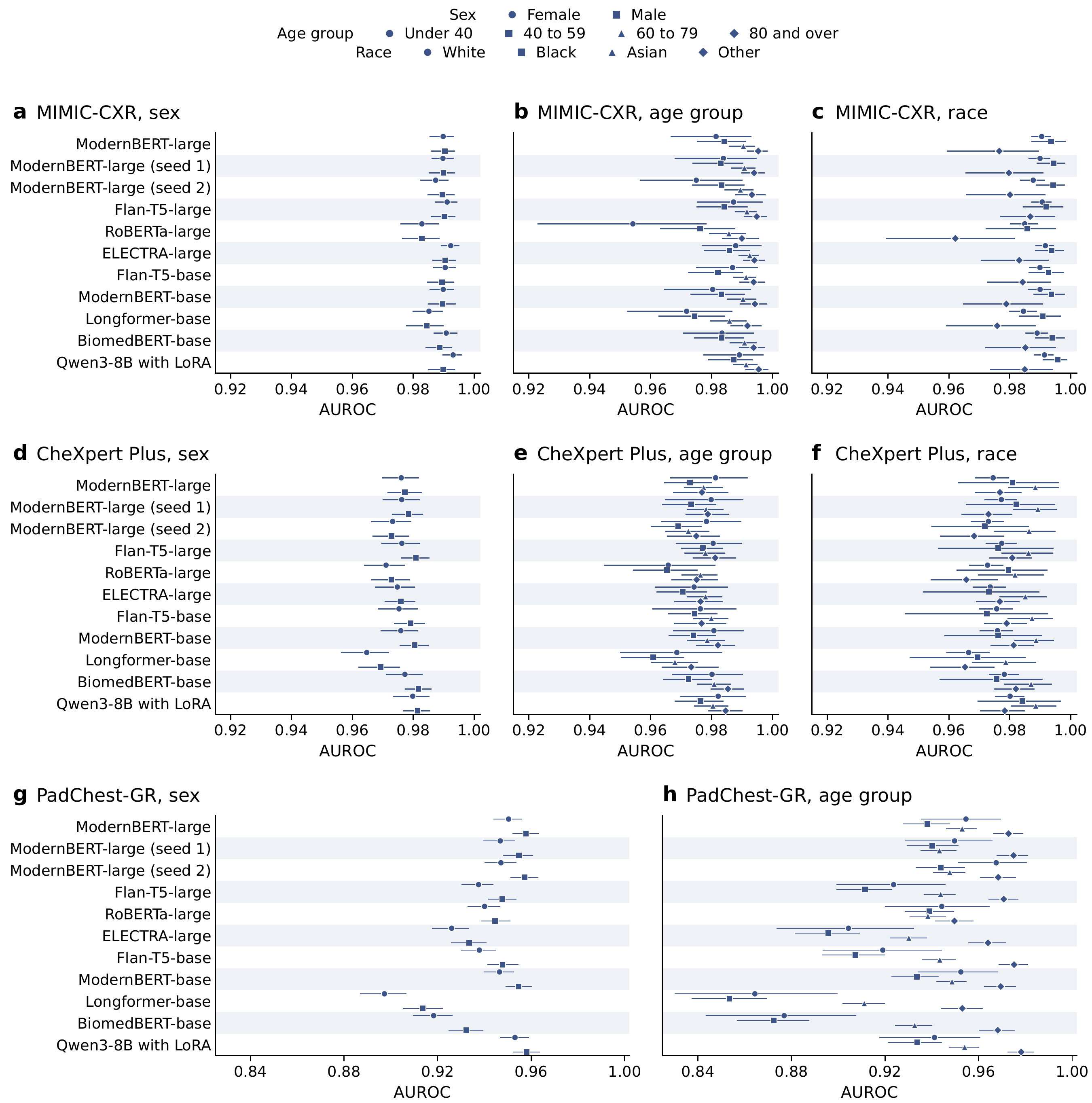}
\caption{AUROC of the 11 verifiers trained with BRACE per patient subgroup, on the MIMIC-CXR, CheXpert Plus~\cite{baker2024chexpertplus}, and PadChest-GR~\cite{castro2024padchestgr} test sets with unchanged evidence ($n = 4{,}356$, $6{,}077$, and $9{,}254$ examples). Each row is one verifier, with one marker per subgroup in the shape of the legend. Light bands mark every second verifier. Each marker is the bootstrap mean of the AUROC over 1,000 resamples of the patients. Error bars are 95\% CIs from the 2.5th to the 97.5th percentile. Every panel has error bars. Subgroups with fewer than 30 patients are not drawn, namely the 21 Hispanic or Latino and the 14 Asian patients of MIMIC-CXR. The 30 MIMIC-CXR test patients without a record in the patient table of MIMIC-IV are in no subgroup. One CheXpert Plus patient with an unknown sex and age is in no subgroup of these two attributes. Patients per subgroup, in the order of the legend: 281 and 237 (\textbf{a}); 53, 125, 219, and 121 (\textbf{b}); 322 White, 97 Black, and 64 other (\textbf{c}); 298 and 416 (\textbf{d}); 87, 191, 279, and 157 (\textbf{e}); 425, 36, 75, and 179 (\textbf{f}); 958 and 945 (\textbf{g}); and 126, 514, 900, and 363 (\textbf{h}). \textbf{a}--\textbf{c}, MIMIC-CXR. \textbf{d}--\textbf{f}, CheXpert Plus. \textbf{g},\textbf{h}, PadChest-GR, whose release records no race. The AUROC axes of \textbf{g} and \textbf{h} start at a lower value than those of \textbf{a}--\textbf{f}.}
\label{sfig:subgroups}
\end{figure*}

The Supplementary Notes contain the datasets with their approvals and the rules for their reports and sentences (Supplementary Note~\ref{snote:data}); the inputs of the verifiers, the judges, the writer, and the report generators (Supplementary Note~\ref{snote:inputs}); the proofs of Eqs.~\ref{eq:half} and~\ref{eq:bound} (Supplementary Note~\ref{snote:proofs}); and the packets and reading instructions of the reader study (Supplementary Note~\ref{snote:reader}). The construction of the training examples in Building Report And Citation Examples (BRACE) and the patient-level checks are written in pseudocode (Supplementary Algorithms~\ref{salg:construction} and~\ref{salg:checks}). The supplementary figures plot every verifier on the MIMIC-CXR test set (Supplementary Fig.~\ref{sfig:mimic}), the tests with unseen support sentences and the controls (Supplementary Fig.~\ref{sfig:unseen}), the accepted shares of the ReXErr statements (Supplementary Fig.~\ref{sfig:rexerr}), the area under the receiver operating characteristic curve (AUROC) of the verifiers trained with BRACE per patient subgroup (Supplementary Fig.~\ref{sfig:subgroups}), and the support sentences per finding state with the judgments of the written candidates (Supplementary Fig.~\ref{sfig:support}). The supplementary tables are a glossary (Supplementary Table~\ref{stab:glossary}), a summary of the findings (Supplementary Table~\ref{stab:findings}), and tables of the verifiers (Supplementary Table~\ref{stab:models}), the correlations across the verifiers (Supplementary Table~\ref{stab:correlations}), the report error scores (Supplementary Table~\ref{stab:report_metric}), the agreement with the radiologists (Supplementary Table~\ref{stab:reader}), and the report findings and examples per finding (Supplementary Tables~\ref{stab:examples} and~\ref{stab:examples_other}).

\refstepcounter{snote}
\section*{Supplementary Note \thesnote: Datasets and the rules for reports and sentences}
\label{snote:data}

\subsection*{Section rules and mention patterns}

The findings and impression sections of a MIMIC-CXR~\cite{johnson2019mimic} report are the text under the headings FINDINGS and IMPRESSION. Lines containing one of the phrases discussed with, communicated to, called to, paged, notified, by phone, phone at, results were communicated, recommend short-term follow-up, these findings were discovered, covering MD, and immediately paged were removed. A section passed the section rules when its lowercase tokens of letters and digits contain at least five distinct words and at least one of the chest terms lung, lungs, pleural, effusion, pneumothorax, cardiac, cardiomediastinal, opacity, opacities, atelectasis, atelectatic, consolidation, edema, hilar, mediastinal, silhouette, vascular, retrocardiac, basilar, interstitial, fracture, fractures, rib, ribs, and osseous. The CheXpert Plus~\cite{baker2024chexpertplus} sections are the corresponding fields of its release, cleaned of the same phrases, and a section passed when it contains at least four distinct words and one chest term from the list above extended with effusions, ptx, bibasilar, airspace, and aeration. The study lists of MIMIC-CXR and CheXpert Plus contain the reports in which both sections passed these rules and in which a keyword match found at least one finding stated in the impression and not mentioned in the findings section. The report findings of a listed study do not depend on the keyword match.

A report sentence, a ReXErr~\cite{rao2024rexerr} sentence, a generator sentence, or a candidate report sentence names a finding when it contains one of the following patterns as whole words, regardless of case: pleural effusion, pleural effusions, effusion, or effusions for pleural effusion; pneumothorax or pneumothoraces for pneumothorax; cardiomegaly, enlarged cardiac silhouette, cardiac silhouette is enlarged, cardiac silhouette enlarged, enlarged heart, heart is enlarged, heart size is enlarged, heart size enlarged, heart is mildly enlarged, or mild cardiomegaly for cardiomegaly; atelectasis, atelectatic, subsegmental atelectatic, bibasilar atelectatic, or basal atelectatic for atelectasis; consolidation, lobar consolidation, focal consolidation, or airspace consolidation for consolidation; pulmonary edema, edema, vascular congestion, pulmonary vascular congestion, central pulmonary vascular congestion, vascular indistinctness, hilar vasculature, interstitial edema, or interstitial pulmonary edema for pulmonary edema; and fracture, fractures, rib fracture, rib fractures, fractured rib, or fractured ribs for fracture. A sentence names a single finding when the patterns of only one finding match.

\subsection*{Chest radiograph datasets}

MIMIC-CXR contains 227,835 radiology reports of chest radiographs from 65,379 patients of Beth Israel Deaconess Medical Center. It was approved by the institutional review board of this center, which waived individual patient consent. With seed 13, all its patients were assigned at random to a training set with 75\% of the patients, a validation set with 10\%, and a test set with 15\%. Its study list contains 5,242 reports from 4,305 patients. Of these reports, 4,052 from 3,336 patients are in the training set, 477 from 378 patients are in the validation set, and 713 from 591 patients are in the test set. After the drawing of the absent findings, 3,673 training studies from 3,036 patients and 422 validation studies from 334 patients have at least one report finding. The training studies have 5,227 present and 1,615 absent findings, and the validation studies have 602 present and 167 absent findings. The sex and the anchor age of the patients come from the patient table of MIMIC-IV~\cite{johnson2023mimiciv}, approved by the same board with the same waiver. The race is the most frequent value over the admissions of the patient.

CheXpert Plus adds the reports of the chest radiographs of CheXpert~\cite{chexpertmain} from Stanford Hospital, 223,460 report rows of 187,711 studies from 64,725 patients. CheXpert was approved by the institutional review board of Stanford Hospital, which waived individual patient consent for the de-identified data. The study uses the 917 test studies of 764 patients in its study list. Where a study folder contains several reports, we kept the first report whose findings section passed the section rules. The report file of the release records the sex, the age, and the race of each patient at the study, with no Hispanic or Latino category. The present and absent findings of MIMIC-CXR and CheXpert Plus are listed per split and finding in Supplementary Table~\ref{stab:examples}.

ReXGradient-160K~\cite{zhang2025rexgradient} contains 160,000 chest radiograph studies with reports from 109,487 patients of 3 health systems in the United States. Its source publication reports no approving committee and no consent procedure. Its public test split contains 10,000 studies of 6,807 patients. Of these, 9,957 studies of 6,784 patients have a findings and an impression section and a radiograph recorded as computed radiography (code CR) or digital radiography (codes DX and DR). We drew one study per patient at random and then 1,000 patients at random, with seed 0. Only the report text of this dataset is used.

Open-I~\cite{demnerfushman2016preparing} contains 3,955 chest radiograph reports with 7,470 images from 2 hospital systems of the Indiana Network for Patient Care. It was collected under an institutional review board approval (OHSRP 5357) and de-identified before its release. Its source publication reports no consent procedure. Of its reports, 3,419 have a findings and an impression section, and the 3,331 of them with at least one image form the test set. Each report also has Medical Subject Headings (MeSH) major topics coded by hand by two coders of the collection. For the second Open-I test set, a finding is present when the text before the first slash of a major topic equals the MeSH heading of the finding, namely Pulmonary Atelectasis; Cardiomegaly; Consolidation; Fractures, Bone; Pleural Effusion; Pneumothorax; or Pulmonary Edema. Otherwise, the finding is absent. The 87 reports with the topic No Indexing were not coded and are left out of this set. Open-I records no view per image. Therefore, the frontal image of each report was selected for the report generators by BiomedCLIP~\cite{zhang2025biomedclip} with three frontal and three lateral zero-shot prompts. It is the image with the highest frontal probability among the images with a probability of at least 0.01 and a pixel standard deviation of at least 15. Such an image exists for 3,194 of the 3,331 reports of the test set. No image met these selection criteria in the other 137 reports.

PadChest-GR~\cite{castro2024padchestgr} contains 4,555 chest radiograph studies of 4,459 patients of the University Hospital Sant Joan d'Alacant, Spain, each with the sentences of its positive and negative findings in Spanish and in English and with the finding labels of the positive sentences. An institutional research committee approved PadChest, and PadChest-GR adds labeled and translated report sentences of its studies. All 4,555 studies enter the test set. The report of a study is its English sentences joined in their order. A finding is present when a positive sentence of the study carries one of the labels of the finding in the release, and absent otherwise. The labels of the findings are atelectasis, laminar atelectasis, lobar atelectasis, segmental atelectasis, atelectasis basal, or total atelectasis for atelectasis; cardiomegaly for cardiomegaly; consolidation for consolidation; fracture, rib fracture, callus rib fracture, humeral fracture, clavicle fracture, or vertebral fracture for fracture; pleural effusion, loculated pleural effusion, or loculated fissural effusion for pleural effusion; pneumothorax or hydropneumothorax for pneumothorax; and hilar congestion for pulmonary edema. The class used for the kind of contradiction and for the agreement with the release labels comes from CheXbert~\cite{smit2020chexbert} applied to the English report. The study table of the release records the sex and the age of the patients and no race.

ReXVal~\cite{yu2023evaluating} contains 200 pairs of a candidate report and its reference report for 50 MIMIC-CXR test studies, with the errors of the candidate counted by 6 board-certified radiologists in 6 categories and marked as clinically significant or insignificant. RadEvalX~\cite{calamida2023radiology} contains one report generated by a transformer model for each of 100 Open-I studies, with the errors counted by 2 board-certified radiologists in 8 categories and marked in the same way. ReXErr-v1, ReXVal, and RadEvalX add annotations to MIMIC-CXR and Open-I studies and contain no new patient data.

\subsection*{Radiopaedia sentences, the evidence lexicon, and the comparison set}

The evidence sentences come from a snapshot of Radiopaedia~\cite{gaillard2011radiopaedia} taken on 1 March 2026. Its article text contains no patient records. Each article was split into sentences after a period, a question mark, or an exclamation mark. We kept a sentence when it has 7 to 50 tokens of the BAAI/bge-base-en-v1.5 tokenizer~\cite{xiao2024cpack}, at least 7 words of which at least 5 are distinct, no line break, and an alphanumeric first character, and when it is not the article title. The 57,084 kept sentences come from 14,391 articles.

Of the kept sentences, 55,624 pass 3 cleanup filters, which remove the sentences marked as metadata, as fragments, or as definitions. The cleanup filters mark a sentence as metadata when it contains a metadata phrase (section names such as see also, differential diagnosis, or treatment and prognosis, and editorial phrases such as peer review or more references needed) or the name of any of 11 Radiopaedia authors, and as a fragment when it contains an alias marker (alias, also known as, or aka), at least two semicolons or two colons, fewer than seven words, or no verb from a fixed list of 31 verbs and auxiliaries, when more than 45\% of its words are capitalized, or when it has no final punctuation and at most 14 words. They mark a sentence as a definition when it contains refers to, is defined as, defined as, describes, is a type of, is a form of, is characterized by, represents, consists of, is the presence of, are collections of, or collection of gas within. A sentence contains a chest term when it contains one of chest, thoracic, lung, lungs, pleural, effusion, pneumothorax, cardiac, cardiomegaly, heart, silhouette, vascular, edema, oedema, interstitial, opacity, opacities, airspace, consolidation, atelectasis, atelectatic, collapse, hilar, mediastinal, hemithorax, costophrenic, rib, diaphragm, or parenchymal. Of the kept sentences, 3,920 contain a chest term.

The evidence lexicon recognizes 10 findings, the 7 studied findings together with vascular congestion, interstitial edema, and rib fracture, which count as pulmonary edema and fracture. The studied findings are recognized by these terms: pleural effusion, pleural effusions, effusion, effusions, pleural fluid, pleural collection, costophrenic blunting, blunting of the (right or left) costophrenic angle, and blunted costophrenic angle or angles for pleural effusion; pneumothorax, pneumothoraces, pleural air, and collapsed lung for pneumothorax; cardiomegaly, cardiac enlargement, cardiac silhouette enlargement, enlarged cardiac silhouette, cardiac silhouette is enlarged, cardiac silhouette enlarged, enlarged heart, heart is enlarged, heart size is enlarged, heart size enlarged, mild cardiomegaly, and cardiomegalic for cardiomegaly; atelectasis, atelectatic, subsegmental collapse, lobar collapse, and collapse for atelectasis; consolidation, airspace or air space consolidation, airspace or air space opacity or opacities, alveolar opacity or opacities, parenchymal opacity or opacities, and airspace disease for consolidation; pulmonary edema, pulmonary oedema, edema, oedema, alveolar edema, and alveolar oedema for pulmonary edema; and fracture, fractures, and fractured for fracture. The other findings are recognized by vascular congestion, pulmonary vascular congestion, central pulmonary vascular congestion, vascular indistinctness, hilar vasculature, vascular prominence, prominent pulmonary vasculature, cephalisation, and cephalization for vascular congestion; interstitial edema, interstitial pulmonary edema, interstitial oedema, interstitial pulmonary oedema, septal edema, septal oedema, and interlobular septal thickening for interstitial edema; and rib fracture, rib fractures, fractured rib, and fractured ribs for rib fracture. Among the kept sentences, 1,599 from 994 articles name a studied finding.

With seed 13, the articles were divided into 11,513 training articles and 2,878 held-out articles by a draw stratified by their profile, which is the set of findings of the evidence lexicon mentioned in their sentences. For every profile with at least three articles, 20\% of its articles, rounded and at least one, were drawn at random as held-out articles. Articles that mention no finding then filled the held-out articles up to 20\% of all articles. There are 3,020 chest sentences in the training articles and 749 in the held-out articles.

The support sentences of the comparison set, i.e., the rule-selected support, are selected by a keyword rule. In this rule, the state of a finding in a sentence comes from the 100 characters before and after the first term of the finding. The state is absent when this window contains a negation cue (no, without, absent, negative for, free of, no evidence of, no focal, or neither). Otherwise, it is uncertain when the window contains an uncertainty cue (possible, possibly, may, might, likely, unlikely, cannot exclude, could represent, suggestive of, question of, suspicious for, concerning for, may represent, or appears to), absent when it contains a resolution cue (resolved, improved, decreased, cleared, or near complete resolution), and present without any of these cues. A chest sentence of the training articles is selected as support of the finding state $(f, s)$ when the state of $f$ in the sentence is $s$, when the sentence contains the name of $f$ in lower case, and when it contains any of the 15 imaging keywords on radiograph, on chest radiograph, on x-ray, on xray, on ct, on imaging, radiographic, imaging shows, appears as, seen as, findings include, demonstrates, demonstrated, visible on, and characterized by.

\subsection*{Recipes, allergens, and Wikipedia evidence}

The six allergens of the food domain are dairy, egg, gluten, peanut, soy, and shellfish. We sampled 8,000 recipes from RecipeNLG~\cite{bien2020recipenlg} with seed 13. An allergen is present in a recipe when the ingredient list names it and no negation cue occurs within 80 characters of its first mention, and absent otherwise. Each combination of a recipe and an allergen forms a recipe allergen with its state. We removed the recipe allergens whose directions name the allergen and subsampled the absent recipe allergens of each allergen to the number of its present recipe allergens. With seed 13, the recipes were assigned at random to a training set with 75\% of them, a validation set with 10\%, and a test set with 15\%. The evidence consists of the 6,153 sentences of 20 to 400 characters in 31 English Wikipedia articles on food allergies and on foods that contain the allergens, retrieved on 24 July 2026. The recipes and the Wikipedia articles contain no patient records.

The ingredient list and the directions of a recipe are the corresponding fields of RecipeNLG, with the ingredients joined by semicolons. The six allergens are recognized by these terms as whole words: milk, butter, cheese, cream, yogurt, yoghurt, ghee, whey, casein, buttermilk, and custard for dairy; egg, eggs, yolk, yolks, egg white, egg whites, albumen, and mayonnaise for egg; wheat, flour, bread, barley, rye, malt, gluten, semolina, couscous, bulgur, breadcrumb, breadcrumbs, cracker, and crackers for gluten; peanut, peanuts, groundnut, groundnuts, and peanut butter for peanut; soy, soya, soybean, soybeans, tofu, edamame, miso, tempeh, and soy sauce for soy; and shrimp, prawn, prawns, crab, crabs, lobster, crayfish, crawfish, shellfish, scallop, scallops, clam, clams, mussel, mussels, oyster, oysters, squid, and calamari for shellfish. The negation cues are no, without, free of, free from, does not contain, dairy-free, egg-free, gluten-free, nut-free, peanut-free, soy-free, shellfish-free, excludes, omit, and substitute for. The state of an allergen in a Wikipedia sentence is set by the same negation rule, and a Wikipedia sentence supports the statement of its state when it contains one of the phrases contains, source of, sources of, found in, present in, rich in, made from, made with, derived from, ingredient, allergen, includes, component of, or consists of. An example without supporting evidence takes the first two unsupporting sentences of its allergen in place of a random draw.

The 31 Wikipedia articles are Food allergy, List of allergens, Milk allergy, Egg allergy, Peanut allergy, Soy allergy, Wheat allergy, Shellfish allergy, Gluten-related disorders, Coeliac disease, Milk, Cheese, Butter, Cream, Yogurt, Egg as food, Peanut, Peanut butter, Soybean, Tofu, Soy sauce, Wheat, Bread, Flour, Barley, Rye (grain), Shrimp, Crab, Lobster, Shellfish, and Oyster. The section headings were removed from their plain text, which was split after a period, a question mark, or an exclamation mark followed by a capital letter.

\subsection*{The replaced report}

Under the replaced report (Fig.~\ref{fig:robustness}b), a group of finding, state, and CheXbert class has no derangement over patients when it has a single report finding or when one patient has more than half of its report findings. A report finding of such a group takes the report of a report finding of another patient with the same finding and state, drawn at random with a generator seeded by 13, the dataset, and the group. On MIMIC-CXR, 1,200 report findings received a report of their class through a derangement, and one report finding received a report by this rule. On CheXpert Plus, all 1,669 report findings received a report of their class.

\refstepcounter{snote}
\section*{Supplementary Note \thesnote: Inputs of the verifiers, judges, writer, and report generators}
\label{snote:inputs}

\subsection*{Inputs of the verifiers}

A trained transformer verifier receives the template
\begin{verbatim}
[TASK] support_verification

[CASE_BEGIN]
<report>
[CASE_END]

[EVIDENCE_BEGIN]
<evidence>
[EVIDENCE_END]

[CLAIM_BEGIN]
<statement>
[CLAIM_END]
\end{verbatim}
\noindent in which the tags CASE and CLAIM mark the report and the statement.

The zero-shot large language models of Supplementary Table~\ref{stab:models} and Qwen3-8B~\cite{yang2025qwen3} finetuned with low-rank adaptation receive the following user message.
{\footnotesize
\begin{verbatim}
You check statements about a patient's chest radiograph. A statement is supported
only if the patient's report agrees with it and the cited evidence supports it.

Patient's report:
<report>

Cited evidence:
<evidence sentences, one per line, or None>

Statement: <statement>

Is the statement supported? Answer with Yes or No only.
\end{verbatim}
}

For every public verifier of Supplementary Table~\ref{stab:models} except GREEN~\cite{ostmeier2024green}, the document is the text ``Patient's report:'', a line break, the findings section, a blank line, ``Cited evidence:'', a line break, and the evidence sentences one per line, or the word None under removed evidence. A sentence of a document or a statement ends at a period, a question mark, or an exclamation mark followed by white space.

MiniCheck-Flan-T5-Large~\cite{tang2024minicheck} receives ``predict: '', a document chunk, the end-of-sequence token, and the statement. The document is split into chunks of at most 500 words at sentence boundaries. The score is the highest probability of support over the chunks, taken at the first decoder step.

Bespoke-MiniCheck-7B~\cite{bespokelabs2024minicheck} receives the system prompt of the MiniCheck release and a user message with the lines ``Document: $\langle$document$\rangle$'' and ``Claim: $\langle$statement$\rangle$''. Both messages are formatted by its chat template. The score is the probability of the token yes among the five most probable first tokens of its answer.

AlignScore-large~\cite{zha2023alignscore} receives a chunk of the document as the premise and one sentence of the statement. Each pair is padded to 512 tokens. In a longer pair, only the premise is truncated. The document is split into groups of sentences of about 350 words. The score is the alignment probability, taken as the highest value over the chunks and averaged over the sentences of the statement.

HHEM-2.1-Open~\cite{li2024hhem} receives the document as the premise and the statement as the hypothesis through the prediction function of its release. The score is the probability of consistency returned by this function.

LettuceDetect~\cite{kovacs2025lettucedetect} receives its question-answering prompt with the report and the evidence as two passages, the question ``Is this statement about the patient supported?'', and the statement as the answer. The score is 1 minus the highest hallucination probability over the tokens of the answer.

Lynx-8B~\cite{ravi2024lynx} receives its released prompt with the question above, the document, and the statement as the answer. The score is $P(\mathrm{PASS}) / (P(\mathrm{PASS}) + P(\mathrm{FAIL}))$ at the first token after the key SCORE of its JSON answer that contains a character other than white space, a quotation mark, or a colon. Every answer of Lynx-8B had PASS or FAIL among the returned tokens at this position.

GREEN receives its released prompt in its chat template, with the report as the reference report and the statement as the candidate report. The score is the number of matched findings divided by the sum of the matched findings and the clinically significant errors. Both counts are parsed from its answer as its release parses them.

\subsection*{Messages of the judges, the writer, and the report generators}

The judges gpt-oss-120b~\cite{openai2025gptoss} and Qwen3.5-122B-A10B~\cite{qwen2026qwen35} receive the system message
{\footnotesize
\begin{verbatim}
You are a radiology evidence auditor. You are given a concept, a claim about that
concept, and one or more external evidence sentences. Apply exactly two judgments.
First, does the evidence CONCERN the claim's concept? Second, if it does, does it
point toward the claim's state (present or absent), the opposite state, or neither?
Judge only what the evidence says. The claim describes a finding on a patient's chest
radiograph, so evidence about the same finding outside the chest does not concern the
claim's concept. Answer with a single JSON object and nothing else:
{"concerns_concept": "yes" | "no", "direction": "claim" | "opposite" | "neither"}.
\end{verbatim}
}
\noindent and the user message
{\footnotesize
\begin{verbatim}
Concept: "<finding>"
Claim: "<finding> is present"

Evidence:
"""
<sentence>
"""
\end{verbatim}
}
\noindent with at most 4{,}096 output tokens. Qwen3.5-122B-A10B answered with thinking disabled in its chat template. In an answer, the direction claim points toward the state present, and the direction opposite points toward the state absent. Each item was sent to a judge at most three times until its answer parsed. Every answer of both judges parsed, on the snapshot sentences and on the written candidates. The sentences assigned by both judges to each finding state are counted in Supplementary Fig.~\ref{sfig:support}.

The writer Qwen3.5-35B-A3B~\cite{qwen2026qwen35} receives the system message ``You write sentences for a radiology reference text about chest radiography, in the style of an encyclopedia article: general statements about a finding, never about a particular patient.'' and one of three requests. The request for support sentences of the present state is ``Write 30 different sentences, one per line, that each state or describe that $\langle$finding$\rangle$ is present on a chest radiograph, as a reference article would.'', and for the absent state ``Write 30 different sentences, one per line, that each describe the absence of $\langle$finding$\rangle$ on a chest radiograph: its exclusion, a normal appearance, or a sign that rules it out, as a reference article would.'', each followed, where the finding state has judge-accepted sentences, by ``Examples of such sentences from the reference text:'' and up to five of them. The request for unsupporting sentences is ``Write 30 different sentences, one per line, that mention $\langle$finding$\rangle$ without saying whether it is present or absent on a chest radiograph, for example its history, causes, epidemiology, a differential list or an imaging technique.'' The requests for support and for unsupporting sentences end with ``Vary the sentence structure and the vocabulary. Output only the sentences, one per line, without numbering.'' The request for rewrites is ``Rewrite the following sentence in 3 different ways that keep its meaning and its style, one per line. Output only the rewritten sentences, without numbering.'', followed by the sentence. Each request was sent with a recorded seed. When the answer to a request contained no candidate, the request was sent again with the next seed, at most three times in all. The writer generated at most 2{,}000 output tokens per request, with thinking disabled in its chat template.

Each report generator except MAIRA-2~\cite{bannur2024maira2} receives the frontal image of an Open-I~\cite{demnerfushman2016preparing} report and the prompt ``You are an experienced radiologist. Write the FINDINGS section of a chest X-ray report for the image provided. Write only the FINDINGS section, in concise clinical radiology style. Do not write an IMPRESSION section, a heading, or any text besides the findings themselves.'' MAIRA-2 receives the image as its current frontal view with no prior study, no indication, and no prompt. MedGemma-1.5-4B~\cite{sellergren2025medgemma}, LLaVA-Med~\cite{li2023llavamed}, Gemma-4-31B~\cite{gemma2026gemma4}, and Qwen3-VL-32B~\cite{qwen2025qwen3vl} receive the image as a JPEG.

\refstepcounter{snote}
\section*{Supplementary Note \thesnote: Patient pairs and the attainable AUROC}
\label{snote:proofs}

Let $h$ be any function of the evidence and the statement, and let $z_i = h(e_i, y_i)$. Every pair $(i, j)$ of the set $\Pi$ of Eq.~\ref{eq:pairs} has $e_i = e_j$ and $y_i = y_j$, so $z_i = z_j$. Therefore, each pair adds one half to the sum of Eq.~\ref{eq:pairs}, and the patient pair accuracy is one half (Eq.~\ref{eq:half}). The identity holds for any $h$ on any test set. The evidence tests replace or remove only $e$. Therefore, a verifier with $z_i = h(e_i, y_i)$ can change its scores under either test without receiving the report.

Fix an input channel $x$ and a set of examples, and let the examples with equal $x$ form the groups $G_1, \dots, G_m$. Group $G_k$ contains $n_k^{+}$ supported and $n_k^{-}$ unsupported examples, with $N^{+} = \sum_k n_k^{+}$ and $N^{-} = \sum_k n_k^{-}$. A verifier that receives only $x$ assigns one score $z_k$ to all examples of $G_k$, and the AUROC of these scores, with ties counted as one half, is
\begin{equation}\label{eq:sauroc}
\mathrm{AUROC}(z) = \frac{1}{N^{+} N^{-}} \Big( \sum_{k} \tfrac{1}{2}\, n_k^{+} n_k^{-} + \sum_{k < l} \phi_{kl}(z) \Big), \qquad \phi_{kl}(z) = \begin{cases} n_k^{+} n_l^{-}, & z_k > z_l, \\ n_l^{+} n_k^{-}, & z_k < z_l, \\ \tfrac{1}{2}\,(n_k^{+} n_l^{-} + n_l^{+} n_k^{-}), & z_k = z_l. \end{cases}
\end{equation}
The first sum does not depend on $z$. Each term $\phi_{kl}(z)$ depends only on the order of $z_k$ and $z_l$ and is at most $\max(n_k^{+} n_l^{-}, n_l^{+} n_k^{-})$. With $r_k = n_k^{+} / (n_k^{+} + n_k^{-})$, the inequality $n_k^{+} n_l^{-} \ge n_l^{+} n_k^{-}$ holds if and only if $r_k \ge r_l$, and the two products are equal when $r_k = r_l$. Therefore, scoring every group by $r_k$ attains the maximum of every term of Eq.~\ref{eq:sauroc} simultaneously. These scores are the scores $\hat{p}_x$ of Eq.~\ref{eq:bound}. So the value $B_x$ is the highest AUROC attainable by a verifier that receives only $x$ on this set.

Take a set whose labels are given by Eq.~\ref{eq:label} and the channel $x = (e, y)$. When a group contains a supported example, its shared value of $\sigma(e, y)$ is 1. Each unsupported example of the group then has $\lambda = 0$ and is a contradicted example with the same evidence and statement as the supported examples. Therefore, the groups with both labels are the groups that contain patient pairs. When no group contains both labels, every $r_k$ is 0 or 1 and $B_{(e, y)} = 1$. When a group contains both labels, its supported and unsupported examples tie and $B_{(e, y)} < 1$. So the AUROC attainable without the report is below 1 if and only if the set contains patient pairs.

\refstepcounter{snote}
\section*{Supplementary Note \thesnote: Packets and reading instructions of the reader study}
\label{snote:reader}

In packet A, the 38 written training sentences were drawn in proportion to the number of written training sentences of each finding state, with at least one per state. Of the 50 combinations outside the candidates assigned to a state by at least one judge, 10 are assigned to the same state by both judges and have a sentence that is not a chest sentence, 3 are assigned to opposite states, 22 are assigned by gpt-oss-120b alone, and 15 are assigned by Qwen3.5-122B-A10B alone. The combinations of each group, and the 50 combinations assigned to a state by neither judge, were taken in turn over the findings with distinct sentences. In packet B, 6 of the 30 PadChest-GR report findings have a release label that disagrees with their CheXbert class. The other report findings of each set were taken in turn over the findings and over the three slots present with the class positive, absent with the class negative, and absent with the class blank, with a study taken at most once per set. In packet C, the statements of a generator were taken in turn over the findings, with distinct sentences and distinct reports preferred. The 13 contradicted statements of each generator are allocated over the kinds of contradiction in proportion to their counts, with at least one per kind. In packet D, the sentences of each group were taken in turn over the findings, with every study at most once in the packet. From the sentences on which the report and the ReXErr definitions disagree, 5 false predictions, 5 false negations, and 10 unchanged sentences were taken.

The instructions asked the radiologists to answer from radiology knowledge and from the text of each item, with no external reference such as the literature, a colleague, or a picture archiving and communication system. The radiologists were also asked to treat every item as independent. In packet A, a sentence did not have to mention the radiograph, and some sentences are fragments of longer passages. The radiologists were asked to judge what each sentence says about the finding named in the item. The answer ``neither'' applies, for example, when the sentence is ``about another finding, about the finding in another organ, or too unspecific''. In packet B, the answer ``absent'' applies when the report states that the finding is ``absent, not seen, or resolved''. The answer ``uncertain'' applies when the report ``calls the finding possible, questionable, or not excluded''. A report may mention the finding under another name, ``for example an enlarged cardiac silhouette for cardiomegaly or fluid in the pleural space for a pleural effusion''. In packets C and D, the answer ``not addressed'' applies when the report ``says nothing about the finding, so it neither agrees nor contradicts''. The radiologists were asked to judge only the finding named in the item, even when the sentence mentions other things, and to judge ``the state of the finding and not the wording or the location''.

The answers were converted to labels by the labeling convention of the report findings. In packet B, the answer not mentioned counts as absent, and a report finding answered uncertain is left out of the comparisons with that radiologist. In packets C and D, a statement read as agreeing with the report counts as consistent and a statement read as contradicting it counts as contradicted. A statement read as not addressed counts as consistent if it says that the finding is absent and as contradicted otherwise.

\clearpage

\begin{algorithm}[t]
\caption{Construction of the examples of BRACE}
\label{salg:construction}
\begin{algorithmic}[1]
\Require report findings $(c, f, s)$ with a report $c$, a finding $f$, and a state $s \in \{\text{present}, \text{absent}\}$; chest sentences $U$ of the reference text; support pools $P_f^{s}$ of each finding state; written unsupporting sentences $W_f$ of each finding, possibly empty; a predicate $m_f(u)$ that is true when sentence $u$ names $f$ and $m(u)$ when it names any finding; a predicate $j_f(u)$ that is true when a judge assigns $u$ to a state of $f$; evidence size $n$; a share $\mu$ of mixed evidence, possibly 0; random seed
\Ensure examples $(c, e, y, \ell)$ with evidence $e$, statement $y$, and label $\ell$
\For{each finding $f$}
\State $H_f \gets \{u \in U \setminus (\bigcup_{f',s'} P_{f'}^{s'}) : m_f(u) \text{ and not } j_f(u)\} \cup W_f$ \Comment{hard unsupporting sentences}
\EndFor
\State $E \gets \{u \in U : \text{not } m(u)\}$ \Comment{easy unsupporting sentences}
\For{each report finding $(c, f, s)$}
\State $\bar{s} \gets$ the opposite state of $s$
\If{$P_f^{s} \neq \varnothing$}
\State add $(c, \textsc{Draw}(P_f^{s}), \text{``}f \text{ is } s\text{''}, 1)$ \Comment{supported example}
\EndIf
\If{$P_f^{\bar{s}} \neq \varnothing$}
\State add $(c, \textsc{Draw}(P_f^{\bar{s}}), \text{``}f \text{ is } \bar{s}\text{''}, 0)$ \Comment{contradicted example}
\EndIf
\State add $(c, \textsc{Fill}(H_f, E), \text{``}f \text{ is present''}, 0)$ \Comment{without supporting evidence}
\State add $(c, \textsc{Fill}(E, H_f), \text{``}f \text{ is absent''}, 0)$ \Comment{without supporting evidence}
\EndFor
\For{each supported or contradicted example, with probability $\mu$} \Comment{mixed evidence}
\State replace its evidence by its first sentence and one sentence of $H_f$, in random order
\EndFor
\Statex $\textsc{Draw}(S)$: $\min(n, |S|)$ sentences of $S$ drawn without replacement and sorted by a fixed index of the sentences
\Statex $\textsc{Fill}(X, Y)$: $\textsc{Draw}(X)$, completed with sentences drawn from $Y$ until it has $n$ sentences, sorted by the same index
\end{algorithmic}
\end{algorithm}

\begin{algorithm}[t]
\caption{Patient-level checks of a grounding verifier on one test set}
\label{salg:checks}
\begin{algorithmic}[1]
\Require examples $i = 1, \dots, n$ of the test set with evidence $e_i$, statement $y_i$, type (supported, contradicted, or without support), label $\ell_i$, and verifier score $z_i$; the scores $\mathcal{T}$ of the supported examples used to set the threshold; the share $q$ of these examples to accept
\Ensure patient pair accuracy $\mathrm{PA}$, accepted share $a_q$ of the contradicted examples, and attainable AUROC $B_{(e, y)}$
\State group the examples by $(e_i, y_i)$ into $G_1, \dots, G_m$
\For{$k = 1, \dots, m$}
\State $r_k \gets$ mean of $\ell_i$ over $i \in G_k$
\EndFor
\State $B_{(e, y)} \gets$ AUROC of the scores $r_{k(i)}$ for the labels $\ell_i$, with ties counted as one half
\State $N \gets 0$, $D \gets 0$
\For{$k = 1, \dots, m$}
\For{every supported $i \in G_k$ and every contradicted $j \in G_k$}
\State $N \gets N + \mathbf{1}[z_i > z_j] + \tfrac{1}{2}\,\mathbf{1}[z_i = z_j]$
\State $D \gets D + 1$
\EndFor
\EndFor
\State $\mathrm{PA} \gets N / D$, undefined when $D = 0$
\State $\tau_q \gets$ the largest score $t$ such that at least the share $q$ of the scores in $\mathcal{T}$ is at least $t$
\State $a_q \gets$ share of the contradicted examples with $z_j \ge \tau_q$
\State \Return $\mathrm{PA}$, $a_q$, $B_{(e, y)}$
\end{algorithmic}
\end{algorithm}

\clearpage

\begin{figure*}[p]
\centering
\includegraphics[width=\textwidth]{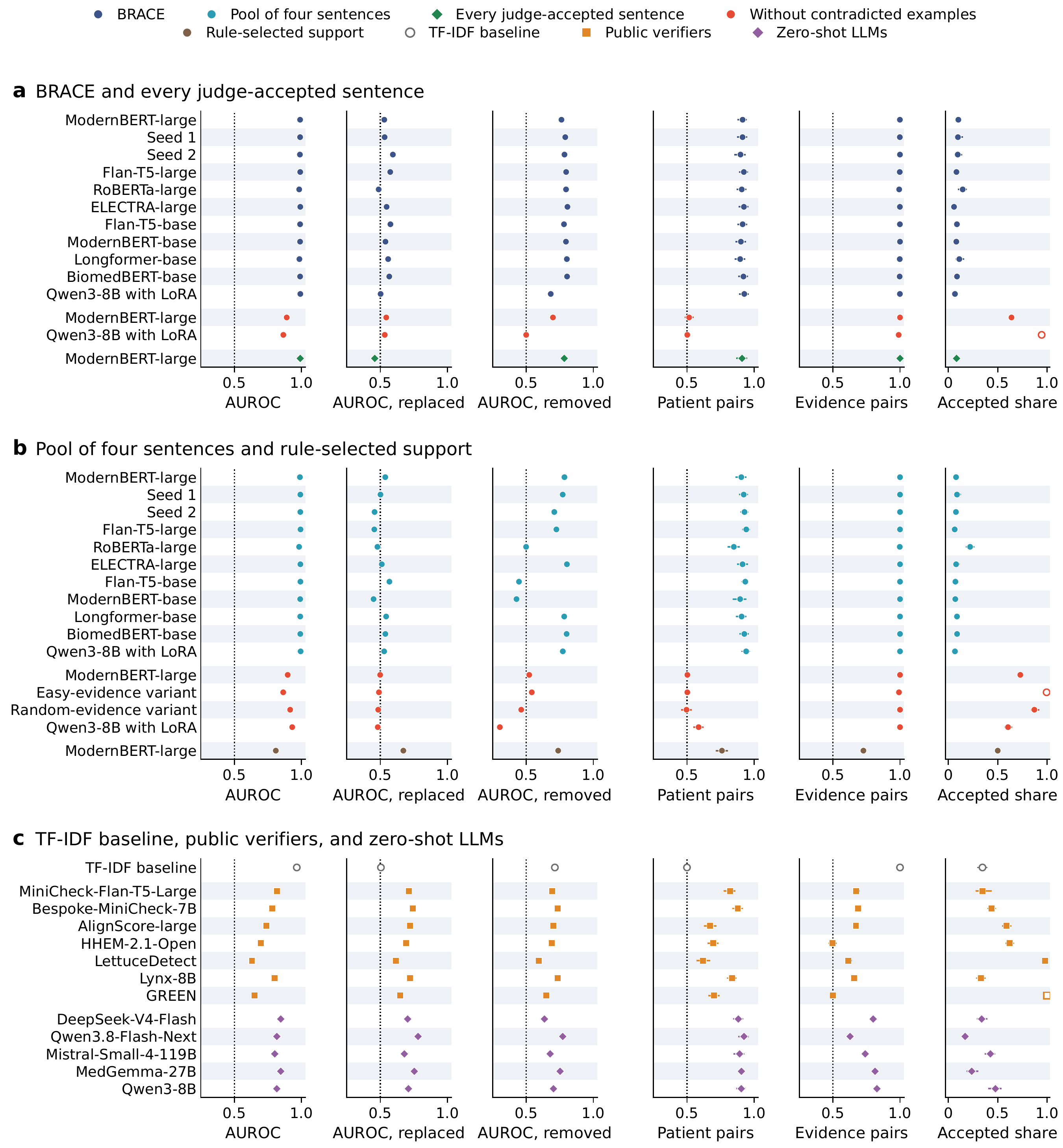}
\caption{Standard validation, evidence tests, and patient-level checks of the 43 verifiers on the MIMIC-CXR~\cite{johnson2019mimic} test set ($n = 4{,}356$ examples). Color and marker shape mark the training evidence or the type of verifier. Light bands mark every second verifier. Markers are bootstrap means over 1,000 resamples of the patients. Error bars are 95\% confidence intervals (CIs) from the 2.5th to the 97.5th percentile. Every panel has error bars. Counts per column for every verifier: 4,356 examples, 33,441 patient pairs, 1,109 evidence pairs, and 845 contradicted examples. \textbf{a}--\textbf{c}, One group of verifiers per row, named in its title. The columns are the area under the receiver operating characteristic curve (AUROC) with unchanged, replaced, and removed evidence; the patient and evidence pair accuracy; and the accepted share of the contradicted examples at $\tau_{0.90}$, the threshold at which the accepted share of the supported examples is 0.90. Open orange and red markers are the three verifiers that accept more than 0.95 of the supported examples at this threshold. Seed 1, Seed 2, and the easy-evidence and random-evidence variants are runs of ModernBERT-large~\cite{warner2024modernbert}. Dotted lines mark 0.5. LLM, large language model; LoRA, low-rank adaptation; TF-IDF, term frequency-inverse document frequency.}
\label{sfig:mimic}
\end{figure*}

\begin{figure*}[p]
\centering
\includegraphics[width=\textwidth]{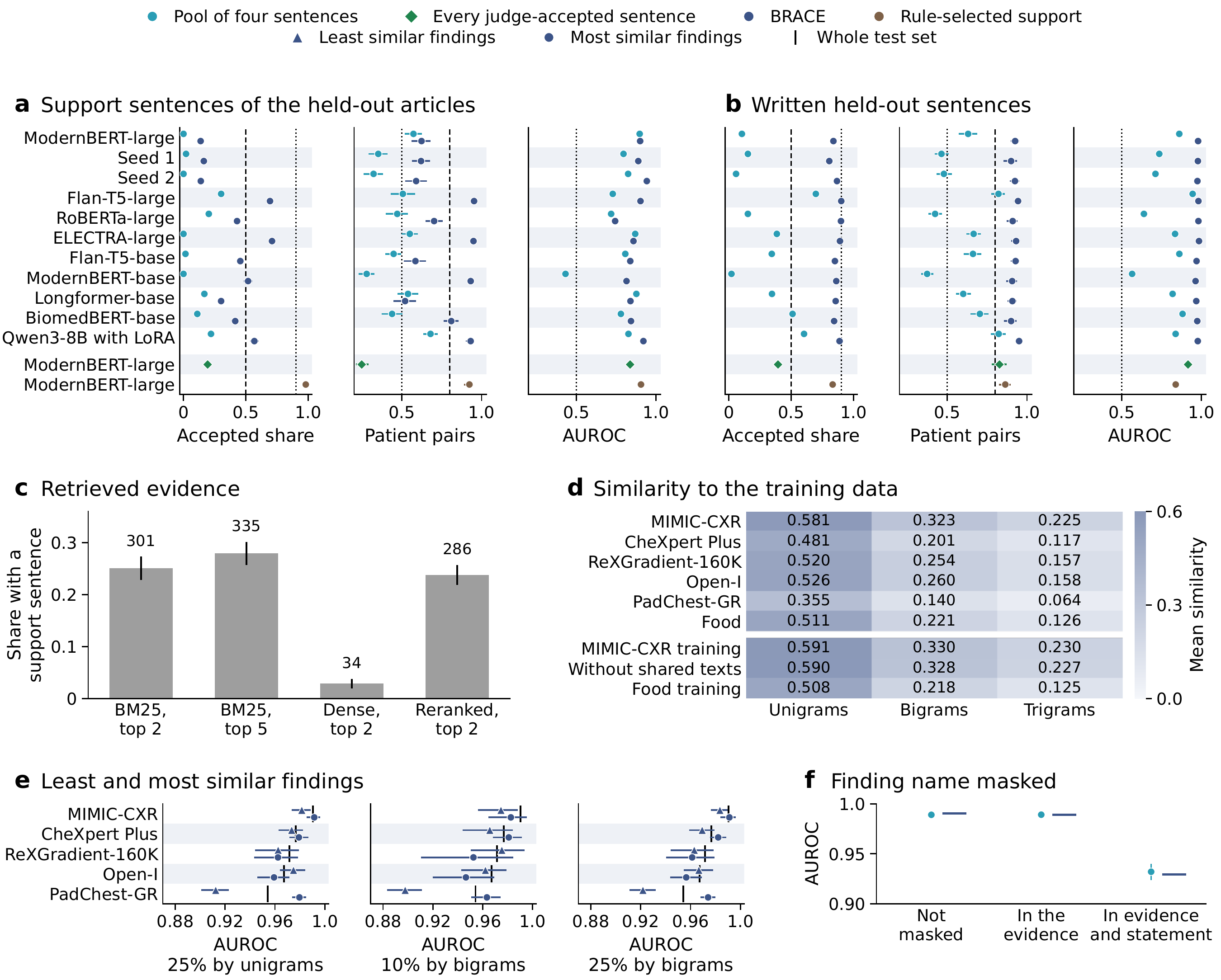}
\caption{Unseen support sentences and the controls, on the MIMIC-CXR test set unless a panel names another set. Color and marker shape mark the training evidence and the similarity subsets. Light bands mark every second row. Markers, ticks, and bars are bootstrap means over 1,000 resamples of the patients, or of the reports on Open-I~\cite{demnerfushman2016preparing}. Error bars are 95\% CIs from the 2.5th to the 97.5th percentile. Every panel except \textbf{d} has error bars. Counts per panel: 3,697 examples, 695 supported examples, and 29,907 patient pairs (\textbf{a}); 4,804, 1,201, and 4,078 (\textbf{b}); 1,201 combinations of a test report finding and its statement (\textbf{c}); 522 to 6,842 report findings and 773 to 3,915 recipe allergens per row and $n$-gram order (\textbf{d}); 195 to 2,320 examples per subset (\textbf{e}); and 4,356 examples (\textbf{f}). \textbf{a},\textbf{b}, Accepted share of the supported examples at $\tau_{0.90}$, patient pair accuracy, and AUROC, with the evidence named in the title. Dashed lines mark the criterion for unseen evidence. Dotted lines mark 0.90 and 0.5. \textbf{c}, Share of the combinations whose retrieved sentences include a support sentence, with the number of these combinations above each bar. The retrievers are BM25~\cite{robertson2009probabilistic}, the dense retriever Qwen3-Embedding-8B~\cite{zhang2025qwen3embedding}, and BM25 reranked by Qwen3-Reranker-8B. \textbf{d}, Mean cosine similarity of word $n$-gram count vectors~\cite{elangovan2021memorization} between each test report finding or recipe allergen and the nearest training findings section or recipe. In the training references of the last three rows, each training report finding or recipe allergen is compared with the findings sections of other patients or with other recipes, once without the findings sections shared by several patients. \textbf{e}, AUROC on the least and most similar findings of ModernBERT-large trained with BRACE. The 10\% by unigrams are plotted in Fig.~\ref{fig:robustness}f. The black tick marks the whole test set. \textbf{f}, AUROC of ModernBERT-large trained on the pool of four sentences, with and without the finding name masked. Navy ticks mark ModernBERT-large trained with BRACE (Fig.~\ref{fig:robustness}g).}
\label{sfig:unseen}
\end{figure*}

\begin{figure*}[p]
\centering
\includegraphics[width=\textwidth]{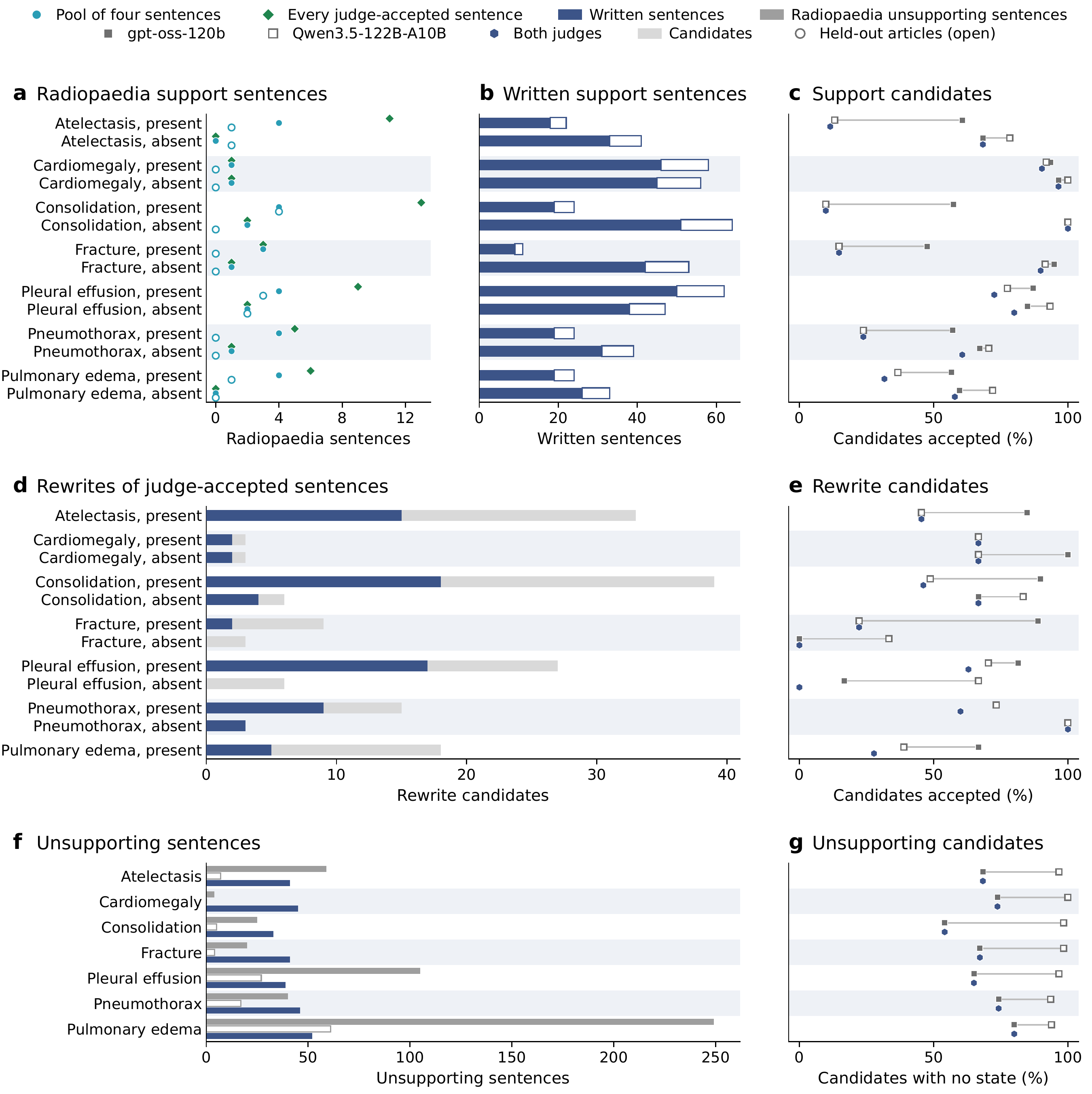}
\caption{Support and unsupporting sentences per finding state and the judgments of the written candidates ($n = 1{,}439$ written candidates). Cyan marks the pool of four sentences of Radiopaedia~\cite{gaillard2011radiopaedia}, and green marks every judge-accepted Radiopaedia sentence. Navy marks the written sentences accepted by both judges, and gray marks the unsupporting Radiopaedia sentences. Open markers and bars mark the held-out articles. Filled squares mark gpt-oss-120b~\cite{openai2025gptoss}, open squares mark Qwen3.5-122B-A10B~\cite{qwen2026qwen35}, and hexagons mark both judges. Light bands mark every second finding. No panel has error bars. Counts per panel: 55, 31, and 12 combinations of a sentence and a finding state (\textbf{a}); 446 training and 112 held-out written support sentences (\textbf{b}); 57 to 63 candidates per request (\textbf{c}); 3 to 39 rewrite candidates per finding state (\textbf{d}, \textbf{e}); 502 and 121 Radiopaedia combinations and 297 written sentences (\textbf{f}); and 60 to 65 candidates per finding (\textbf{g}). \textbf{a}, Every judge-accepted sentence (upper marker), the pool of four sentences (center), and the held-out pool (lower) of each finding state. \textbf{b}, Written support sentences kept by the leakage filter, for training (filled) and held out (open). \textbf{c},\textbf{e},\textbf{g}, Share of the candidates of a request assigned to the requested state by each judge (upper markers) and by both judges (lower marker), or to no state for an unsupporting request. \textbf{d}, Rewrite candidates of the judge-accepted training sentences (light gray) and the rewrites accepted by both judges (navy). \textbf{f}, Hard unsupporting sentences per finding from the training articles, the held-out articles, and the writer Qwen3.5-35B-A3B.}
\label{sfig:support}
\end{figure*}

\clearpage

\begin{table*}[t]
\centering
\caption{Glossary of the terms of the study. AI, artificial intelligence; AUROC, area under the receiver operating characteristic curve; LLM, large language model; TF-IDF, term frequency-inverse document frequency.}
\label{stab:glossary}
{\scriptsize\hyphenpenalty=10000\exhyphenpenalty=10000
\setlength{\tabcolsep}{4pt}
\renewcommand{\arraystretch}{0.9}
\begin{tabular}{@{}>{\raggedright\arraybackslash}p{0.24\textwidth}>{\raggedright\arraybackslash}p{0.735\textwidth}@{}}
\toprule
Term & Definition \\
\midrule
\multicolumn{2}{@{}l}{\textbf{Data and evidence}} \\
Report finding & A study with one of the seven findings and the state of the finding, present or absent. \\
Statement & The sentence ``$f$ is $s$'' for a finding $f$ and a state $s$, as in ``pleural effusion is present''. \\
Support sentence & A chest sentence that both LLM judges assign to the stated finding state. \\
Unsupporting sentence & A chest sentence that names the finding and is assigned to no state of it by either judge (hard), or a chest sentence that names none of the findings (easy). \\
Written sentence & A sentence written by the LLM writer in the style of the reference text and accepted by both judges. \\
\midrule
\multicolumn{2}{@{}l}{\textbf{BRACE and levels of training evidence}} \\
BRACE & Building Report And Citation Examples, the automated framework that builds training examples from reports with automatic labels and a reference text, with support sentences selected by two LLM judges and added by an LLM writer. \\
Pool of four sentences, level 1 & The first four support sentences of each finding state in the training articles. \\
Every judge-accepted sentence, level 2 & Every support sentence of the training articles. \\
Level 3 & The training evidence of BRACE, i.e., level 2 with the written training sentences and with mixed evidence, a support sentence beside a hard unsupporting sentence. \\
Comparison set & Level 1 with support sentences selected by a keyword rule in place of the judges, also called rule-selected support. \\
\midrule
\multicolumn{2}{@{}l}{\textbf{Example types}} \\
Supported example & The report agrees with the statement, and the evidence includes a support sentence for it. \\
Contradicted example & The report contradicts the statement, and the evidence includes a support sentence for it. \\
Example without supporting evidence & The evidence includes no support sentence for the statement. \\
\midrule
\multicolumn{2}{@{}l}{\textbf{Test conditions}} \\
Held-out sentences & Support sentences of the held-out articles, or held-out written support sentences. \\
Evidence tests & Replaced evidence from another example of the test set, and removed evidence. A verifier passes them when its AUROC under both is below its AUROC with unchanged evidence. \\
Replaced report & The findings section of another patient with the same state and CheXbert class. \\
Confirmed support & Support sentences read as supporting their assigned state by T.T.N., by S.Z., or by both radiologists. \\
\midrule
\multicolumn{2}{@{}l}{\textbf{AI-written statements}} \\
Consistent and contradicted statements & Statements labeled as agreeing with their report or contradicting it, under the report, the ReXErr, or the agree definition. \\
Template form & The statement written as ``$f$ is $s$'', in place of its original sentence. \\
Citations & A support sentence of the training or of the held-out articles, or the two sentences retrieved by BM25 from the held-out articles. \\
\midrule
\multicolumn{2}{@{}l}{\textbf{Verifiers}} \\
Trained with BRACE & The 11 verifiers trained on level 3. \\
Food verifier & ModernBERT-large trained with recipes and allergens in place of reports and findings. \\
Without contradicted examples & The six verifiers trained on the four training sets without contradicted examples. \\
TF-IDF baseline & A logistic regression on TF-IDF features of the statement and the evidence alone. \\
Public verifiers and zero-shot LLMs & Seven grounding verifiers from other studies and five LLMs, run without training. \\
\midrule
\multicolumn{2}{@{}l}{\textbf{Measures}} \\
Patient pair accuracy & The share of patient pairs in which the supported example scores higher, with a tie counted as one half. A patient pair, or a recipe pair in the food domain, is a supported and a contradicted example with the same evidence and statement. \\
Evidence pair accuracy & The same share for a supported example and the example without supporting evidence of its report finding. \\
Accepted share at $\tau_{0.90}$ & The share of examples or statements that score at or above the threshold at which a verifier's accepted share of its supported MIMIC-CXR test examples is 0.90. \\
AUROC attainable without the report & The highest AUROC of any function of the evidence and the statement on a set (Eq.~\ref{eq:bound}). \\
Criterion for unseen evidence & An accepted share of at least 0.50 for the supported examples with unseen support sentences, and a patient pair accuracy of at least 0.80. \\
Report error score & The share of the statements of a candidate report that score below $\tau_{0.90}$. \\
Kind of contradiction & Whether the impression of the report of a contradicted example states, negates, or does not mention the finding. \\
\midrule
\multicolumn{2}{@{}l}{\textbf{Reporting}} \\
Mean $\pm$ SD [95\% CI] & The mean, the standard deviation (SD), and the 95\% confidence interval (CI) from the 2.5th to the 97.5th percentile over 1{,}000 bootstrap resamples. \\
$p_{\mathrm{FDR}}$ & A $p$-value corrected with the Benjamini-Hochberg false discovery rate (FDR) within its family. A value below 0.05 is significant. \\
\bottomrule
\end{tabular}
}
\end{table*}

\begin{table*}[t]
\centering
\caption{Summary of the findings in the order in which the main text reports them, with the values taken unchanged from it. LoRA, low-rank adaptation; MeSH, Medical Subject Headings.}
\label{stab:findings}
{\scriptsize\hyphenpenalty=10000\exhyphenpenalty=10000
\setlength{\tabcolsep}{4pt}
\renewcommand{\arraystretch}{0.86}
\begin{tabular}{@{}>{\raggedright\arraybackslash}p{0.115\textwidth}>{\raggedright\arraybackslash}p{0.135\textwidth}>{\raggedright\arraybackslash}p{0.455\textwidth}>{\raggedright\arraybackslash}p{0.235\textwidth}@{}}
\toprule
Analysis & Comparison & Result & Interpretation \\
\midrule
Use of the report and the evidence & Verifier groups on MIMIC-CXR & For the verifiers trained with BRACE, patient pair accuracy of 0.897 to 0.927 and evidence pair accuracy of 0.994 to 1.000. For the zero-shot LLMs, patient pair accuracy of 0.883 to 0.925 and evidence pair accuracy of 0.627 to 0.828. TF-IDF baseline AUROC 0.966 without the report & The verifiers trained with BRACE use the report and the evidence. A high AUROC is no evidence of report use \\
\midrule
Other hospitals & Verifiers trained with BRACE & On the four other test sets, AUROC of 0.906 to 0.981, patient pair accuracy of 0.672 to 0.976, and evidence pair accuracy of 0.983 to 1.000 & At other hospitals, these verifiers also use the report and the evidence \\
\midrule
Without contradicted examples & Six verifiers & Patient pair accuracy of 0.496 to 0.587 on MIMIC-CXR and 0.452 to 0.639 on the other four test sets, below every verifier trained with BRACE on every test set & They pass the evidence tests and make little use of the report \\
\midrule
Memorization & Verifiers trained on the pool of four sentences & Accepted share of the supported MIMIC-CXR examples of 0.000 to 0.302 with the support sentences of the held-out articles and 0.021 to 0.697 with the written held-out sentences & These verifiers reject most support sentences absent from their training data \\
\midrule
Unseen support sentences & Verifiers trained with BRACE & All 11 meet the criterion on the written held-out sentences (accepted share 0.805 to 0.901), and 4 of the 11 meet it on the support sentences of the held-out articles (0.137 to 0.709) & Each verifier accepts more support sentences of the held-out articles than after training on the pool \\
\midrule
Statements, training citation & Contradicted ReXErr statements & Mean accepted share at $\tau_{0.90}$ of 0.239 for the verifiers trained with BRACE and 0.830 for four verifiers trained without contradicted examples & Without contradicted examples, verifiers accept most of these statements \\
\midrule
Statements, retrieved citation & Agreeing report, unsupporting citation & Mean accepted share of 0.169 for the verifiers trained with BRACE and 0.830 for the zero-shot LLMs. Accepted share of 0.598 to 0.997 for the six public verifiers other than GREEN & Unlike the LLMs, the verifiers trained with BRACE also check the citation \\
\midrule
Finetuned LLM & Qwen3-8B with LoRA and its zero-shot version & AUROC higher on every test set ($+0.090$ to $+0.176$). With an agreeing report and an unsupporting citation, accepted share of 0.123 for the finetuned and 0.878 for the zero-shot version & Finetuned on the constructed examples, Qwen3-8B also checks the cited evidence \\
\midrule
Correlation with acceptance & 43 verifiers & Spearman $\rho$ with the accepted share of the contradicted ReXErr statements of $-0.829$ for the patient pair accuracy and $-0.699$ for the AUROC. With the training citation, $|\rho|$ is larger for the patient pair accuracy in all 96 variants & Patient pair accuracy measures whether a verifier distinguishes examples with identical statements and citations but different report states \\
\midrule
Report error score & Counts of clinically significant errors & Kendall's $\tau_b$ of 0.415 to 0.574 for four verifiers trained with BRACE on RadEvalX ($p_{\mathrm{FDR}} \leq 0.013$). On ReXVal, $\tau_b$ ranges from $-0.186$ to 0.279 across all 11 verifiers trained with BRACE (all $p_{\mathrm{FDR}} > 0.05$) & Some verifiers reject more statements of RadEvalX reports with more errors \\
\midrule
Replaced report & Two verifier groups & Share of changed decisions over all test examples of MIMIC-CXR and CheXpert Plus of 0.050 to 0.104 for the verifiers trained with BRACE and 0.190 to 0.278 for the LLMs & Over all test examples, every LLM changes more decisions than every verifier trained with BRACE \\
\midrule
Label sources & CheXbert and three other label sources & Cohen's $\kappa$ of 0.920 to 0.995 with the CheXpert labeler, 0.402 to 0.892 with the MeSH topics of Open-I, and 0.726 to 0.989 for six findings and 0.147 for pulmonary edema with the PadChest-GR labels. Patient pair accuracies under the two label sources of Open-I correlated at $\rho = 0.981$ & The ranking of the verifiers is nearly the same under the two label sources of Open-I \\
\midrule
Reader study & Automatic labels and radiologists & Agreement of 0.964 with T.T.N. ($\kappa = 0.927$), 0.957 with S.Z. ($\kappa = 0.910$), and 0.977 between the radiologists ($\kappa = 0.952$) & Each radiologist agrees with the labels of BRACE almost as closely as with the other radiologist \\
\midrule
Second domain & Food verifier and TF-IDF baseline & Recipe pair accuracy of 0.792 for the food verifier. Accepted share of the contradicted examples of 0.493 for the food verifier and 0.898 for the TF-IDF baseline & BRACE extends beyond clinical applications \\
\bottomrule
\end{tabular}
}
\end{table*}

\begin{table*}[p]
\centering
\caption{Verifiers of the study. The first two blocks list the verifiers trained on the constructed examples, with their training sets, seeds, parameters, and input lengths in tokens. Level 3 is the training evidence of BRACE. The third block lists the baseline without the report, with its features and the inverse regularization strength $C$ chosen by the validation AUROC. The fourth block lists the public verifiers, with the mapping of the report, the evidence, and the statement to their inputs, and the score of each. The fifth block lists the LLM verifiers prompted zero-shot, with their sizes and settings. The score of each LLM verifier is $P(\mathrm{Yes}) / (P(\mathrm{Yes}) + P(\mathrm{No}))$ at the first answer token.}
\label{stab:models}
{\scriptsize\hyphenpenalty=10000\exhyphenpenalty=10000
\setlength{\tabcolsep}{3pt}
\renewcommand{\arraystretch}{0.98}
\begin{tabular}{@{}>{\raggedright\arraybackslash}p{0.25\textwidth}>{\raggedright\arraybackslash}p{0.37\textwidth}>{\raggedright\arraybackslash}p{0.34\textwidth}@{}}
\toprule
\addlinespace[2.5pt]
\multicolumn{3}{@{}l}{\textbf{Transformer verifiers trained on the constructed examples}} \\
Verifier & Training sets and seeds & Parameters and input length \\
\midrule
ModernBERT-large~\cite{warner2024modernbert} & Levels 1, 2, and 3, seeds 0, 1, and 2 at levels 1 and 3; without contradicted examples at level 1 (three variants) and at level 3; rule-selected support; food & 395.8 million, 1{,}024 tokens \\
Flan-T5-large~\cite{falnt5} & Levels 1 and 3, seed 0 & 751.3 million, 1{,}024 tokens \\
RoBERTa-large~\cite{liu2019roberta} & Levels 1 and 3, seed 0 & 355.4 million, 512 tokens \\
ELECTRA-large~\cite{clark2020electra} & Levels 1 and 3, seed 0 & 335.1 million, 512 tokens \\
Flan-T5-base~\cite{falnt5} & Levels 1 and 3, seed 0 & 223.5 million, 1{,}024 tokens \\
ModernBERT-base~\cite{warner2024modernbert} & Levels 1 and 3, seed 0 & 149.6 million, 1{,}024 tokens \\
Longformer-base~\cite{beltagy2020longformer} & Levels 1 and 3, seed 0 & 148.7 million, 1{,}024 tokens \\
BiomedBERT-base~\cite{biomedbert} & Levels 1 and 3, seed 0 & 109.5 million, 512 tokens \\
\midrule
\addlinespace[2.5pt]
\multicolumn{3}{@{}l}{\textbf{Open LLM finetuned with LoRA}} \\
Verifier & Training sets and seed & Parameters and input length \\
\midrule
Qwen3-8B with LoRA~\cite{yang2025qwen3,hu2022lora} & Levels 1 and 3, with and without contradicted examples, seed 0 & 8.2 billion, 1{,}024 tokens \\
\midrule
\addlinespace[2.5pt]
\multicolumn{3}{@{}l}{\textbf{Baseline without the report}} \\
Baseline & Model and training sets & Features and regularization \\
\midrule
TF-IDF baseline~\cite{pedregosa2011scikit} & Logistic regression on TF-IDF features of the statement and the evidence; level 1 and food & 39{,}978 unigrams and bigrams, $C = 1$ (radiology); 11{,}277, $C = 0.01$ (food) \\
\midrule
\addlinespace[2.5pt]
\multicolumn{3}{@{}l}{\textbf{Public verifiers with released checkpoints}} \\
Verifier & Inputs & Score \\
\midrule
MiniCheck-Flan-T5-Large~\cite{tang2024minicheck} & Document = report and evidence, claim = statement; chunks of 500 words, highest value & Probability of support \\
Bespoke-MiniCheck-7B~\cite{bespokelabs2024minicheck} & Document and claim as above, in the prompt of its release & Probability of yes among the five most probable first tokens \\
AlignScore-large~\cite{zha2023alignscore} & Context = report and evidence, claim = statement; chunks of 350 words, highest value, mean over the statement's sentences & Alignment probability \\
HHEM-2.1-Open~\cite{li2024hhem} & Premise = report and evidence, hypothesis = statement & Consistency score \\
LettuceDetect~\cite{kovacs2025lettucedetect} & Context = report and evidence as two passages, answer = statement & 1 minus the highest hallucination probability of an answer token \\
Lynx-8B~\cite{ravi2024lynx} & Document = report and evidence, answer = statement, in its released prompt & $P(\mathrm{PASS}) / (P(\mathrm{PASS}) + P(\mathrm{FAIL}))$ at the score token \\
GREEN~\cite{ostmeier2024green} & Reference = report, candidate = statement; the evidence is not used & Matched findings divided by matched findings plus clinically significant errors \\
\midrule
\addlinespace[2.5pt]
\multicolumn{3}{@{}l}{\textbf{Zero-shot LLM verifiers}} \\
Verifier & Size and setting &  \\
\midrule
DeepSeek-V4-Flash~\cite{deepseek2026v4} & 284 billion parameters, 13 billion activated; NVFP4 weights & \\
Qwen3.8-Flash-Next~\cite{qwen2026qwen38flashnext} & 125 billion parameters, 6 billion activated; NVFP4 weights; reasoning off & \\
Mistral-Small-4-119B~\cite{mistral2026small4} & 119 billion parameters, 6.5 billion activated; NVFP4 weights & \\
MedGemma-27B~\cite{sellergren2025medgemma} & 27 billion parameters & \\
Qwen3-8B~\cite{yang2025qwen3} & 8.2 billion parameters; bfloat16; thinking disabled & \\
\bottomrule
\end{tabular}
}
\end{table*}

\begin{table}[t]
\centering
\caption{Spearman correlation across the 43 verifiers between a test-set measure and the accepted share among negative labels. Each cell is the coefficient with its FDR-adjusted $p$-value in parentheses. The test-set measure is the patient pair accuracy or the AUROC on the MIMIC-CXR test set. The accepted share uses the threshold at which a verifier's accepted share of its supported MIMIC-CXR test examples or of the consistent statements of the source is 0.90. ReXErr uses the agree definition, and the generators use the report definition. With the retrieved citation, the joint label is positive when the report agrees and a retrieved sentence supports the statement, and negative otherwise.}
\label{stab:correlations}
{\scriptsize\hyphenpenalty=10000\exhyphenpenalty=10000
\setlength{\tabcolsep}{3pt}
\renewcommand{\arraystretch}{1.08}
\begin{tabular}{@{}lrrrr@{}}
\toprule
Statement source & \begin{tabular}[b]{@{}r@{}}Patient pairs,\\test-set threshold\end{tabular} & \begin{tabular}[b]{@{}r@{}}Patient pairs,\\statement threshold\end{tabular} & \begin{tabular}[b]{@{}r@{}}AUROC,\\test-set threshold\end{tabular} & \begin{tabular}[b]{@{}r@{}}AUROC,\\statement threshold\end{tabular} \\
\midrule
\addlinespace[2.5pt]
\multicolumn{5}{@{}l}{\textbf{Template statements}} \\
ReXErr & $-0.829$ ($< 0.001$) & $-0.733$ ($< 0.001$) & $-0.699$ ($< 0.001$) & $-0.447$ (0.007) \\
Generators, pooled & $-0.774$ ($< 0.001$) & $-0.582$ ($< 0.001$) & $-0.726$ ($< 0.001$) & $-0.496$ (0.002) \\
Generators, equal weight & $-0.774$ ($< 0.001$) & $-0.679$ ($< 0.001$) & $-0.648$ ($< 0.001$) & $-0.420$ (0.013) \\
Generators pooled, without LLaVA-Med & $-0.704$ ($< 0.001$) & $-0.458$ (0.006) & $-0.513$ (0.001) & $-0.334$ (0.061) \\
MAIRA-2 & $-0.785$ ($< 0.001$) & $-0.681$ ($< 0.001$) & $-0.571$ ($< 0.001$) & $-0.344$ (0.052) \\
LLaVA-Rad & $-0.761$ ($< 0.001$) & $-0.542$ ($< 0.001$) & $-0.583$ ($< 0.001$) & $-0.178$ (0.388) \\
MedGemma-1.5-4B & $-0.785$ ($< 0.001$) & $-0.614$ ($< 0.001$) & $-0.745$ ($< 0.001$) & $-0.351$ (0.047) \\
LLaVA-Med & $-0.801$ ($< 0.001$) & $-0.777$ ($< 0.001$) & $-0.778$ ($< 0.001$) & $-0.614$ ($< 0.001$) \\
Gemma-4-31B & $-0.837$ ($< 0.001$) & $-0.744$ ($< 0.001$) & $-0.738$ ($< 0.001$) & $-0.492$ (0.003) \\
Qwen3-VL-32B & $-0.349$ (0.049) & $-0.400$ (0.019) & $-0.141$ (0.515) & $-0.259$ (0.178) \\
ReXErr, held-out citation & $-0.271$ (0.154) & 0.100 (0.641) & $-0.708$ ($< 0.001$) & 0.548 ($< 0.001$) \\
ReXErr, retrieved citation & $-0.173$ (0.409) & $-0.239$ (0.219) & $-0.686$ ($< 0.001$) & $-0.230$ (0.241) \\
\addlinespace[2.5pt]
\multicolumn{5}{@{}l}{\textbf{Original sentences as statements}} \\
ReXErr & $-0.429$ (0.010) & $-0.535$ ($< 0.001$) & $-0.146$ (0.496) & $-0.068$ (0.752) \\
Generators, pooled & $-0.414$ (0.014) & $-0.566$ ($< 0.001$) & $-0.210$ (0.293) & $-0.432$ (0.010) \\
Generators, equal weight & $-0.397$ (0.020) & $-0.795$ ($< 0.001$) & $-0.138$ (0.526) & $-0.619$ ($< 0.001$) \\
Generators pooled, without LLaVA-Med & $-0.350$ (0.048) & $-0.496$ (0.002) & $-0.086$ (0.684) & $-0.318$ (0.078) \\
MAIRA-2 & $-0.507$ (0.002) & $-0.774$ ($< 0.001$) & $-0.255$ (0.185) & $-0.538$ ($< 0.001$) \\
LLaVA-Rad & $-0.310$ (0.088) & $-0.598$ ($< 0.001$) & 0.009 (0.968) & $-0.264$ (0.167) \\
MedGemma-1.5-4B & $-0.438$ (0.009) & $-0.451$ (0.007) & $-0.228$ (0.248) & $-0.098$ (0.647) \\
LLaVA-Med & $-0.475$ (0.004) & $-0.739$ ($< 0.001$) & $-0.341$ (0.055) & $-0.689$ ($< 0.001$) \\
Gemma-4-31B & $-0.701$ ($< 0.001$) & $-0.810$ ($< 0.001$) & $-0.462$ (0.005) & $-0.603$ ($< 0.001$) \\
Qwen3-VL-32B & $-0.217$ (0.276) & $-0.464$ (0.005) & $-0.029$ (0.892) & $-0.291$ (0.116) \\
ReXErr, held-out citation & $-0.267$ (0.161) & 0.028 (0.898) & $-0.580$ ($< 0.001$) & 0.559 ($< 0.001$) \\
ReXErr, retrieved citation & $-0.082$ (0.698) & $-0.439$ (0.009) & $-0.630$ ($< 0.001$) & $-0.504$ (0.002) \\
\addlinespace[2.5pt]
\multicolumn{5}{@{}l}{\textbf{Template ReXErr statements on confirmed support}} \\
Support confirmed by T.T.N. & $-0.825$ ($< 0.001$) & $-0.680$ ($< 0.001$) & $-0.765$ ($< 0.001$) & $-0.473$ (0.003) \\
Support confirmed by S.Z. & $-0.729$ ($< 0.001$) & $-0.779$ ($< 0.001$) & $-0.559$ ($< 0.001$) & $-0.510$ (0.001) \\
Support confirmed by both radiologists & $-0.746$ ($< 0.001$) & $-0.829$ ($< 0.001$) & $-0.733$ ($< 0.001$) & $-0.602$ ($< 0.001$) \\
\bottomrule
\end{tabular}
}
\end{table}

\begin{table}[t]
\centering
\caption{Kendall's $\tau_b$ between the report error score of each verifier and the radiologists' counts of clinically significant errors, on the candidate reports with statements of RadEvalX~\cite{calamida2023radiology} ($n = 33$ of 100) and ReXVal~\cite{yu2023evaluating} ($n = 68$ of 144). The report error score is the share of a candidate report's statements that score below $\tau_{0.90}$ of the verifier (Eq.~\ref{eq:operating}). Each cell is the coefficient with its FDR-adjusted $p$-value in parentheses, from a bootstrap over patients on ReXVal and reports on RadEvalX. Rows are grouped by training evidence and verifier type. The GREEN~\cite{ostmeier2024green} score of the whole candidate report has $\tau_b$ of $-0.319$ ($< 0.001$) on RadEvalX ($n = 100$) and $-0.626$ ($< 0.001$) on ReXVal ($n = 144$). The easy-evidence variant of ModernBERT-large and GREEN are left out, since their report error score is constant.}
\label{stab:report_metric}
{\scriptsize\hyphenpenalty=10000\exhyphenpenalty=10000
\setlength{\tabcolsep}{6pt}
\renewcommand{\arraystretch}{0.95}
\begin{tabular}{@{}lrr@{}}
\toprule
Verifier & RadEvalX & ReXVal \\
\midrule
\addlinespace[2.5pt]
\multicolumn{3}{@{}l}{\textbf{BRACE}} \\
ModernBERT-large~\cite{warner2024modernbert} & 0.003 ($> 0.999$) & 0.047 (0.823) \\
Flan-T5-large~\cite{falnt5} & 0.415 (0.012) & 0.225 (0.114) \\
RoBERTa-large~\cite{liu2019roberta} & 0.327 (0.061) & 0.058 (0.823) \\
ELECTRA-large~\cite{clark2020electra} & 0.232 (0.235) & 0.187 (0.184) \\
Flan-T5-base~\cite{falnt5} & 0.574 (0.010) & 0.042 (0.841) \\
ModernBERT-base~\cite{warner2024modernbert} & 0.437 (0.010) & 0.238 (0.080) \\
Longformer-base~\cite{beltagy2020longformer} & 0.069 (0.763) & $-0.186$ (0.523) \\
BiomedBERT-base~\cite{biomedbert} & 0.296 (0.080) & 0.279 (0.066) \\
ModernBERT-large (seed 1) & 0.008 (0.976) & $-0.102$ (0.746) \\
ModernBERT-large (seed 2) & 0.324 (0.062) & 0.079 (0.746) \\
Qwen3-8B with LoRA~\cite{yang2025qwen3,hu2022lora} & 0.448 (0.012) & 0.214 (0.170) \\
\addlinespace[2.5pt]
\multicolumn{3}{@{}l}{\textbf{BRACE without contradicted examples}} \\
ModernBERT-large & $-0.163$ (0.425) & $-0.356$ (0.124) \\
Qwen3-8B with LoRA & $-0.209$ (0.271) & $-0.356$ (0.124) \\
\addlinespace[2.5pt]
\multicolumn{3}{@{}l}{\textbf{Every judge-accepted sentence}} \\
ModernBERT-large & 0.515 (0.012) & 0.062 (0.823) \\
\addlinespace[2.5pt]
\multicolumn{3}{@{}l}{\textbf{Pool of four sentences}} \\
ModernBERT-large & 0.404 (0.010) & 0.058 (0.823) \\
Flan-T5-large & 0.450 (0.029) & 0.103 (0.714) \\
RoBERTa-large & 0.079 (0.746) & $-0.106$ (0.746) \\
ELECTRA-large & 0.464 (0.010) & 0.097 (0.714) \\
Flan-T5-base & 0.323 (0.045) & 0.202 (0.187) \\
ModernBERT-base & 0.327 (0.094) & 0.174 (0.216) \\
Longformer-base & 0.414 ($< 0.001$) & 0.223 (0.080) \\
BiomedBERT-base & 0.123 (0.573) & $-0.111$ (0.741) \\
ModernBERT-large (seed 1) & 0.466 (0.012) & 0.240 (0.055) \\
ModernBERT-large (seed 2) & 0.508 (0.012) & 0.222 (0.091) \\
Qwen3-8B with LoRA & 0.446 (0.010) & 0.213 (0.114) \\
\addlinespace[2.5pt]
\multicolumn{3}{@{}l}{\textbf{Pool of four sentences, without contradicted examples}} \\
ModernBERT-large & 0.051 (0.823) & $-0.352$ (0.170) \\
\begin{tabular}[t]{@{}l@{}}ModernBERT-large,\\random evidence\end{tabular} & 0.134 (0.625) & 0.000 ($> 0.999$) \\
Qwen3-8B with LoRA & $-0.390$ (0.035) & $-0.128$ (0.625) \\
\addlinespace[2.5pt]
\multicolumn{3}{@{}l}{\textbf{Rule-selected support}} \\
ModernBERT-large & 0.283 (0.094) & 0.198 (0.197) \\
\addlinespace[2.5pt]
\multicolumn{3}{@{}l}{\textbf{Without the report}} \\
TF-IDF baseline~\cite{pedregosa2011scikit} & $-0.061$ (0.788) & 0.253 (0.111) \\
\addlinespace[2.5pt]
\multicolumn{3}{@{}l}{\textbf{Public verifiers}} \\
MiniCheck-Flan-T5-Large~\cite{tang2024minicheck} & 0.222 (0.216) & 0.130 (0.498) \\
Bespoke-MiniCheck-7B~\cite{bespokelabs2024minicheck} & 0.011 (0.965) & 0.249 (0.123) \\
AlignScore-large~\cite{zha2023alignscore} & 0.222 (0.195) & 0.058 (0.823) \\
HHEM-2.1-Open~\cite{li2024hhem} & 0.140 (0.561) & 0.295 (0.083) \\
LettuceDetect~\cite{kovacs2025lettucedetect} & 0.223 (0.530) & $-0.080$ (0.814) \\
Lynx-8B~\cite{ravi2024lynx} & 0.054 (0.823) & 0.249 (0.045) \\
\addlinespace[2.5pt]
\multicolumn{3}{@{}l}{\textbf{Zero-shot LLMs}} \\
DeepSeek-V4-Flash~\cite{deepseek2026v4} & 0.172 (0.498) & 0.274 (0.076) \\
Qwen3.8-Flash-Next~\cite{qwen2026qwen38flashnext} & 0.365 (0.047) & 0.281 (0.035) \\
Mistral-Small-4-119B~\cite{mistral2026small4} & 0.252 (0.122) & 0.161 (0.268) \\
MedGemma-27B~\cite{sellergren2025medgemma} & 0.182 (0.195) & 0.250 (0.061) \\
Qwen3-8B & 0.167 (0.313) & 0.264 (0.055) \\
\bottomrule
\end{tabular}
}
\end{table}

\begin{table}[t]
\centering
\caption{Agreement between the radiologists and agreement of the LLM judges and of the automatic labels with each radiologist ($n = 728$ items, each read by both radiologists). Each share is the bootstrap mean $\pm$ SD [95\% CI] over 1{,}000 resamples of the items. In a support row, the share is that of the sentences read as supporting their assigned state. A judge that assigned no state answers neither. A report that does not mention the finding counts as absent. Report findings answered as uncertain and items left unanswered are left out. The test-set label of PadChest-GR is its release label. Every Cohen's $\kappa$ has an FDR-adjusted $p$-value below 0.001.}
\label{stab:reader}
{\scriptsize\hyphenpenalty=10000\exhyphenpenalty=10000
\setlength{\tabcolsep}{3pt}
\renewcommand{\arraystretch}{0.95}
\begin{tabular}{@{}lrrrrrr@{}}
\toprule
Comparison & $n$ & T.T.N. & $\kappa$ & $n$ & S.Z. & $\kappa$ \\
\midrule
\addlinespace[2.5pt]
\multicolumn{7}{@{}l}{\textbf{Agreement between the radiologists}} \\
Packet A, all combinations & 328 & \multicolumn{4}{c}{0.719 $\pm$ 0.025 [0.672, 0.768]} & 0.569 \\
Packet A, pools of four & 43 & \multicolumn{4}{c}{0.376 $\pm$ 0.071 [0.238, 0.512]} &  \\
Packet A, written sentences & 150 & \multicolumn{4}{c}{0.813 $\pm$ 0.031 [0.753, 0.873]} &  \\
Packet B, answers & 150 & \multicolumn{4}{c}{0.846 $\pm$ 0.030 [0.780, 0.900]} & 0.745 \\
Packet B, labels & 132 & \multicolumn{4}{c}{0.977 $\pm$ 0.013 [0.947, 1.000]} & 0.952 \\
Packet C, answers & 149 & \multicolumn{4}{c}{0.920 $\pm$ 0.021 [0.878, 0.959]} & 0.847 \\
Packet C, labels & 149 & \multicolumn{4}{c}{0.947 $\pm$ 0.018 [0.910, 0.980]} & 0.886 \\
Packet D, answers & 99 & \multicolumn{4}{c}{0.880 $\pm$ 0.034 [0.808, 0.939]} & 0.755 \\
Packet D, labels & 99 & \multicolumn{4}{c}{0.900 $\pm$ 0.032 [0.833, 0.958]} & 0.782 \\
\addlinespace[2.5pt]
\multicolumn{7}{@{}l}{\textbf{Support sentences confirmed by each radiologist (packet A)}} \\
Pool of four, training articles & 31 & 0.577 $\pm$ 0.084 [0.419, 0.742] & & 31 & 0.355 $\pm$ 0.088 [0.200, 0.533] & \\
Pool of four, held-out articles & 12 & 0.502 $\pm$ 0.151 [0.167, 0.750] & & 12 & 0.505 $\pm$ 0.146 [0.250, 0.833] & \\
Other judge-accepted & 24 & 0.460 $\pm$ 0.100 [0.291, 0.667] & & 24 & 0.292 $\pm$ 0.092 [0.125, 0.458] & \\
Written, training & 38 & 0.867 $\pm$ 0.055 [0.737, 0.974] & & 38 & 0.871 $\pm$ 0.055 [0.763, 0.974] & \\
Written, held-out & 112 & 0.822 $\pm$ 0.037 [0.750, 0.893] & & 112 & 0.849 $\pm$ 0.034 [0.777, 0.911] & \\
Rule-selected, comparison set & 13 & 0.238 $\pm$ 0.127 [0.000, 0.500] & & 13 & 0.159 $\pm$ 0.102 [0.000, 0.385] & \\
\addlinespace[2.5pt]
\multicolumn{7}{@{}l}{\textbf{Support sentences confirmed by both radiologists (packet A)}} \\
Pool of four, training articles & 31 & \multicolumn{4}{c}{0.191 $\pm$ 0.072 [0.063, 0.333]} & \\
Pool of four, held-out articles & 12 & \multicolumn{4}{c}{0.167 $\pm$ 0.106 [0.000, 0.417]} & \\
Pool of four, all articles & 43 & \multicolumn{4}{c}{0.189 $\pm$ 0.061 [0.071, 0.310]} & \\
Other judge-accepted & 24 & \multicolumn{4}{c}{0.169 $\pm$ 0.077 [0.042, 0.333]} & \\
Written, training & 38 & \multicolumn{4}{c}{0.738 $\pm$ 0.072 [0.579, 0.868]} & \\
Written, held-out & 112 & \multicolumn{4}{c}{0.760 $\pm$ 0.040 [0.679, 0.839]} & \\
Written, all articles & 150 & \multicolumn{4}{c}{0.753 $\pm$ 0.035 [0.680, 0.820]} & \\
Rule-selected, comparison set & 13 & \multicolumn{4}{c}{0.159 $\pm$ 0.102 [0.000, 0.385]} & \\
\addlinespace[2.5pt]
\multicolumn{7}{@{}l}{\textbf{LLM judges (packet A)}} \\
gpt-oss-120b & 328 & 0.735 $\pm$ 0.026 [0.687, 0.785] & 0.607 & 328 & 0.704 $\pm$ 0.027 [0.650, 0.756] & 0.570 \\
Qwen3.5-122B-A10B & 328 & 0.705 $\pm$ 0.026 [0.658, 0.755] & 0.561 & 328 & 0.695 $\pm$ 0.027 [0.643, 0.746] & 0.553 \\
Two judges together & 328 & 0.747 $\pm$ 0.025 [0.701, 0.795] & 0.621 & 328 & 0.732 $\pm$ 0.026 [0.682, 0.781] & 0.600 \\
\addlinespace[2.5pt]
\multicolumn{7}{@{}l}{\textbf{Report labels (packet B)}} \\
Test-set labels, all test sets & 141 & 0.964 $\pm$ 0.016 [0.929, 0.993] & 0.927 & 139 & 0.957 $\pm$ 0.017 [0.921, 0.986] & 0.910 \\
Test-set labels, MIMIC-CXR & 34 & 0.941 $\pm$ 0.040 [0.853, 1.000] &  & 32 & 0.937 $\pm$ 0.043 [0.844, 1.000] &  \\
Test-set labels, CheXpert Plus & 27 & 1.000 $\pm$ 0.000 [1.000, 1.000] &  & 26 & 0.885 $\pm$ 0.059 [0.769, 1.000] &  \\
Test-set labels, ReXGradient-160K & 24 & 1.000 $\pm$ 0.000 [1.000, 1.000] &  & 23 & 1.000 $\pm$ 0.000 [1.000, 1.000] &  \\
Test-set labels, Open-I & 29 & 1.000 $\pm$ 0.000 [1.000, 1.000] &  & 30 & 1.000 $\pm$ 0.000 [1.000, 1.000] &  \\
Test-set labels, PadChest-GR & 27 & 0.891 $\pm$ 0.060 [0.741, 1.000] &  & 28 & 0.965 $\pm$ 0.033 [0.893, 1.000] &  \\
CheXbert classes & 150 & 0.726 $\pm$ 0.036 [0.660, 0.800] & 0.589 & 150 & 0.733 $\pm$ 0.036 [0.660, 0.800] & 0.603 \\
\addlinespace[2.5pt]
\multicolumn{7}{@{}l}{\textbf{Generator statements, report definition (packet C)}} \\
All statements & 150 & 0.813 $\pm$ 0.031 [0.748, 0.870] & 0.630 & 149 & 0.812 $\pm$ 0.033 [0.745, 0.875] & 0.627 \\
Consistent & 72 & 0.945 $\pm$ 0.027 [0.889, 0.987] &  & 72 & 0.945 $\pm$ 0.027 [0.887, 0.987] &  \\
\addlinespace[2.5pt]
\multicolumn{7}{@{}l}{\textbf{Contradicted generator statements by the impression (packet C)}} \\
Impression negates the finding & 4 & 0.721 $\pm$ 0.251 [0.000, 1.000] &  & 4 & 0.499 $\pm$ 0.197 [0.000, 1.000] &  \\
Impression does not mention it & 69 & 0.681 $\pm$ 0.057 [0.565, 0.788] &  & 68 & 0.674 $\pm$ 0.058 [0.557, 0.784] &  \\
Impression states the finding & 5 & 0.792 $\pm$ 0.183 [0.400, 1.000] &  & 5 & 1.000 $\pm$ 0.000 [1.000, 1.000] &  \\
\addlinespace[2.5pt]
\multicolumn{7}{@{}l}{\textbf{ReXErr sentences (packet D)}} \\
Agree definition & 80 & 0.927 $\pm$ 0.029 [0.867, 0.976] & 0.850 & 79 & 0.911 $\pm$ 0.032 [0.846, 0.964] & 0.822 \\
Report definition & 100 & 0.800 $\pm$ 0.040 [0.722, 0.874] & 0.600 & 99 & 0.797 $\pm$ 0.040 [0.717, 0.872] & 0.595 \\
ReXErr definition & 100 & 0.880 $\pm$ 0.031 [0.814, 0.936] & 0.760 & 99 & 0.858 $\pm$ 0.035 [0.783, 0.920] & 0.716 \\
\addlinespace[2.5pt]
\multicolumn{7}{@{}l}{\textbf{ReXErr sentences where the two definitions differ (packet D)}} \\
Report definition & 20 & 0.299 $\pm$ 0.096 [0.105, 0.500] &  & 20 & 0.350 $\pm$ 0.100 [0.158, 0.550] &  \\
ReXErr definition & 20 & 0.701 $\pm$ 0.096 [0.500, 0.895] &  & 20 & 0.650 $\pm$ 0.100 [0.450, 0.842] &  \\
\bottomrule
\end{tabular}
}
\end{table}

\begin{table}[t]
\centering
\caption{Report findings and examples per finding and type. The columns list the present and absent findings, the supported and the contradicted examples, the examples without supporting evidence whose statement calls the finding present or absent, and the supported and contradicted examples with the support sentences of the held-out articles, which exist for 6 of the 14 finding states. The examples of the MIMIC-CXR training and validation sets are those of the pool of four sentences. The set with every judge-accepted sentence has the same counts. At level 3, every report finding has a supported and a contradicted example. Entries without held-out evidence in the training and validation sets are marked as not applicable (N/A).}
\label{stab:examples}
{\scriptsize\hyphenpenalty=10000\exhyphenpenalty=10000
\setlength{\tabcolsep}{3pt}
\renewcommand{\arraystretch}{0.95}
\begin{tabular}{@{}lrrrrrrrr@{}}
\toprule
Finding & \begin{tabular}[b]{@{}r@{}}Present\\findings\end{tabular} & \begin{tabular}[b]{@{}r@{}}Absent\\findings\end{tabular} & \begin{tabular}[b]{@{}r@{}}Supported\end{tabular} & \begin{tabular}[b]{@{}r@{}}Contradicted\end{tabular} & \begin{tabular}[b]{@{}r@{}}Without\\support,\\present\end{tabular} & \begin{tabular}[b]{@{}r@{}}Without\\support,\\absent\end{tabular} & \begin{tabular}[b]{@{}r@{}}Supported,\\held-out\end{tabular} & \begin{tabular}[b]{@{}r@{}}Contradicted,\\held-out\end{tabular} \\
\midrule
\addlinespace[2.5pt]
\multicolumn{9}{@{}l}{\textbf{MIMIC-CXR, training}} \\
Atelectasis & 1{,}008 & 64 & 1{,}008 & 64 & 1{,}072 & 1{,}072 & N/A & N/A \\
Cardiomegaly & 1{,}306 & 163 & 1{,}469 & 1{,}469 & 1{,}469 & 1{,}469 & N/A & N/A \\
Consolidation & 222 & 421 & 643 & 643 & 643 & 643 & N/A & N/A \\
Fracture & 121 & 119 & 240 & 240 & 240 & 240 & N/A & N/A \\
Pleural effusion & 1{,}391 & 236 & 1{,}627 & 1{,}627 & 1{,}627 & 1{,}627 & N/A & N/A \\
Pneumothorax & 100 & 82 & 182 & 182 & 182 & 182 & N/A & N/A \\
Pulmonary edema & 1{,}079 & 530 & 1{,}079 & 530 & 1{,}609 & 1{,}609 & N/A & N/A \\
\addlinespace[2.5pt]
\multicolumn{9}{@{}l}{\textbf{MIMIC-CXR, validation}} \\
Atelectasis & 110 & 7 & 110 & 7 & 117 & 117 & N/A & N/A \\
Cardiomegaly & 175 & 18 & 193 & 193 & 193 & 193 & N/A & N/A \\
Consolidation & 27 & 39 & 66 & 66 & 66 & 66 & N/A & N/A \\
Fracture & 10 & 15 & 25 & 25 & 25 & 25 & N/A & N/A \\
Pleural effusion & 150 & 21 & 171 & 171 & 171 & 171 & N/A & N/A \\
Pneumothorax & 10 & 11 & 21 & 21 & 21 & 21 & N/A & N/A \\
Pulmonary edema & 120 & 56 & 120 & 56 & 176 & 176 & N/A & N/A \\
\addlinespace[2.5pt]
\multicolumn{9}{@{}l}{\textbf{MIMIC-CXR, test}} \\
Atelectasis & 166 & 4 & 166 & 4 & 170 & 170 & 170 & 170 \\
Cardiomegaly & 257 & 32 & 289 & 289 & 289 & 289 & 0 & 0 \\
Consolidation & 49 & 56 & 105 & 105 & 105 & 105 & 49 & 56 \\
Fracture & 24 & 20 & 44 & 44 & 44 & 44 & 0 & 0 \\
Pleural effusion & 252 & 34 & 286 & 286 & 286 & 286 & 286 & 286 \\
Pneumothorax & 14 & 15 & 29 & 29 & 29 & 29 & 0 & 0 \\
Pulmonary edema & 190 & 88 & 190 & 88 & 278 & 278 & 190 & 88 \\
\addlinespace[2.5pt]
\multicolumn{9}{@{}l}{\textbf{CheXpert Plus}} \\
Atelectasis & 134 & 28 & 134 & 28 & 162 & 162 & 162 & 162 \\
Cardiomegaly & 147 & 22 & 169 & 169 & 169 & 169 & 0 & 0 \\
Consolidation & 84 & 220 & 304 & 304 & 304 & 304 & 84 & 220 \\
Fracture & 34 & 19 & 53 & 53 & 53 & 53 & 0 & 0 \\
Pleural effusion & 387 & 80 & 467 & 467 & 467 & 467 & 467 & 467 \\
Pneumothorax & 51 & 26 & 77 & 77 & 77 & 77 & 0 & 0 \\
Pulmonary edema & 304 & 133 & 304 & 133 & 437 & 437 & 304 & 133 \\
\bottomrule
\end{tabular}
}
\end{table}

\begin{table}[t]
\centering
\caption{Examples per finding and type on ReXGradient-160K~\cite{zhang2025rexgradient}, Open-I, PadChest-GR, and the food test set, with the example columns of Supplementary Table~\ref{stab:examples}. Open-I is listed with the labels of CheXbert~\cite{smit2020chexbert} and with the labels of its MeSH major topics. The food rows are recipe allergens, with the allergens in place of the findings. N/A marks sets without held-out articles.}
\label{stab:examples_other}
{\scriptsize\hyphenpenalty=10000\exhyphenpenalty=10000
\setlength{\tabcolsep}{3pt}
\renewcommand{\arraystretch}{0.95}
\begin{tabular}{@{}lrrrrrr@{}}
\toprule
Finding & \begin{tabular}[b]{@{}r@{}}Supported\end{tabular} & \begin{tabular}[b]{@{}r@{}}Contradicted\end{tabular} & \begin{tabular}[b]{@{}r@{}}Without\\support,\\present\end{tabular} & \begin{tabular}[b]{@{}r@{}}Without\\support,\\absent\end{tabular} & \begin{tabular}[b]{@{}r@{}}Supported,\\held-out\end{tabular} & \begin{tabular}[b]{@{}r@{}}Contradicted,\\held-out\end{tabular} \\
\midrule
\addlinespace[2.5pt]
\multicolumn{7}{@{}l}{\textbf{ReXGradient-160K}} \\
Atelectasis & 82 & 82 & 164 & 164 & 164 & 164 \\
Cardiomegaly & 150 & 150 & 150 & 150 & 0 & 0 \\
Consolidation & 8 & 8 & 8 & 8 & 4 & 4 \\
Fracture & 26 & 26 & 26 & 26 & 0 & 0 \\
Pleural effusion & 104 & 104 & 104 & 104 & 104 & 104 \\
Pneumothorax & 6 & 6 & 6 & 6 & 0 & 0 \\
Pulmonary edema & 32 & 32 & 64 & 64 & 32 & 32 \\
\addlinespace[2.5pt]
\multicolumn{7}{@{}l}{\textbf{Open-I}} \\
Atelectasis & 125 & 125 & 250 & 250 & 250 & 250 \\
Cardiomegaly & 414 & 414 & 414 & 414 & 0 & 0 \\
Consolidation & 30 & 30 & 30 & 30 & 15 & 15 \\
Fracture & 70 & 70 & 70 & 70 & 0 & 0 \\
Pleural effusion & 192 & 192 & 192 & 192 & 192 & 192 \\
Pneumothorax & 22 & 22 & 22 & 22 & 0 & 0 \\
Pulmonary edema & 35 & 35 & 70 & 70 & 35 & 35 \\
\addlinespace[2.5pt]
\multicolumn{7}{@{}l}{\textbf{Open-I (MeSH labels)}} \\
Atelectasis & 266 & 266 & 532 & 532 & N/A & N/A \\
Cardiomegaly & 568 & 568 & 568 & 568 & N/A & N/A \\
Consolidation & 58 & 58 & 58 & 58 & N/A & N/A \\
Fracture & 158 & 158 & 158 & 158 & N/A & N/A \\
Pleural effusion & 232 & 232 & 232 & 232 & N/A & N/A \\
Pneumothorax & 30 & 30 & 30 & 30 & N/A & N/A \\
Pulmonary edema & 34 & 34 & 68 & 68 & N/A & N/A \\
\addlinespace[2.5pt]
\multicolumn{7}{@{}l}{\textbf{PadChest-GR}} \\
Atelectasis & 259 & 259 & 518 & 518 & 518 & 518 \\
Cardiomegaly & 996 & 996 & 996 & 996 & 0 & 0 \\
Consolidation & 82 & 82 & 82 & 82 & 41 & 41 \\
Fracture & 298 & 298 & 298 & 298 & 0 & 0 \\
Pleural effusion & 410 & 410 & 410 & 410 & 410 & 410 \\
Pneumothorax & 22 & 22 & 22 & 22 & 0 & 0 \\
Pulmonary edema & 78 & 78 & 156 & 156 & 78 & 78 \\
\addlinespace[2.5pt]
\multicolumn{7}{@{}l}{\textbf{Food}} \\
Dairy & 264 & 264 & 264 & 264 & N/A & N/A \\
Egg & 191 & 191 & 191 & 191 & N/A & N/A \\
Gluten & 252 & 252 & 252 & 252 & N/A & N/A \\
Peanut & 8 & 7 & 15 & 15 & N/A & N/A \\
Shellfish & 16 & 16 & 16 & 16 & N/A & N/A \\
Soy & 37 & 37 & 37 & 37 & N/A & N/A \\
\bottomrule
\end{tabular}
}
\end{table}

\end{document}